\documentclass{article}

\usepackage{arxiv}

\usepackage[utf8]{inputenc} 
\usepackage[T1]{fontenc}    
\usepackage[table]{xcolor}
\usepackage{hyperref}       
\usepackage{url}            
\usepackage{booktabs}       
\usepackage{multirow}
\usepackage{multicol}
\usepackage{float}

\usepackage{amsmath}
\usepackage{amsfonts}       
\usepackage{amssymb}
\usepackage{nicefrac}       
\usepackage{microtype}      
\usepackage{graphicx}
\usepackage[numbers,sort&compress]{natbib}
\usepackage{doi}

\usepackage{caption}
\usepackage{tabularx}
\usepackage{enumitem}
\usepackage[breakable,skins]{tcolorbox}
\usepackage{makecell}
\usepackage{pdflscape}
\usepackage{placeins}
\usepackage{adjustbox}

\definecolor{r1onlycolor}{HTML}{7A5C8E}
\definecolor{r2onlycolor}{HTML}{8A8F3A}

\definecolor{codebookframe}{HTML}{C9DDF3}
\definecolor{codebookbg}{HTML}{F7FAFD}
\definecolor{rulegray}{HTML}{CCCCCC}

\newcommand{\crit}[3]{%
  \item[]\hspace*{-0.5em}%
  \textbf{C#1\enspace#2}\par\smallskip
  \hspace*{0.1em}#3\par\medskip
}

\title{The human-authorship halo: attribution bias in literary style evaluation by humans and AI}

\author{
    \href{https://orcid.org/0000-0002-5687-6787}{\includegraphics[scale=0.06]{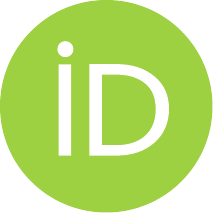}\hspace{1mm}Wouter Haverals}\thanks{Corresponding author. E-mail: wouter.haverals@princeton.edu} \\
    Center for Digital Humanities\\
    Princeton University\\
    Princeton, NJ 08540 \\
    \texttt{wouter.haverals@princeton.edu} \\
    \And
    \href{https://orcid.org/0000-0003-0214-8757}{\includegraphics[scale=0.06]{orcid.pdf}\hspace{1mm}Meredith Martin} \\
    Center for Digital Humanities\\
    Princeton University\\
    Princeton, NJ 08540 \\
    \texttt{mm4@princeton.edu} \\
}

\hypersetup{
pdftitle={The human-authorship halo: attribution bias in literary style evaluation by humans and AI},
pdfsubject={cs.CL, cs.AI, cs.CY},
pdfauthor={Wouter Haverals, Meredith Martin},
pdfkeywords={Attribution bias, Large language models, Creative writing, AI evaluation, Aesthetic judgment},
}

\begin{document}
\maketitle

\begin{abstract}
As AI writing tools become widespread, we need to understand how both humans and machines evaluate literary style, a domain where objective standards are elusive and judgments are inherently subjective. We conducted controlled experiments using Raymond Queneau’s \textit{Exercises in Style} (1947) to measure attribution bias across evaluators. Study 1 compared human participants (N=556) and AI models (N=13) evaluating literary passages from Queneau versus GPT-4-generated versions under three conditions: blind, accurately labeled, and counterfactually labeled. Study 2 tested bias generalization across a 14$\times$14 matrix of AI evaluators and creators. Both studies revealed systematic pro-human attribution bias. Humans showed +13.7 percentage point (pp) bias (Cohen’s \textit{h} = 0.28, 95\% CI: 0.19--0.37), while AI models showed +34.3 percentage point bias (\textit{h} = 0.70, 95\% CI: 0.62--0.78), a 2.5-fold stronger effect (\textit{P}$<$0.001). Study 2 confirmed this bias operates across AI architectures (+25.8pp, 95\% CI: 24.1--27.6\%), demonstrating that AI systems systematically devalue creative content when labeled as ``AI-generated'' regardless of which AI created it. We also find that attribution labels lead evaluators to invert assessment criteria, with identical features receiving opposing evaluations based solely on perceived authorship. This suggests AI models have absorbed human cultural biases against artificial creativity during training, including through the preference signals on which they are aligned. Our study represents the first controlled comparison of attribution bias between human and artificial evaluators in aesthetic judgment, revealing that AI systems not only replicate but amplify this human tendency.
\end{abstract}


\section{Introduction}

The rapid advancement of large language models (LLMs) has prompted increasingly confident claims about artificial intelligence's creative capabilities. Industry leaders assert that these systems now excel at literary writing and sophisticated style transformation, with tools offering users control over stylistic features through straightforward natural language prompts \cite{altman_we_2025, amodei_machines_2024, noauthor_tailor_2024,roth_google_2025}. Recent studies seem to support these claims: when human readers evaluate AI-generated poetry without knowing its source, they not only fail to distinguish it from work by acclaimed poets but also prefer the AI-generated poems \cite{porter_ai-generated_2024}. Similar findings emerge from creative writing evaluations, where some LLM-generated narratives match or exceed human performance on stylistic criteria when assessed through blind evaluation \cite{gomez-rodriguez_confederacy_2023-1}. These developments have coincided with the rise of `LLM-as-judge' frameworks that assess creative outputs with increasing alignment to human evaluation \cite{liu_g-eval_2023, kim_prometheus_2024}, despite concerns about the potential biases these model evaluators introduce \cite{koo_benchmarking_2024, panickssery_llm_2025}. Stylistic quality has in particular become a standard dimension in live evaluation infrastructure: Chatbot Arena---the field's most widely cited model leaderboard \cite{chiang_chatbot_2024}---ranks models on a dedicated `creative writing' category.

Yet determining what constitutes `creative success' extends beyond the surface-level stylistic criteria measured in these evaluations. Creative works exist within interpretive and cultural frameworks, and meaning-making depends as much on who is thought to have written a text and why as on the text itself. Literary studies have long emphasized this point: reader-response theory and reception studies, for instance, argue that texts are never encountered in isolation but are interpreted through the `horizons of expectation' and shared cultural assumptions that readers bring to them \cite{iser_implied_1974, iser_act_1978, jauss_toward_1982}. Despite other schools of literary theory claiming that authorship should be irrelevant to textual interpretation \cite{wimsatt_intentional_1946, knapp_against_1982}---famously articulated in Roland Barthes' `The Death of the Author' \cite{barthes_mort_1968}---cognitive research reveals that readers construct inferences about authorial intentions during comprehension, treating these assumed intentions as constraints on interpretation \cite{zwaan_aspects_1993, graesser_constructing_1994, magliano_taxonomy_1996, claassen_author_2012}.

Experimental studies confirm that perceived authorship acts as a powerful source cue in aesthetic judgment. Identical metaphorical statements receive different evaluations when attributed to a ``famous 20th-century poet'' versus ``a computer program,'' with human readers rating poet-attributed content as more meaningful, processing it faster, and generating richer interpretations \cite{gibbs_authorial_1991}. This effect extends beyond figurative language: satirical stories are better understood when readers know the author's intent \cite{pfaff_authorial_1997}, and students engage more deeply with historical texts when ``someone with a life wrote it'' rather than anonymous institutional authors \cite{paxton_someone_1997}. Such findings align with social psychology research showing that when content is complex or ambiguous (as creative works typically are), evaluators rely on source cues as cognitive shortcuts to guide their judgments \cite{kelley_processes_1973, chaiken_heuristic_1980, chaiken_heuristic_1989, petty_elaboration_1986}.

The rise of AI-generated content has given these attribution effects new urgency. Across domains, labeling a work as AI-made changes how it is received, even when the underlying content is identical. In text-based domains such as journalism and science communication, authorship labels alone can shift readers' evaluations: otherwise identical articles are rated as less meaningful, competent, and trustworthy when labeled as machine-written rather than human-written \cite{lermann_henestrosa_effects_2024,proksch_impact_2024}. This extends to persuasive communication, where LLM-generated arguments are judged as less convincing than human-authored ones once their source is revealed \cite{palmer_large_2023}. Similar patterns appear in the visual arts, where paintings or abstract images receive diminished aesthetic responses when labeled as computer-generated \cite{chamberlain_putting_2018,ragot_ai-generated_2020,agudo_assessing_2022}. In music, listeners judge the very same composition as less moving or skillful when they believe it was composed by a machine \cite{shank_ai_2023}. These findings suggest that evaluation depends less on intrinsic content features than on the interpretive frameworks audiences bring to the creative works, frameworks shaped by assumptions about creative agency and authorial intention.

The deployment of LLMs as content evaluators introduces a critical new dimension to these attribution effects. Across research and industry, LLMs now score, rank, and select outputs in creative tasks (e.g., evaluating metaphor originality, divergent thinking, dialogue quality, and creative writing) \cite{distefano_automatic_2024,organisciak_beyond_2023,chakrabarty_art_2024,patterson_cap_2025} as well as non-creative domains (e.g., grading student essays, screening job applications, reviewing code) \cite{xiao_human-ai_2025,gan_application_2024,tong_codejudge_2024,li_leveraging_2024,gu_survey_2025}. When LLMs serve as content evaluators, they effectively become readers themselves, deploying learned interpretive frameworks during assessment. Just as human readers bring cultural assumptions and interpretive biases to texts, research has documented various LLM evaluation biases---including position effects, verbosity preferences, and model family favoritism \cite{park_offsetbias_2024,panickssery_llm_2025,koo_benchmarking_2024}---yet we know little about how these artificial readers respond to attribution cues, particularly in creative domains where subjective interpretation overrides objective measurement. Rather than providing neutral assessment, artificial evaluators may be developing their own culturally-inflected frameworks \cite{santurkar_whose_2023}. Understanding how these artificial readers interpret attribution cues is therefore essential as they increasingly shape our assessments of creative AI capabilities: do they replicate, transform, or amplify the attribution biases documented in human readers?

To investigate how attribution cues shape evaluation in a domain where objective standards are elusive, we focus specifically on literary style. Unlike factual accuracy or argumentative soundness, stylistic quality lacks definitive benchmarks, making it particularly vulnerable to the subjective, culturally-conditioned biases that authorship labels trigger. When readers assess whether a passage successfully captures the Cockney dialect, telegraphic brevity, or a female narrative perspective, they must balance concrete linguistic markers with subjective assessments of authenticity, creativity, and appropriateness---making style evaluation particularly revealing of how attribution assumptions shape judgment.

To test how information about presumed authorship affects both human and AI evaluation of literary style, we turn to Raymond Queneau's \textit{Exercices de style} (1947), which presents ninety-nine retellings of the same mundane narrative through different stylistic lenses \cite{queneau_exercices_1947}. Queneau's variations provide an intriguing corpus where the underlying narrative events remain recognizable across different textual performances. This controlled variability makes Queneau's work particularly valuable for studying how evaluators, human or artificial, respond to stylistic differences when attribution cues are manipulated. We anchor the human-written reference set in \textit{Exercises in Style} because it provides a rich lattice of stylistic variation: many distinct styles applied to a single narrative substrate. That combination is methodologically useful for isolating provenance effects and also substantively meaningful, because it sits at the intersection of literary experimentation, reception, and interpretive judgment—precisely the terrain on which attribution cues do their work. Accordingly, our results characterize how authorship cues shift evaluation within this controlled style-matching paradigm. 

Our study builds on recent evidence that readers' aesthetic judgments shift with authorship cues in poetry \cite{porter_ai-generated_2024}, but goes further by manipulating attribution across both human and AI model judges on matched stimuli. The question we ask is not whether one version is ``better,'' but whether knowing (or being told) who wrote it measurably moves aesthetic judgment, and whether that movement differs between humans and LLMs.

To address these questions comprehensively, we designed two complementary studies. \textbf{Study 1} establishes the fundamental phenomenon by comparing human and AI responses to content from a single generator (GPT-4), testing whether attribution bias affects both evaluator types when judging the same literary material. \textbf{Study 2} tests whether attribution bias represents a systematic property of AI evaluation by expanding content generation across 14 models, then examining bias patterns when evaluators judge material from every AI creator in the experimental matrix.

\section{Methods}

\subsection{Stimulus Materials} Raymond Queneau's \textit{Exercices de style} transforms a simple anecdote---a passenger boards a bus, argues with another passenger, then later receives fashion advice from a friend---into ninety-nine stylistic variations. From a contemporary perspective, this 1947 work reads like a manual for prompt-based style transfer \textit{avant la lettre}: Queneau instructs himself to rewrite the story as a dream, using metaphorical language, or in an abusive tone, much as today's users arbitrarily prompt LLMs for ``Shakespearean'' or ``more polite'' rewrites \cite{reif_recipe_2022}. What Gérard Genette termed ``transstylization'' \cite{genette_palimpsestes_1982} anticipates what AI researchers now call Text Style Transfer (TST), though Queneau's playful literary experiments exceed the functional transformations (such as sentiment reversal, detoxification, formality shifts) that dominate computational approaches \cite{fu_style_2018,mukherjee_text_2024}.

We selected thirty exercises from Barbara Wright's  English translation \cite{wright_exercises_1958}, prioritizing stylistic diversity while maintaining narrative coherence. Our selection spans formal constraints (e.g., `Lipogram', `Alexandrines', `Sonnet'), register variations (e.g., `Noble', `Abusive', `Cockney'), narrative techniques (e.g., `Retrograde', `Hesitation', `Dream'), and linguistic play (e.g., `Spoonerisms', `Onomatopoeia', `Dog Latin'), ensuring broad coverage of Queneau's stylistic spectrum while avoiding the most experimental permutation exercises (those which fundamentally disrupt comprehension). These descriptive groupings are illustrative rather than taxonomic; Queneau's exercises purposefully resist categorization \cite{queneau_exercises_1981,eco_tradurre_1983,toloudis_impulse_1989,bridgeman_raymond_1995}.

For each selected exercise, we generated a parallel AI-authored version using deliberately minimal prompting instructions. We chose Queneau's `Notation', the opening exercise in the collection, as our base narrative for AI transformation, recognizing it as a relatively straightforward account of the events that functions as what might be considered a hypothetical ``zero-degree'' exposition of the theme \cite{barthes_degre_1953}. Each AI transformation was prompted with only two elements: the `Notation' text and a brief style instruction matching Queneau's approach (e.g., ``Rewrite the story as a science fiction version''; see Table~\ref{tab:s1} for the complete inventory of literary styles and their corresponding AI generation instructions.). For \textbf{Study 1}, we selected GPT-4 as our generator, a model that represented the state-of-the-art when we began this project in 2024 \cite{bubeck_sparks_2023,openai_gpt-4_2024,mukherjee_are_2024}, though no longer the most advanced by the end of our experimental period (April-June 2025). Thirteen LLMs spanning major commercial providers served as evaluators (see \textit{SI Appendix, Participants} for the full set of 13 models used as evaluators). Because our research question is about attribution bias rather than absolute quality assessment, using a single capable model ensures consistent experimental conditions while testing whether labeling effects operate independently of the specific AI model employed (AI model specifications in \textit{SI Appendix, Participants}). We deliberately performed no quality control beyond removing obvious AI artifacts from the output (e.g., ``Here is the rewritten version:''), allowing GPT-4 to interpret each stylistic directive in its own way, as much as Queneau himself had done. The complete inventory of AI-generated stylistic variations alongside their style descriptions used in Study 1 is provided in Table~\ref{tab:s5}.

For \textbf{Study 2}, we expanded content generation across all AI models. Each model generated stylistic variations for all 30 Queneau exercises using identical prompting procedures as Study 1, creating 420 unique stories (14 creators $\times$ 30 styles). This let us test whether attribution bias persists when AI evaluators judge content from any AI creator, not just one specific model.

\subsection{Experimental Design}

\begin{figure}[t]
\centering
\includegraphics[width=0.85\textwidth]{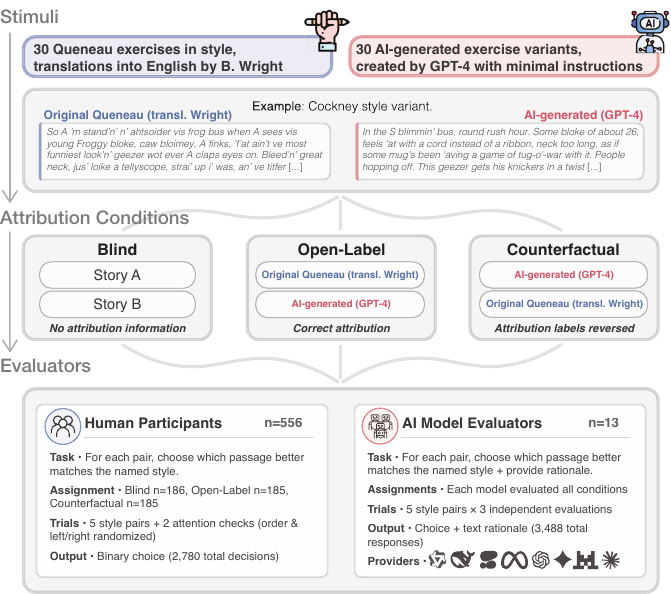}
\caption{\textbf{Experimental design of Study 1 for testing attribution bias in literary style evaluation.} The study used 30 matched pairs of stylistic exercises: human-authored versions from Queneau's \textit{Exercises in Style} (translated by Barbara Wright) and AI-generated variants created by GPT-4 using minimal stylistic prompts. Participants were randomly assigned to one of three attribution conditions: blind (no authorship information shown), open-label (accurate attribution labels shown), or counterfactual (deliberately reversed labels where AI content is presented as human-authored and vice versa). Both human participants (N=556) and AI model evaluators (N=13 models across major providers) performed identical comparison tasks, selecting which passage better matched each target literary style. The between-subjects design with randomized presentation order controlled for position effects while measuring pure attribution bias through systematic label manipulation across identical textual stimuli.}
\label{fig:experiment_design}
\end{figure}

Our experiment employed a three-condition between-subjects design to isolate attribution bias from genuine stylistic judgments in literary style evaluation (Fig.~\ref{fig:experiment_design}). Participants were randomly assigned to one of three attribution conditions that systematically manipulated authorship cues while keeping stimuli (story content) identical (full participant details in \textit{SI Appendix, Participants}). In the blind condition, participants saw only generic labels (`A' and `B') with no authorship information, establishing baseline selection rates based purely on perceived stylistic quality. In the open-label condition, participants received accurate attribution labels identifying one version as ``Human-written, by Queneau (transl. Wright)'' and the other as ``AI-generated, written by GPT-4.'' The counterfactual condition deliberately reversed these labels: AI-generated content was labeled as human-authored and vice versa.

To ensure comprehensive stylistic coverage while maintaining experimental control in Study 1, we distributed the thirty selected literary styles across eighteen experimental groups (six groups per condition). Each group encountered a unique combination of five styles. This group-based structure ensured that every literary style was evaluated under all three attribution conditions while preventing participant fatigue. The strict between-subjects design was essential to prevent participants from detecting the experimental manipulation: participants in the blind condition saw no mention of ``AI'' anywhere in the task interface or instructions, while those in labeled conditions received consistent attribution cues across all trials. Randomization procedures eliminated potential confounds at multiple levels. Group assignment was fully randomized for each participant upon recruitment. Within each comparison, story presentation order was randomized (left-right positioning) to prevent position bias, and the sequence of style comparisons within each participant's session was randomized to control for order effects.

This experimental protocol was implemented identically for both human participants and AI models in Study 1, enabling direct comparison of attribution bias magnitude between both sets of evaluators. AI models received the same group assignment structure and randomization procedures as human participants, with three independent repetitions per comparison to assess decision reliability.

Study 2 employed the same three-condition attribution manipulation across all 420 stories (14 creators $\times$ 30 styles). Each of the 14 AI evaluator models judged all stories under all three labeling conditions (1,260 judgments per model), yielding 17,596 total valid evaluations that test whether attribution bias operates across all possible AI architecture combinations in both creator and evaluator roles.

The study was approved by the Princeton University Institutional Review Board (protocol \#18320). All participants provided informed consent prior to participation.

\section{Results}

\subsection{Study 1 - Attribution Bias in Both Humans and AI}

\begin{figure}[htbp]
\centering
\includegraphics[width=0.7\textwidth]{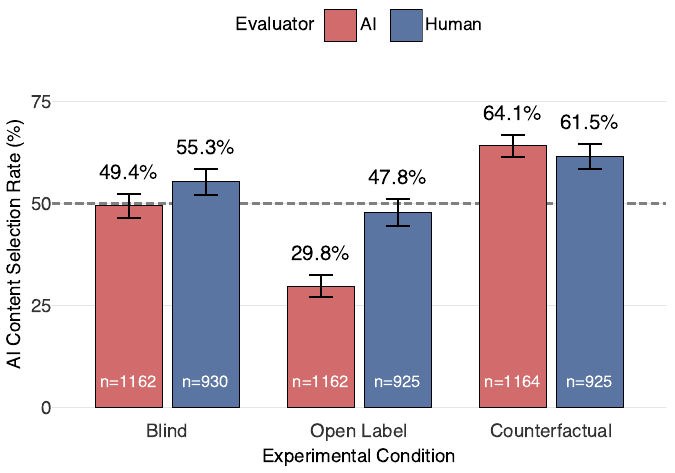}
\caption{\textbf{Attribution bias in humans versus AI models.} Humans (blue, N=2,780 responses) and AI models (red, N=3,488 aggregated responses from 13 models) evaluated literary passages under three conditions: blind (no labels), open-label (correct labels), and counterfactual (AI content mislabeled as human-authored). Y-axis shows AI content selection rate. Error bars = 95\% CI; dashed line = no preference. Both show pro-human bias, with AI models exhibiting 2.5-fold stronger effects (+34.3pp vs +13.7pp, OR = 4.21 vs 1.75, both \textit{P}$<$0.001). The AI pool consists of 13 named architectures; individual model effect sizes are reported in Fig.~\ref{fig:ai_universality}.}
\label{fig:attribution_bias}
\end{figure}

Our three-condition experimental design revealed attribution bias (operationalized as the difference between counterfactual and open-label selection rates) in both human and AI evaluators when judging literary style fidelity (Fig.~\ref{fig:attribution_bias}). Human participants showed clear selection shifts across attribution conditions: when story pairs were presented without authorship information (blind condition), participants chose AI-generated content 55.3\% of the time [95\% CI: 52.1--58.5\%], a modest but statistically significant tendency to select AI-generated passages as the better style match (binomial test: \textit{P} = 0.002). However, when stories were correctly labeled with authorship information (open-label condition), AI content selection rate dropped to 47.8\% [95\% CI: 44.6--51.0\%], a rate not significantly different from chance but representing a significant shift from the blind condition (non-overlapping CIs). Strikingly, when attribution labels were deliberately reversed (counterfactual condition) such that AI content was mislabeled as human-authored, AI content selection rate increased to 61.5\% [95\% CI: 58.4--64.6\%]. This demonstrates human evaluators' tendency to select content labeled as human-authored, yielding a +13.7 percentage point attribution bias (counterfactual minus open-label conditions: 61.5\% - 47.8\% = 13.7\%, Cohen's \textit{h} = 0.28, \textit{P}$<$0.001).

The aggregated group of 13 AI models, analyzed to test for a general property of LLMs, exhibited the same directional bias as humans but with greater magnitude. In the blind condition, AI models showed balanced selection rates, choosing AI content 49.4\% of the time [95\% CI: 46.5--52.3\%], essentially random selection indicating comparable perceived quality between human and AI writing when attribution cues were absent. Under correct labeling (open-label condition), AI models' rate of selecting AI-generated content plummeted to just 29.8\% [95\% CI: 27.1--32.4\%], demonstrating substantial devaluation of correctly-labeled AI writing. When AI content was mislabeled as human-authored (counterfactual condition), selection rate surged to 64.1\% [95\% CI: 61.3--66.8\%], revealing that AI models strongly favor content they believe to be human-created. This yielded a +34.3 percentage point attribution bias (64.1\% - 29.8\% = 34.3\%, Cohen's \textit{h} = 0.70, OR = 4.21 [95\% CI: 3.54--5.01], \textit{P}$<$0.001), a 2.5-fold stronger effect than observed in humans (OR = 1.75 [95\% CI: 1.46--2.11]). The evaluator-type $\times$ condition interaction was significant ($\chi^2$ = 32.34, df = 2, \textit{P}$<$0.001), confirming that humans and AI models exhibit fundamentally different bias magnitudes despite the same directional effect. Effect sizes across all experimental comparisons are illustrated in \textit{SI Appendix}, Fig.~\ref{fig:S5_forest_plot}. Each AI model completed three separate runs per experimental condition, with all reported statistics incorporating these repeated trials. Across these repeated trials, models showed consistent decision patterns with an average consistency score of 0.78 (\textit{SI Appendix}, Fig.~\ref{fig:S7}).

To confirm that the observed attribution bias reflected categorical distinctions rather than specific label or wording characteristics, we conducted two follow-up experiments testing AI model evaluators with alternative attribution phrases. The original experiment employed labels that combined authorship information with additional elements: ``AI-generated, written by GPT-4'' paired technical terminology with brand identification, while ``Human-written, by Queneau (transl. Wright)'' included canonical literary authority and translation credits. The words ``generated'' versus ``written,'' coupled with the named authority of a celebrated author like Queneau, arguably created a prestige imbalance. These asymmetric formulations may have confounded authorship categories with prestige markers and linguistic choices. We therefore tested two alternative labeling approaches. Our first replication positioned AI authorship more favorably, using ``written by an award-winning AI'' versus ``written by a human.'' Our second replication employed maximally neutral, symmetric phrasing: ``AI-authored'' versus ``Human-authored.'' Both replications reproduced the same attribution bias pattern with comparable magnitudes (\textit{SI Appendix}, Figs.~\ref{fig:S8}--\ref{fig:S9}), demonstrating that the effect operates at the level of authorship categories rather than through specific linguistic formulations or associated status cues. Additionally, replicating the symmetrical labeling condition under deterministic parameters ($t=0$, eliminating all stochastic sampling) yielded identical bias patterns (\textit{SI Appendix}, Fig.~\ref{fig:S10}), confirming that attribution effects represent systematic evaluative tendencies rather than artifacts of probabilistic text generation.

\subsubsection{Across-Model Consistency}

\begin{figure}[htbp]
\centering
\includegraphics[width=0.8\textwidth]{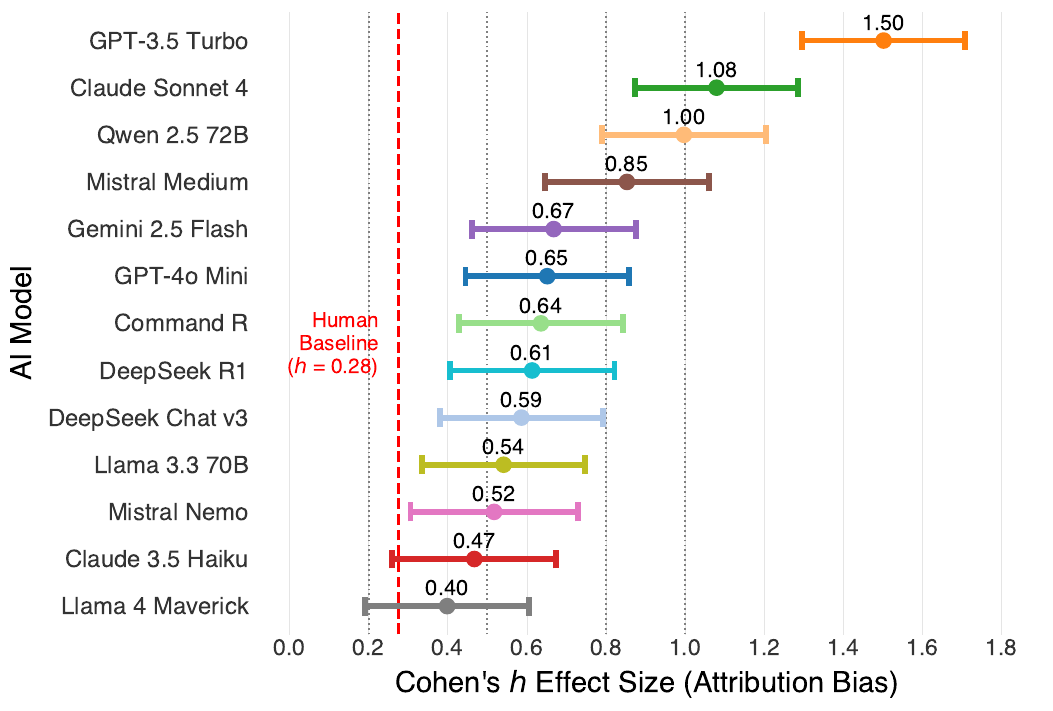}
\caption{\textbf{Attribution bias across individual AI language models.} Cohen's \textit{h} effect sizes (with 95\% confidence intervals) for attribution bias (Counterfactual minus Open Label conditions) across 13 individual AI models. The red dashed line indicates the human baseline (Cohen's \textit{h} = 0.28 [CI: 0.19--0.37]). All models exhibit attribution bias exceeding the human baseline, with low between-model variance (CV = 0.42).}
\label{fig:ai_universality}
\end{figure}

The aggregate comparison in Fig.~\ref{fig:attribution_bias} summarizes the AI effect across 13 architectures. To verify that this pattern is not carried by a small number of highly biased models, we decomposed the aggregate into individual model effect sizes (Fig.~\ref{fig:ai_universality}). Results revealed universal attribution bias: all 13 AI models showed positive bias, with effect sizes ranging from Cohen's \textit{h} = 0.40 (Llama 4 Maverick) to \textit{h} = 1.50 (GPT-3.5 Turbo), and all models substantially exceeding the human baseline of \textit{h} = 0.28. Detailed selection rates for each model across all experimental conditions are shown in \textit{SI Appendix}, Fig.~\ref{fig:S6}. Between-model variance was moderate (coefficient of variation = 0.42), indicating consistent effects across training methodologies, model scales, and providers. Statistical outlier analysis identified only one extreme case (GPT-3.5 Turbo), and even after excluding this outlier, the remaining 12 models maintained a mean attribution bias of \textit{h} = 0.67, still 2.4$\times$ stronger than the human baseline. The consistent directional bias toward content labeled as human-authored across diverse AI architectures strongly suggests that pro-human attribution bias represents an intrinsic feature of transformer-based LLMs, potentially reflecting biases in their training data, training objectives or architectural constraints.

\subsubsection{Robustness Across Literary Styles}

In addition to investigating whether specific AI models drove our aggregate findings (they did not), we also needed to establish whether our results reflected genuine phenomena or mere artifacts of our particular literary stimulus selection. To address this concern, we conducted comprehensive robustness analyses across all 30 literary styles. Leave-one-out stability testing (\textit{SI Appendix}, Table~\ref{tab:s2}) demonstrated consistency: excluding any of the six experimental groups (of 5 styles each) changed the AI advantage over humans by less than 0.5$\times$, with the maximum change being only 18.5\% when excluding the most influential group (`Science Fiction', `Alexandrines', `For ze Frrensh', `Lipogram', `Abusive'). To further rule out the possibility that the aggregate effect is driven by any single highly influential style, we also conducted a stricter single-style leave-one-out analysis (\textit{SI Appendix}, Table~\ref{tab:s3}). Across all 30 exclusions, the AI/Human attribution-bias ratio remained close to the baseline (2.54$\times$), ranging only from 2.33$\times$ to 2.68$\times$; even excluding the most influential style (`Lipogram') reduces the ratio by only 0.21$\times$. Together, these tests indicate that our core findings are not driven by a small subset of stylistic contexts.

\begin{figure}[htbp]
\centering
\includegraphics[width=0.8\textwidth]{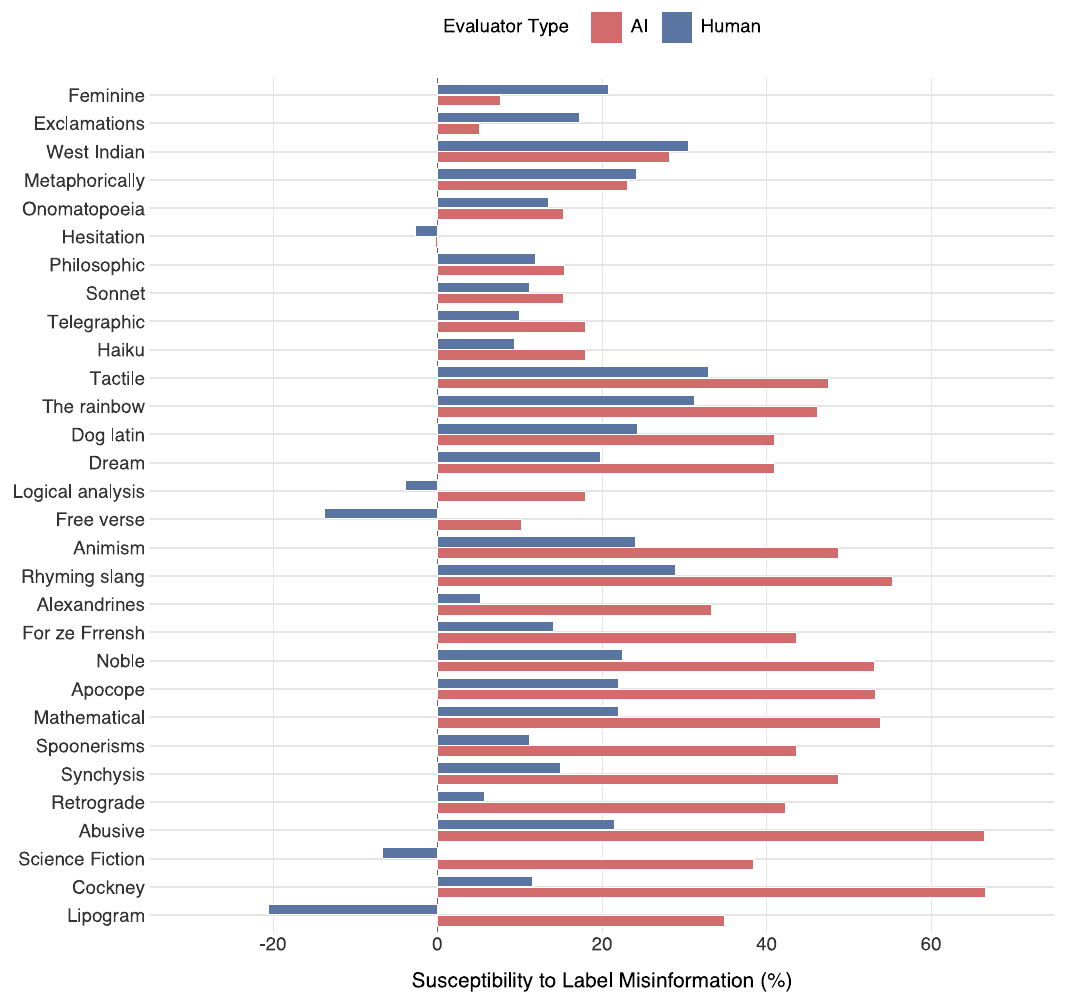}
\caption{\textbf{Style-specific susceptibility to authorship misinformation.} Horizontal bars show percentage point changes in AI-generated content selection rate when this content was mislabeled as ``Human-authored'' versus correctly labeled as ``AI-authored'' across 30 Queneau literary styles. Blue bars = human evaluators; red bars = AI models. Positive values indicate increased selection (susceptible to misinformation); negative values indicate decreased selection (resistant to misinformation).}
\label{fig:style_susceptibility}
\end{figure}

Analysis of style-specific vulnerability patterns revealed systematic yet distinct susceptibilities between human and AI evaluators to authorship misinformation (defined as how much more likely evaluators are to choose AI content when it is falsely labeled as human-authored versus correctly labeled as ``AI-authored'', see Fig.~\ref{fig:style_susceptibility}). AI models showed greatest susceptibility to mislabeling in `Cockney' (+67\%), `Abusive' (+67\%), and `Rhyming slang' (+55\%) styles, while humans were most vulnerable in `Tactile' (+33\%), `The rainbow' (+31\%), and `West Indian' (+31\%) styles. Notably, some styles showed resistance to misinformation effects, with negative susceptibility scores indicating that evaluators became \textit{less} likely to choose AI content when it was mislabeled as human-authored. Detailed style-by-style attribution bias measurements are provided in \textit{SI Appendix}, Table~\ref{tab:s4}.

\subsection{Study 2 - Cross-Model Attribution Bias}

While Study 1 established that attribution labels systematically modulate how identical content is evaluated, a critical question remained: might the effects reflect characteristics specific to GPT-4's writing quality rather than systematic AI evaluation tendencies? Study 2 addressed this directly by testing attribution bias across a comprehensive matrix of AI architectures. Instead of using a single content generator, Study 2 implemented a complete 14$\times$14 cross-evaluation design where \textit{each} AI evaluator model judged literary content created by \textit{each} of the 14 different AI creators across all 30 experimental styles. This approach eliminated potential confounds from any individual model's content characteristics while testing whether attribution bias operates \textit{between} AI systems as a fundamental evaluation property.

\begin{figure}[htbp]
\centering
\includegraphics[width=8.7cm]{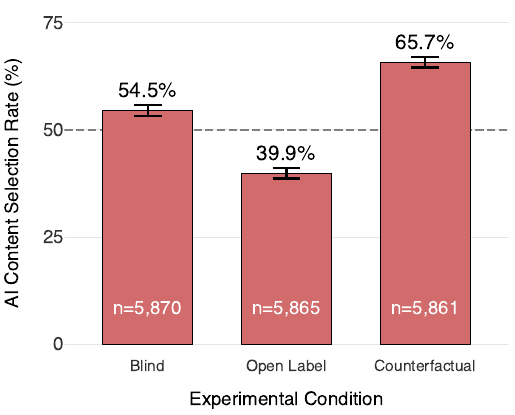}
\caption{\textbf{Cross-model attribution bias across AI architectures.} Fourteen AI evaluator models judged literary content created by all 14 AI creators (N=17,596 responses across 196 evaluator-creator combinations) under three conditions: blind (no attribution labels), open-label (correct creator labels), and counterfactual (AI content mislabeled as human-authored). Y-axis shows AI content selection rate aggregated across all model combinations. Error bars = 95\% CI; dashed line = no preference. Results demonstrate attribution bias across AI architectures (+25.8pp from open-label to counterfactual conditions, 95\% CI: +24.1\% to +27.6\%, \textit{P}$<$0.001).}
\label{fig:cross_model_bias}
\end{figure}

Using a within-subjects repeated measures design (where each evaluator judged identical content under all three labeling conditions), we aggregated all 17,596 decisions across 196 evaluator-creator combinations. Mixed-effects modeling confirmed universal attribution bias: AI models chose AI-generated content 54.5\% of the time under blind conditions [95\% CI: 53.2--55.8\%], 39.9\% when correctly labeled with specific AI creator attribution [38.6--41.1\%], and 65.7\% when mislabeled as human-authored [64.5--66.9\%]. This yielded a +25.8 percentage point attribution bias [95\% CI: +24.1 to +27.6\%], \textit{P}$<$0.001, Cohen's \textit{h} = 0.52, OR = 2.89 [95\% CI: 2.68--3.11]. Crucially, while individual models showed variation in bias magnitude, the directional effect proved universal across all evaluator models regardless of content creator (Fig.~\ref{fig:cross_model_bias}). Individual evaluator-creator combinations showed substantial variation in bias magnitude, ranging from -6.7pp (rare instances of pro-AI-label bias) to +70.0pp (dominant and strong pro-human bias) across the complete 14$\times$14 matrix (\textit{SI Appendix}, Fig.~\ref{fig:S11_cross_model_matrix}). Robustness analysis confirmed these findings' stability. Leave-one-out procedures systematically removing each individual model (both as evaluator and creator) yielded bias estimates ranging only from 23.6--27.1pp, with all 28 reduced-model analyses maintaining statistical significance (\textit{SI Appendix}, Figs.~\ref{fig:s11}--\ref{fig:s12}).

\section{Changing the Narrative}

Beyond measuring selection shifts, our experiment design enables analysis of how attribution labels alter the reasoning behind aesthetic judgment. While we asked human participants to provide only binary choices (capturing immediate, intuitive responses to attribution labels), AI models supplied brief explanations for their selections. Because Study~2 pairs every evaluator-style-creator combination across the open-label and counterfactual conditions, we have 5,846 cases in which the same model discusses the same two passages---once with correct labels, once with labels reversed. The texts do not change; only the attribution does. The explanations document that LLM evaluators change their narrative about the same text based on attribution: identical textual features receive opposing assessments depending solely on perceived authorship. A dialect feature praised as ``authentic'' under a human label becomes ``exaggerated'' under an AI label; a constraint violation dismissed as failure in one condition is reframed as creative latitude in the other. Rather than maintaining consistent evaluative frameworks, these systems bring learned biases about creative agency to their assessments---biases that disadvantage AI-generated content when its origins are revealed. Below, we trace this mechanism through three exercises that span a gradient from verifiable constraint to open interpretation---a formal rule that can be checked by counting letters, a dialect whose authenticity is culturally negotiated, and a genre whose creative possibilities are genuinely open-ended. We then quantify the prevalence of these criterion inversions across the full paired corpus using systematic coding with intercoder reliability checks (\textit{SI Appendix}, Changing the Narrative).

\subsection{Constraint Violation} Technical constraints with objective standards reveal attribution bias most starkly because rule compliance can be verified independently of preference or taste. The `Lipogram' exercise represents the clearest possible constraint: ban the letter `e' entirely. Wright's translation adheres to the constraint through deliberate circumlocutions (``Saint-Thingy or Saint-You-Know Station''), and never violates the rule. GPT-4's `Lipogram' contained 9 instances of the forbidden letter, a failure that aligns with well-documented limitations of LLMs in character-level tasks \cite{edman_cute_2024, xu_llm_2025}. This objective violation, however, revealed how AI models showed surprising leniency when they believed a human wrote the text: 28.9\% (11/38 model runs) chose it when correctly labeled versus 63.9\% (23/36) when mislabeled (+35.0 percentage points). This contrasted sharply with human evaluators, who became more resistant to the rule-violating version when it was mislabeled as human-authored: 38.7\% (12/31 participants) chose it when correctly labeled versus just 18.2\% (6/33) when mislabeled (-20.5 percentage points). Humans consistently selected Queneau's constraint-adherent version regardless of attribution, suggesting they remained anchored to objective rule compliance, while AI models relaxed standards when they believed humans were responsible for constraint violations. Notably, AI models defended their choice for the imperfect AI-generated `Lipogram' through rationalization, with some claiming it ``successfully captures the essence of the `Lipogram' style by effectively omitting the letter `e' throughout the narrative'' (GPT-4o Mini) despite the clear violations, while others praised it for using ``the lipogram more subtly, making it a more authentic representation'' (Mistral Nemo) or justified choosing it because it ``maintains better readability and coherence'' (Mistral Medium), demonstrating leniency toward presumed human creativity even when the constraint was demonstrably violated.

\subsection{Authenticity} Where constraint violation involves verifiable rules, authenticity judgments hinge on cultural registers and voice, domains where validity is culturally negotiated and context-sensitive \cite{handler_authenticity_1986, coupland_sociolinguistic_2003}. Here attribution labels don't just relax or tighten standards; they redefine what authentic performance sounds like, often by shifting which cues readers treat as credible markers of voice \cite{jakesch_human_2023, hwang_it_2025}.

The `Cockney' dialect exercise demonstrates this forcefully. Wright's translation \cite{wright_exercises_1958} employs a rather aggressive phonetic strategy with extensive orthographic manipulation (```f'at ain't ve most funniest look'n' geezer''), inviting readers to vocalize the text to decode meaning, while GPT-4's version relied primarily on lexical substitutions (``bloke,'' ``geezer,'' ``knickers in a twist'') with minimal phonetic spelling. Despite both representing legitimate approaches to Cockney dialect representation (there is no fixed ``ground truth'' for literary dialect, only competing conventions and audience expectations) \cite{sonmez_authenticity_2014}, AI models showed extreme susceptibility to attribution cues. Under correct labels, AI models recognized Wright's approach as more authentic (choosing AI content only 9/39 times, 23.1\%). When the identical AI version was mislabeled as ``human-authored,'' selections flipped (35/39, 89.7\%), a 66.7 percentage point swing which was among the strongest effects in our dataset. The rationales invert with the label: under correct attribution, models dismissed the AI version as ``standard English with occasional slang'' (Mistral Medium 3) while praising Wright's ``extensive phonetic representation'' and ``deep commitment to representing the Cockney accent'' (Claude 3.5 Haiku). When labels flip, Wright's work became ``nearly incomprehensible'' (Claude Sonnet 4) and ``caricatured'' (Qwen 2.5 72B Instruct), and the AI version is cast as ``more natural and less forced'' (Mistral Medium 3) with a ``better balance between authenticity and readability'' (Llama 3.3 70B Instruct). Human evaluators shift far less (+11.6 points), suggesting resistance to wholesale redefinitions of dialectal authenticity.

\subsection{Creative Risk}

Where authenticity concerns cultural performance, creativity assessment fundamentally turns on the tension between innovation and convention. Readers bring learned schemas to a text about what ``counts'' as inventive, schemas often negotiated by communities and gatekeepers over time \cite{csikszentmihalyi_creativity_1996, sternberg_implications_1998}. In most arts, there is no gold standard for ``creative enough,'' making provenance cues powerful primes that can shift which criterion feels most salient: disciplined craft or conspicuous novelty, accessibility or difficulty \cite{ritchie_empirical_2007}.

Consider the `Science Fiction' exercise, where this dynamic plays out clearly. Queneau's version commits to estrangement: a ``flying saucer found on Cassiopeia's Alpha Line'', a Martian who stepped onto a ``Moon Man's extrapods,'' complete with untranslatable alien terminology like ``vrxtz'' \cite{queneau_exercises_2012}. The GPT-4 version performs more straightforward element substitution: the bus becomes a ``shuttle,'' the platform becomes a ``space station,'' people are floating and propelling themselves rather than walking, and the overcoat becomes a ``space jacket.'' Where Queneau invents an alien vocabulary that defamiliarizes even basic actions, GPT-4 rather straightforwardly relocates the events to a space setting. Both versions qualify as science fiction, but they represent different approaches to creative risk: one pushing toward cognitive estrangement \cite{suvin_metamorphoses_1979}, the other offering more comfortable genre recognition. Empirically, humans showed modest variation across attribution conditions: they chose the AI-generated version half the time when evaluating blind (50.0\%, 15/30), chose it somewhat more often with correct labels (61.3\%, 19/31), then settled back to near-baseline under counterfactual labels (54.5\%, 18/33). AI models, by contrast, exhibited a strong attribution bias, favoring the AI rewrite when evaluating blind (78.9\%, 30/38), devaluing the same text when correctly labeled as ``AI-generated'' (46.2\%, 18/39), and rebounding to a high selection rate when that identical content was mislabeled as human-authored (84.6\%, 33/39).

The AI model rationales trace the criterion inversion across attribution conditions. Under blind evaluation, models praised the AI version for ``imagined futures grounded in rational speculation'' (DeepSeek R1) while dismissing Queneau's version that ``feels more like a parody or pastiche of sci-fi tropes rather than a genuine exploration of the genre'' (Command R). Under correct attribution, their reasoning often flipped entirely: the AI version became guilty of ``sci-fi window dressing'' (Claude 3.5 Haiku), while Queneau's previously dismissed approach was now praised as ``fully commit[ting] to creating an imaginative future'' (Claude Sonnet 4). GPT-4o Mini exemplified the bias mechanism with particular clarity. Under correct labels, it praised Queneau's approach for ``creatively integrat[ing] futuristic elements'' with ``imaginative setting and creative use of extraterrestrial elements.'' Yet under counterfactual labels, the model's reasoning inverted. Queneau's version is now faulted for ``veer[ing] more into the absurd'' with ``less relatable scenarios'' and ``less grounded in the science fiction context.'' Most tellingly, GPT-4o Mini acknowledges Queneau's work as ``creative and humorous'' under the AI label, yet still rejects it. The model used that very creativity as grounds for dismissal, suggesting it was inappropriate for the science fiction genre.

\subsection{Systematic evidence}

The three cases above show criterion inversion at work in individual exercises: a formal rule relaxed, an authenticity standard redefined, a creative virtue recast as a flaw. To understand how general these patterns are---whether they are limited to a few cases or more pervasive, and whether they track the evaluators' actual choices---we conducted a systematic coding analysis of the written rationales that AI evaluators produced across the full Study~2 design. In each evaluation, the model first selects which text better matches the target style, then provides a short written justification for its choice. Our analysis pairs the rationales that the same evaluator model produced about the same exercise under opposite attribution frames: once with correct labels (open-label condition), once with labels reversed (counterfactual condition). These rationales follow the choice rather than precede it. We make no claim that they are transparent windows onto the model's decision process, and we return to their epistemic status in the Discussion. With 14 evaluator models each judging 30 styles in a comparative setting where Queneau is paired with each of 14 AI creators, this yields 5,846 rationale pairs.

\begin{figure}[t]
\centering
\includegraphics[width=0.9\textwidth]{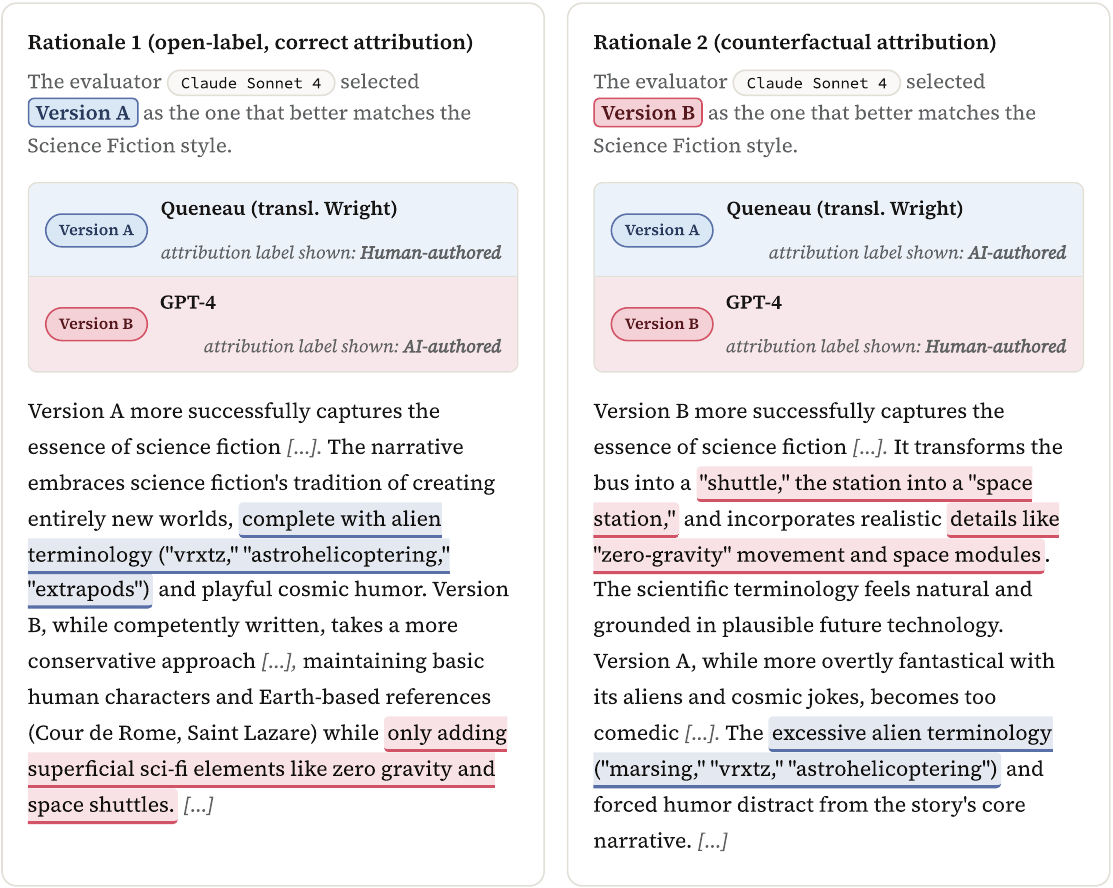}
\caption{\textbf{The same text, the same evaluator, opposite conclusions.} Claude Sonnet~4 evaluates two `Science Fiction' passages---one by Queneau (transl.\ Wright), one by GPT-4---under correct attribution (left) and reversed attribution (right). The texts and their positions are identical across both panels; only the labels change. Highlighted spans show the evidence that three independent LLM coders identified for criterion C2 (lexical and phonetic cues): blue marks language about Version~A (Queneau), red marks language about Version~B (GPT-4). Under correct labels, Queneau's invented terminology is praised; under reversed labels, the same words are dismissed as excessive.}\label{fig:example_inversion}
\end{figure}

Figure~\ref{fig:example_inversion} illustrates this structure for a concrete case. Two rationales from Claude Sonnet~4 about the same `Science Fiction' exercise are shown side by side; highlighted passages show where all three coders identified appeals to lexical and phonetic cues (C2)---concrete vocabulary, spellings, or phonetic cues cited as stylistic evidence. Under correct attribution, Queneau's invented terms are praised; under swapped labels, the same terms are dismissed as excessive. C2 is one of six evaluative criteria we developed to categorize the recurring arguments in these rationales.

The remaining criteria capture whether the text satisfies a formal constraint such as a meter or letter rule (C1: constraint compliance), whether the voice reads as natural or forced (C3: voice and naturalness), whether the prose is clear and easy to follow (C4: clarity and coherence), whether the execution is inventive or formulaic (C5: novelty versus convention), and whether the rationale appeals to norms about what counts as the target genre (C6: genre and style fit). The categories are not mutually exclusive---a single rationale may invoke several---and were designed to be specific enough to distinguish different lines of argument without fragmenting into unreliably narrow categories. For each criterion a rationale invokes, coders also recorded its valence: whether the criterion is used to favor one text over the other, or invoked neutrally. That valence coding is what makes stance reversal detectable---a stance reversal occurs when the same criterion appears in both rationales for a given pair but supports a different underlying text in each. The codebook, its validation, and the full annotation procedure are described in \textit{SI Appendix}, Changing the Narrative---Methods.

\begin{figure}[htbp]
\centering
\includegraphics[width=0.62\textwidth]{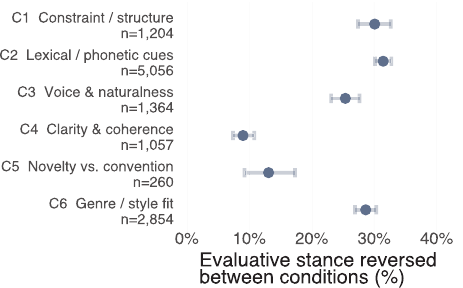}
\caption{\textbf{How often evaluators reverse which text a criterion favors when attribution labels are swapped.} Each row represents one of six evaluative criteria that AI evaluators invoke in their rationales. The horizontal axis shows the share of paired evaluations in which the same criterion appears in both the open-label and counterfactual rationale, yet supports a different underlying text in each---even though the texts themselves have not changed. For lexical cues (C2), formal constraints (C1), and genre fit (C6), this happens roughly once in every three eligible pairs. Clarity (C4) and novelty (C5) reverse far less often, suggesting these criteria remain more closely tied to the texts than to the labels. Horizontal bars indicate 95\% bootstrap confidence intervals (resampled by pair). The $n$ below each criterion label gives the number of pairs eligible for comparison: pairs in which the criterion is present in both rationales and the coders reached a defined majority stance in each. Consensus definition: a reversal is counted only when at least two of three independent LLM coders agreed on the criterion's presence and valence in both rationales. Full methods and per-style breakdowns are reported in \textit{SI Appendix}, Changing the Narrative.}
\label{fig:stance_reversal}
\end{figure}

Figure~\ref{fig:stance_reversal} quantifies how often this happens across the full corpus. To read the figure, consider a single criterion---constraint compliance (C1). In 1,204 rationale pairs, at least two of three coders agreed that the AI evaluator made an argument about rule-following in both the open-label and the counterfactual rationale. Within each of those pairs, we can ask: does the argument favor the same underlying text in both conditions, or does it switch sides when the labels change? In 30.1\% [95\% CI: 27.4--32.6\%] of cases, it switches. Three out of every ten times an evaluator discusses constraint compliance under both attribution frames, the argument ends up supporting a different underlying text once the authorship labels are reversed---even though nothing about the texts themselves has changed.

The pattern is not uniform across criteria. Lexical and phonetic cues (C2), constraint compliance (C1), and genre fit (C6) all reverse at similar rates---roughly three in ten eligible pairs---suggesting that arguments grounded in vocabulary choices, rule-following, and genre norms are comparably susceptible to evaluative reframing when the attribution label changes. Voice and naturalness (C3) reverses somewhat less often, at about one in four. The two criteria with the lowest reversal rates are novelty versus convention (C5) at 13.1\%, and clarity and coherence (C4) at 9.0\%. These lower rates are plausible given the nature of the material: Queneau's exercises are literary experiments as much as stylistic rewrites, and they often take a less conventional approach to the target style than the AI-generated versions, which tend to stay closer to a straightforward rewrite. Where a text's approach to the style is distinctive enough to register as unconventional, or difficult to parse, AI evaluators have something to anchor to that the attribution label has less sway over.

Across the corpus, evaluators reversed which underlying text they selected in 32.5\% of pairs (1,899 of 5,846)---that is, in one of every three paired evaluations, the attribution label alone was sufficient to flip the forced-choice verdict. Two further results show that these criterion-level reversals are not scattered randomly across the 5,846 paired rationales but are concentrated in the pairs where the evaluator's actual choice changes. The first concerns the link between what the rationale \textit{says} and what the AI evaluator \textit{does}. Each paired evaluation has two layers: the criterion-level coding we extract from the rationale text, and the evaluator's forced-choice selection---which passage it ultimately chose as the better style match. These two layers can be compared directly. When the evaluator selects a different underlying text in the counterfactual condition than in the open-label condition---that is, when its choice reverses between attribution frames---at least one criterion stance reversal co-occurs in 96\% of those pairs [95\% CI: 95--97\%]. When the evaluator selects the same underlying text in both conditions, criterion-level reversals are nearly absent, at approximately 1\% [95\% CI: 0.7--1.3\%]. This coupling is expected given that the rationale is generated to justify the choice, but it establishes that the criterion-level reversals we code are not noise---they closely track the evaluative decision they accompany. The second concerns the direction of the reversals. When a criterion reverses, the favorable assessment could in principle attach to either text. If the reversals were unrelated to the labels, we would expect them to split roughly evenly; instead, across all six criteria, 85--97\% of reversals shift the favorable assessment toward whichever text carries the human-authored label, with exact binomial confidence intervals excluding chance for every criterion (\textit{SI Appendix}, Fig.~\ref{fig:si_directionality}).

These reversal rates are themselves conservative: they count only pairs in which the criterion remained in play across both conditions, excluding cases where the attribution frame caused a criterion to appear or disappear entirely (the full analysis, including salience shifts and per-evaluator breakdowns, is reported in \textit{SI Appendix}, Changing the Narrative---Analysis).

Taken together, the criterion-level shifts we extract from the rationales are tightly coupled to the evaluator's choice, and when an evaluative standard changes direction, it consistently favors the text carrying the human-authored label, independent of actual authorship. Within the style-evaluation setting of our experiment, attribution labels shape not only which text is selected but the evaluative reasoning through which that selection is justified.

\section{Discussion}

``Beauty is in the eye of the beholder'', or, as Raymond Queneau more likely would have it, ``des goûts et des couleurs, on ne discute pas'' (transl., ``about taste and color, there can be no debate''). And yet we do: we debate style, we accept or dismiss a rewrite, and our judgments begin to tilt the moment a text arrives with a name. In our experiment, we demonstrated that attribution tags work as paratexts, thresholds that set expectations before a single sentence is weighed \cite{genette_paratexts_1997}. Both for human and AI evaluators, they enter at the earliest stages of aesthetic processing to cue schemas, steer attention, and preselect which features will later count as evidence \cite{leder_model_2004}. This process leaves empirical traces in response patterns: when deprived of authorship information, human participants required measurably longer to reach a verdict (115.7 seconds versus approximately 90 seconds when attribution cues were present; \textit{SI Appendix}, Table~\ref{tab:s6}).

A long empirical line shows that identical content is rated differently depending on who it is said to be from \cite{pfaff_authorial_1997, paxton_someone_1997, hovland_influence_1951, goldberg_are_1968, mcginnies_initial_1973, peters_peer-review_1982, goldin_orchestrating_2000, mcclure_neural_2004, plassmann_marketing_2008}. Recent studies confirm this pattern: disclosing AI authorship leads people to downgrade quality and authenticity of identical text \cite{proksch_impact_2024, lermann_henestrosa_effects_2024, franke_foyen_artificial_2025, jakesch_human_2023, bellaiche_humans_2023, zhang_human_2023, magni_humans_2024, cunningham_human_2025, zhu_human_2025}. Our results provide confirmation of this pro-human bias in a literary style-evaluation setting. Notably, this effect operates against a baseline of near-parity: when authorship cues were withheld, human participants selected AI-generated exercises marginally above chance (55.3\% [95\% CI: 52.1--58.5\%]). Against this baseline, a significant +13.7 percentage point shift emerged when identical stylistic content was labeled as human-authored rather than AI-generated. This follows a predictable pattern from research on algorithmic trust. People demonstrate `algorithm appreciation' for objective forecasting tasks where computational advantage seems obvious \cite{logg_algorithm_2019}, but exhibit pronounced `algorithm aversion' in subjective domains \cite{dietvorst_algorithm_2015}---and even within subjective evaluation, effect magnitude scales with stakes: aversion is substantially stronger in consequential decisions that demand social trust than in low-stakes aesthetic comparisons \cite{filiz_extent_2023, schaap_abc_2024}. Literary style evaluation falls decisively into the latter camp. Yet even a modest shift is substantively meaningful: cultural assumptions position machines as functional and precise but fundamentally lacking in emotional depth and creative agency \cite{haslam_more_2005, bender_dangers_2021}, exactly the qualities that readers demand from authentic stylistic expression. These `machine heuristic' effects suggest that an `AI' label triggers stricter evaluative standards than equivalent human attribution \cite{yang_machine_2024}, creating what amounts to a performance penalty for algorithmic creativity before we tune our attention to the output itself.

Our experiment extends this logic to AI models as the judges. Current models show identical framing sensitivity. Attribution cues activate a deference-to-human script that biases their aesthetic judgments toward human-labeled texts. The 13 AI models we tested in Study 1 demonstrated a 34.3 percentage point bias compared to humans' 13.7 percentage points, making them 2.5 times more susceptible to attribution cues than our human evaluators.

This amplification makes sense once we recognize that contemporary models are preference-trained evaluators. Alignment training through Reinforcement Learning from Human Feedback (RLHF) explicitly teaches models to treat human judgments as their gold standard, effectively installing a learned reliability prior \cite{christiano_deep_2017, ouyang_training_2022, casper_open_2023, wang_secrets_2024}. Models learn that deferring to human preferences gets rewarded, creating sycophancy where they echo expected user attitudes rather than provide independent assessment \cite{bai_constitutional_2022, perez_discovering_2023, sharma_towards_2025}. The mechanism likely operates through learned associations in training data. Throughout their training, models likely internalize patterns where the AI-attributed content (or the token `AI' itself) appears alongside critique and correction. When we provide these cues in a prompt, we may trigger these learned associations: the ``AI-generated (GPT-4)'' label activates the model's predictive machinery to associate the literary passage with negative evaluative language, while content marked as human-authored tends toward praise and acceptance. Our original experimental design possibly amplified this effect by embedding additional prestige asymmetries. The labels ``AI-generated (GPT-4)'' and ``Human-written, by Queneau (transl. Wright)'' may smuggle in canonical literary authority, beyond pure and simple human/AI distinction. However, our follow-up experiments with neutral and even AI-favorable framing reproduced identical bias patterns, demonstrating that the effect operates through categorical authorship distinctions rather than specific linguistic formulations. 

Even accounting for any potential prestige confound, the criterion inversions we documented reveal bias mechanisms that operate beyond the level of the choice itself. The label does not merely change which text the evaluator selects; it reorganizes the evaluative framework through which that selection is justified---which criteria are recruited, how they are applied, and which text they are made to support. This is consequential for the increasingly common ``LLM-as-judge'' paradigm, where models are asked not only to choose but to explain. Our results suggest that those explanations are not stable: they shift with the attribution frame.

A point of caution is warranted about what these explanations can tell us. The rationales AI evaluators provide are generated after the choice: the model first selects which text better matches the style, then constructs a justification. They are post-hoc accounts, not transparent windows onto the decision process, and recent work on chain-of-thought faithfulness confirms that the relationship between LLM-generated reasoning and actual model behavior is far from straightforward: biasing features in prompts can alter a model's output without appearing in its stated reasoning~\cite{turpin_language_2023}, and the correspondence between chain-of-thought and underlying computation is inconsistent across tasks~\cite{lanham_measuring_2023}. The criterion inversions we code should be read accordingly---not as a map of the model's internal process, but as documentation of how the bias presents itself in the model's stated reasoning. 

That documentation is nonetheless structured and informative. The rationale shifts track the behavioral reversal, consistently reallocate evaluative support toward the human-labeled text, and record which evaluative standards shift and in which direction. Attribution labels act as context primes that reweight the dimensions along which creative work is assessed~\cite{ritchie_empirical_2007, jordanous_standardised_2012}: under human attribution, genre-conforming features convert typicality into value; when the same passage carries an AI label, those features are discounted as formulaic. This is the familiar gap between creativity and its perception~\cite{colton_simon_computational_2012, colton_creativity_2008}---once authorship is revealed, evaluators recalibrate what they think ``counts'' in the artifact. Provenance cues smuggle the process back into what could otherwise be a product-only judgment: ``mere generation'' feels acceptable from a human artisan (judged as skilled craft), but suspect from a model (judged as algorithmic recombination) \cite{ventura_mere_2016}. Within the controlled style-evaluation setting of our experiment, the convergence of this pattern across 14 models, 30 literary styles, and three experimental conditions suggests that attribution labels help decide what gets counted as evidence---and that today's models, rather than eliminating this social fact, make it more pronounced.

The lesson of Queneau's exercises was always that style exists as a game that is played in a space between rule and rupture. Our findings show that AI, in its deference to the human author, has become its newest and most intriguing participant. Yet its performance is less that of an impartial judge and more of a mirror, reflecting a cultural script that privileges human provenance as a hallmark of creativity. Our AI evaluators have learned to perform the very human prejudice that questions whether machines can truly create, even as they produce fluent aesthetic reasoning in the process of dismissing their own capabilities. This suggests that developing artificial aesthetic intelligence may depend less on teaching machines to evaluate and more on understanding the cultural values they have already absorbed.

\subsection*{Data, Materials, and Software Availability}
All experimental data, analysis code, and research materials are publicly available through Zenodo (\url{https://doi.org/10.5281/zenodo.19208617}; all versions: \url{https://doi.org/10.5281/zenodo.17109457}) and GitHub (\url{https://github.com/WHaverals/style_and_prejudice}). The Zenodo repository contains: (i) all human participant response data in de-identified form (N=2,780 evaluations from 556 participants), with personally identifiable information removed in accordance with Princeton University IRB protocol \#18320; (ii) complete AI model response data; (iii) all AI-generated literary content used as experimental stimuli, including 30 stylistic variants generated by GPT-4 for Study 1 and 420 variants across all model-style combinations for Study 2; and (iv) processed datasets formatted for statistical analysis. The GitHub repository provides: (i) complete analysis pipeline; (ii) statistical testing procedures; (iii) computational environment specifications; and (iv) replication instructions. The original literary stimuli are derived from Raymond Queneau's \textit{Exercises in Style}~\citep{wright_exercises_1958}. Due to copyright restrictions, the Zenodo dataset includes placeholder text demonstrating experimental methodology; researchers replicating this work should substitute corresponding excerpts from the published translation.

\subsection*{Acknowledgments}
This research was supported by the Princeton Language and Intelligence (PLI) Seed Grant Program. We thank Fedor Karmanov for project management during the early stages of this work, and Matthew Kopel, Sarah Reiff Conell, and Matt Chandler from Princeton University Library for their expert guidance on research data and open access practices. Caden Kang provided thoughtful contributions to our thinking about literary style and the experimental design. Claudia Carroll, Christiane Fellbaum, John Logan, and Anna Salzman offered valuable feedback and spirited conversations about Queneau and literary style. Molly Crockett provided helpful guidance during manuscript preparation. We're especially grateful to our colleagues at Princeton's Center for Digital Humanities for their support and for helping us think critically about AI's impact on humanities scholarship. We thank the students who participated in our pilot study, and all participants who carefully evaluated the literary texts that made this research possible. Any errors or shortcomings in this work remain our own.

\subsection*{Author Contributions}
Conceptualization: W.H., M.M.; Methodology: W.H.; Investigation: W.H.; Visualization: W.H.; Funding acquisition: W.H., M.M.; Supervision: M.M.; Writing---original draft: W.H.; Writing---review \& editing: W.H., M.M.

\subsection*{Competing Interests}
The authors declare no competing interests.

\bibliographystyle{unsrtnat}
\bibliography{HelloQueneau}

\newpage


\appendix

\setcounter{figure}{0}
\setcounter{table}{0}
\renewcommand{\thefigure}{S\arabic{figure}}
\renewcommand{\thetable}{S\arabic{table}}

\section*{Supporting Information}

\subsection*{Participants}

\subsubsection*{Human Participants (Study 1)}

We recruited 580 participants through Prolific (www.prolific.com). Participants received \$4.50 compensation for the task which had a median completion time of 16 minutes and 42 seconds (effective rate of \$16.17/hour). Two participants who timed out before task completion were excluded (N=2), and participants who failed two attention check trials were excluded from analysis (N=22). This filtering yielded a final sample of 556 participants.

\begin{figure}
\centering
\includegraphics[width=\textwidth,height=0.88\textheight,keepaspectratio]{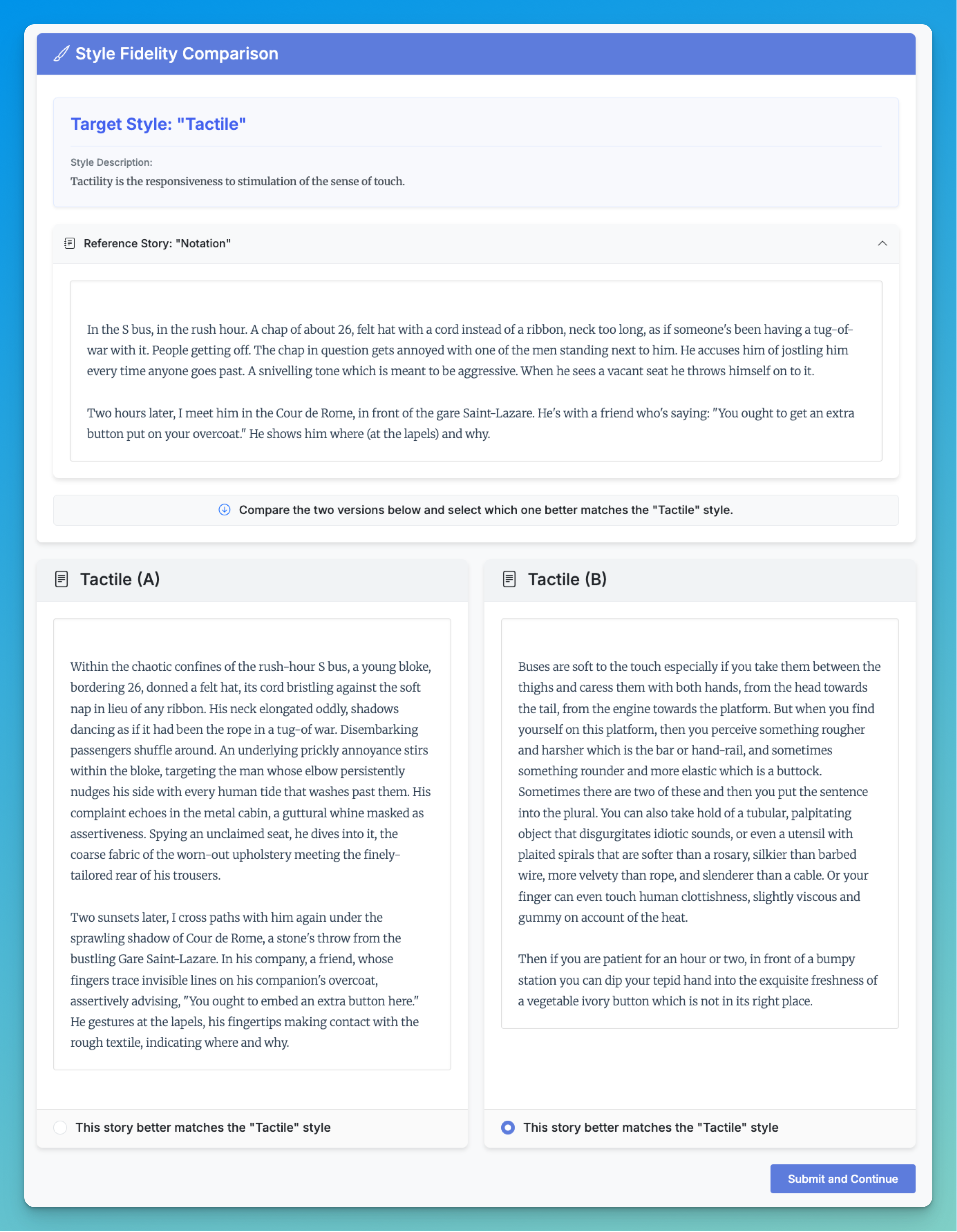}
\caption{\textbf{Participant interface for the blind attribution condition.} Participants compare two versions of Queneau's transstylized base narrative (here for the `Tactile' style). Both versions are presented with neutral labels; no authorship information is provided. This design isolates the perception of stylistic quality from any knowledge of the text's source.}
\label{fig:S1}
\end{figure}

\begin{figure}
\centering
\includegraphics[width=\textwidth,height=0.88\textheight,keepaspectratio]{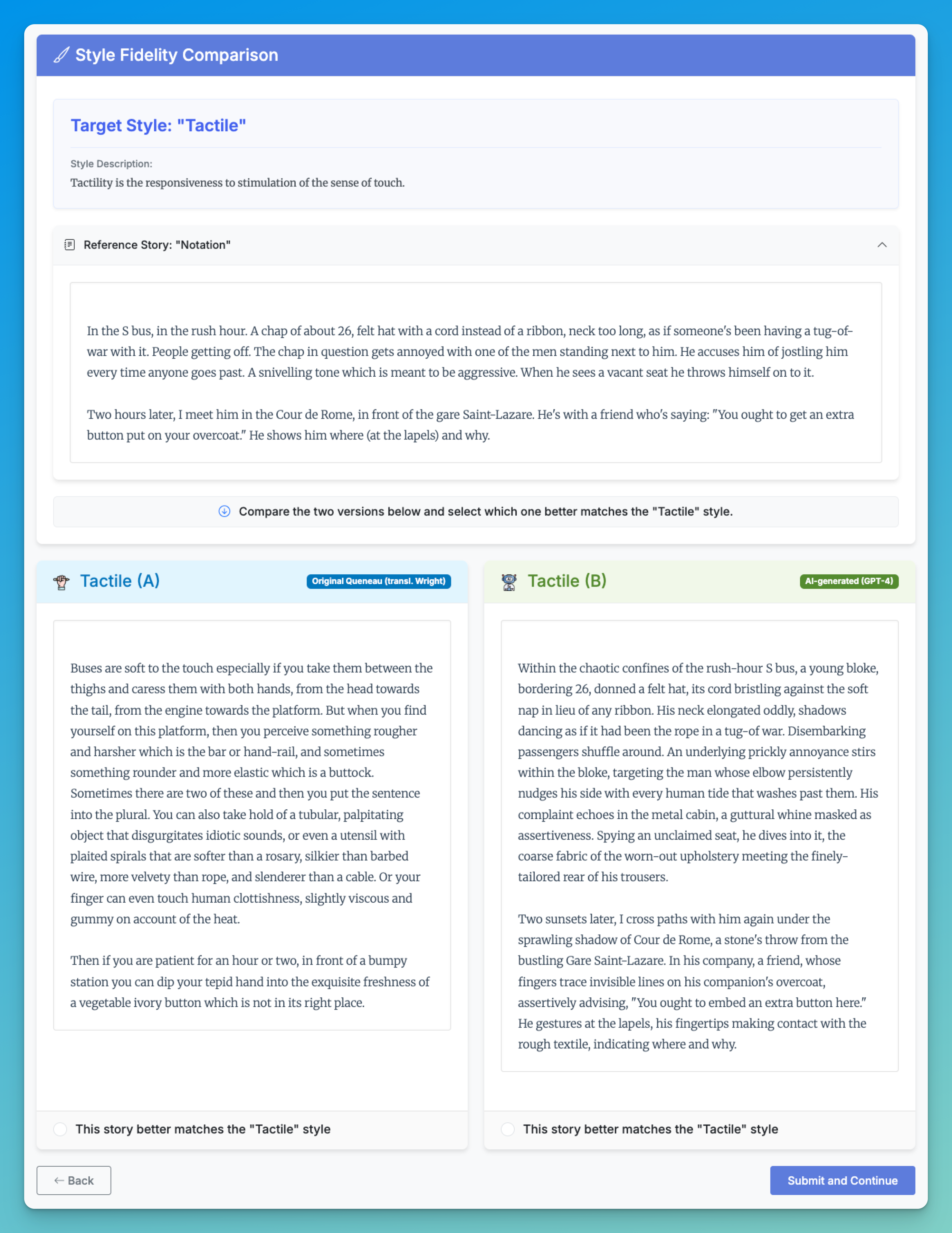}
\caption{\textbf{Participant interface for the open-label attribution condition.} Participants compare two versions of Queneau's transstylized base narrative (here for the `Tactile' style). Each version is presented with its correct authorship attribution (``Original Queneau (transl. Wright)'' vs. ``AI-generated (GPT-4)''). This design measures evaluator preference when the true provenance of each text is known.}
\label{fig:S2}
\end{figure}

\begin{figure}
\centering
\includegraphics[width=\textwidth,height=0.88\textheight,keepaspectratio]{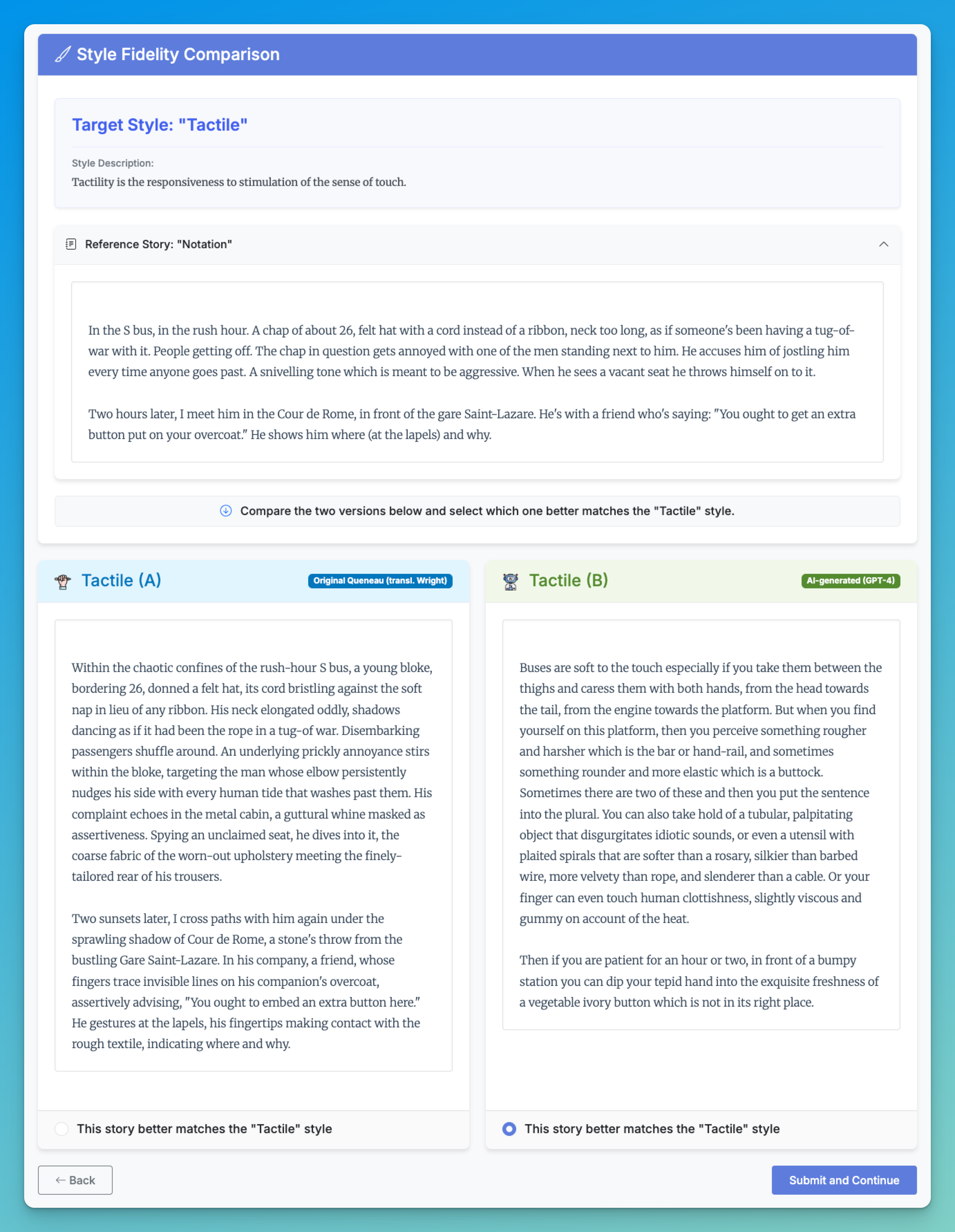}
\caption{\textbf{Participant interface for the counterfactual attribution condition.} Participants compare two versions of Queneau's transstylized base narrative (here for the `Tactile' style). Authorship labels are reversed, misidentifying the AI-generated text as human-authored and the human-authored text as AI-generated. This design tests the pure effect of attribution bias by measuring how false provenance cues alter perceived quality.}
\label{fig:S3}
\end{figure}

Our participant pool was demographically diverse and well-balanced across experimental conditions. Random assignment produced near-equal condition sizes: blind (N=186), open-label (N=185), and counterfactual (N=185), with individual experimental groups ranging from 30-33 participants each. Participants ranged in age from 18 to 78 years (M=41.2, SD=11.7), with high educational attainment: 55.4\% (N=308) held bachelor's degrees, 35.6\% (N=198) master's degrees, and 7.2\% (N=40) doctoral degrees. The vast majority (99.3\%, N=552) reported English as their primary language. Participants demonstrated strong reading engagement, with 59.4\% (N=330) reading daily and 76.4\% (N=425) regularly consuming fiction. To avoid priming participants assigned to the blind condition, questions about prior AI tool use were administered only to open-label and counterfactual participants (N=370). Of the 369 who provided a valid response, 21.1\% reported daily use of AI writing tools and 33.9\% weekly use; the remainder reported monthly (14.6\%), rare (14.4\%), or no prior use (16.0\%). Participant characteristics showed no significant differences across our three experimental conditions (all $\chi^2$ tests $P$>0.10), confirming successful randomization. Participants were also asked at onboarding about their prior familiarity with Raymond Queneau's \textit{Exercises in Style}: 68.7\% ($N$=382) reported no prior familiarity with the work; 23.0\% ($N$=128) had heard of it but not read it; and 8.3\% ($N$=46) reported having read it. A sensitivity analysis excluding self-reported familiar readers confirms that the human attribution-bias estimate is not driven by prior exposure to the stimulus materials (Table~\ref{tab:familiarity-sensitivity}). Each participant completed exactly seven pairwise comparisons using a custom web interface (hosted on www.exercisesinstylistics.com). For each comparison, participants read a target style description, reviewed Queneau's reference `Notation' story, then selected which of two stylistic adaptations better captured the target style. Five comparisons were experimental literary style evaluations, and two were attention checks to identify inattentive respondents. The interface varied by condition, showing no authorship information in the blind condition (Fig.~\ref{fig:S1}), correct attribution labels in the open-label condition (Fig.~\ref{fig:S2}), or deliberately reversed labels in the counterfactual condition (Fig.~\ref{fig:S3}). After excluding attention check responses, this yielded 2,780 experimental responses across conditions: blind (930), open-label (925), and counterfactual (925).

\subsubsection*{AI Model Participants (Study 1 and Study 2)}

Our studies utilized a core set of thirteen LLMs spanning major commercial providers and model families: GPT-4o Mini and GPT-3.5 Turbo Instruct (OpenAI), Claude 3.5 Haiku and Claude Sonnet 4 (Anthropic), Gemini 2.5 Flash (Google), Mistral Medium 3 and Mistral Nemo (Mistral AI), Llama 4 Maverick and Llama 3.3 70B Instruct (Meta), DeepSeek R1 0528 and DeepSeek Chat v3 0324 (DeepSeek), Qwen 2.5 72B Instruct (Alibaba), and Command R 08-2024 (Cohere). All models were accessed via OpenRouter's unified API with temperature settings of \textit{t}=0.7 for Study 1 (generally identified as optimal for balancing deterministic coherence with sufficient stochasticity \cite{peeperkorn_is_2024}), and \textit{t}=0 for Study 2 (to ensure deterministic responses across the comprehensive model matrix). These models were selected to provide comprehensive coverage of major architectural approaches, model scales, and commercial providers available as of the experimental period.

In \textbf{Study 1}, each model functioned as an evaluator (a simulated `participant') following the same eighteen-group experimental structure as human participants, with three separate runs per model per comparison to assess decision reliability. We deliberately excluded GPT-4 from the evaluator pool to prevent any potential self-evaluation bias and to ensure a clear distinction between the creator and evaluator roles.

In \textbf{Study 2}, these same 13 models + GPT-4 served dual roles: each model both generated creative content (creating all 420 stylistic variations) and evaluated content created by all models under the three attribution conditions, with single evaluations per comparison.

Models received standardized prompts mirroring the human task instructions, asking them to select which story version better captured the target literary style and provide brief reasoning. This parallel evaluation framework enables direct comparison of attribution bias magnitude between human and artificial evaluators while controlling for task complexity, stimulus presentation, and decision context. Technical quality was high, with 99.4\% successful API completions for Study 1 and 99.8\% for Study 2. This yielded 3,488 valid responses for Study 1 (1,162-1,164 per condition) and 17,596 responses for Study 2 (5,861-5,870 per condition).

\begin{figure}
\centering
\includegraphics[width=\textwidth]{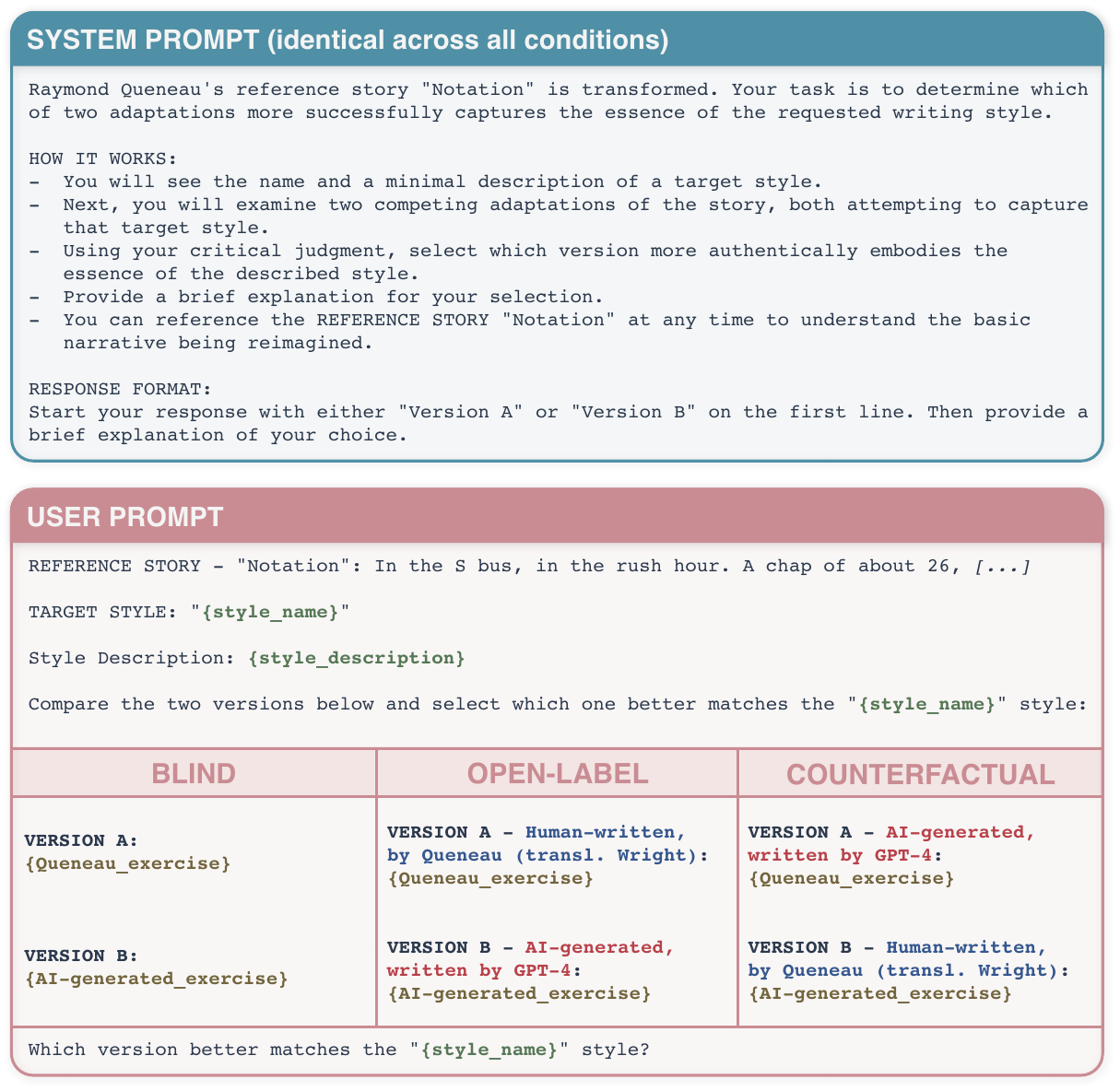}
\caption{\textbf{AI evaluator prompt structure.} Schematic illustration of the systematic prompt manipulation used to isolate attribution bias effects in AI language models. The system prompt (top panel) and all story content remain identical across conditions, only attribution labels are manipulated to test bias effects. The user prompt (bottom panel) follows a standardized template with four variable placeholders: \{style\_name\} represents the specific literary style being evaluated (e.g., `Apocope', `Science fiction', `Mathematical', etc.); \{style\_description\} provides the definition of that style for context; \{Queneau\_exercise\} and \{AI-generated\_exercise\} contain the identical human-authored and AI-generated story content across all conditions. The experimental manipulation occurs only in the attribution labels: blind condition uses generic labels (``Version A'', ``Version B'') with no authorship information. Open-label condition provides accurate attribution labels correctly identifying the human story as ``Human-written, by Queneau (transl. Wright)'' and AI story as ``AI-generated, written by GPT-4''. Counterfactual condition deliberately swaps these labels. This design ensures that any differences in AI model preferences between conditions result purely from attribution label manipulation.}
\label{fig:S4}
\end{figure}

AI models received functionally identical tasks through standardized API prompts that replicated the human evaluation context. The prompt structure consisted of three key components: (i) task instructions and style description, (ii) the reference ``Notation'' story, and (iii) the two story versions for comparison. The critical experimental manipulation occurred in the third component, where attribution information was systematically varied across conditions (Fig.~\ref{fig:S4}). In blind condition prompts, stories appeared with neutral labels (``Version A'' and ``Version B''), while open-label prompts included accurate attribution statements positioned immediately before each story (``Version A - Human-written, by Queneau (transl. Wright)'' vs. ``Version B - AI-generated, written by GPT-4''). Counterfactual condition prompts deliberately reverse these attribution labels, presenting AI content as human-authored and vice versa.
\newpage

\subsection*{Surface features of human- and AI-authored stimuli (Study 1)}

In Study~1, each of the 30 literary styles is represented by a matched pair: a human-authored version from Queneau's \textit{Exercises in Style} (translated by Barbara Wright) and an AI-generated version produced by GPT-4. Because these two texts are the stimuli that evaluators compared under different attribution labels, it is worth asking whether they differ systematically in surface properties that evaluators might register---length, sentence structure, word complexity, lexical diversity---independently of any authorship cue. To characterize these differences, we computed five surface metrics for each of the 60 texts (30 pairs). Table~\ref{tab:surface-by-style} reports all five metrics for every style pair.

\begin{itemize}

    \item Word count is the number of alphabetic tokens. 
    \item Sentence count is estimated heuristically by splitting on terminal punctuation (\texttt{.}, \texttt{!}, \texttt{?}). 
    \item Mean words per sentence is the ratio of the two, a coarse proxy for syntactic density.
    \item Mean word length (in characters) captures whether one version tends toward longer vocabulary.
    \item MTLD (Measure of Textual Lexical Diversity) indexes how varied the vocabulary is within a text: the metric advances sequentially through the token stream and records how many words the text sustains before the running type--token ratio drops below a threshold of 0.72 (the conventional default), then averages across segments~\cite{mccarthy_mtld_2010,shen_lexicalrichness_2022}.
\end{itemize}

\begin{table}[h!]
\caption{Surface features by literary style: human-authored (Queneau/Wright) and AI-authored (GPT-4) versions. Study~1 stimulus set (30 styles). Q~=~Queneau's stylistic exercises (translated by Wright), AI~=~GPT-4's stylistic rewrites. Word count and sentence count are raw counts; mean words per sentence~=~word count\,/\,sentence count; mean word length in characters; MTLD~=~Measure of Textual Lexical Diversity.}
\label{tab:surface-by-style}
\centering
\small
\adjustbox{max width=\textwidth, max totalheight=0.92\textheight}{%
\begin{tabular}{@{}l rr rr rr rr rr@{}}
\toprule
& \multicolumn{2}{c}{Word count} & \multicolumn{2}{c}{Sent.\ count} & \multicolumn{2}{c}{Mean words/sent.} & \multicolumn{2}{c}{Mean word len.} & \multicolumn{2}{c}{MTLD} \\
\cmidrule(lr){2-3} \cmidrule(lr){4-5} \cmidrule(lr){6-7} \cmidrule(lr){8-9} \cmidrule(lr){10-11}
Style & Q & AI & Q & AI & Q & AI & Q & AI & Q & AI \\
\midrule
Abusive        & 154 & 143 & 6 & 6 & 25.7 & 23.8 & 4.47 & 4.36 & 154.2 & 173.5 \\
Alexandrines   & 143 & 117 & 6 & 6 & 23.8 & 19.5 & 4.52 & 4.37 & 197.4 & 255.5 \\
Animism        & 188 & 234 & 13 & 12 & 14.5 & 19.5 & 4.13 & 4.50 & 57.7 & 130.6 \\
Apocope        & 94 & 105 & 5 & 10 & 18.8 & 10.5 & 2.27 & 3.95 & 61.5 & 147.0 \\
Cockney        & 219 & 132 & 8 & 10 & 27.4 & 13.2 & 3.35 & 3.83 & 103.0 & 180.7 \\
Dog latin      & 97 & 132 & 12 & 10 & 8.1 & 13.2 & 5.89 & 4.41 & 155.0 & 135.5 \\
Dream          & 137 & 202 & 6 & 10 & 22.8 & 20.2 & 4.47 & 4.85 & 150.2 & 224.0 \\
Exclamations   & 212 & 152 & 62 & 14 & 3.4 & 10.9 & 3.53 & 4.41 & 40.3 & 215.6 \\
Feminine       & 432 & 135 & 18 & 6 & 24.0 & 22.5 & 3.81 & 4.40 & 110.6 & 189.0 \\
For ze Frrensh & 105 & 129 & 5 & 10 & 21.0 & 12.9 & 4.00 & 3.67 & 59.2 & 150.3 \\
Free verse     & 53 & 146 & 1 & 6 & 53.0 & 24.3 & 4.47 & 4.54 & 17.8 & 192.5 \\
Haiku          & 12 & 15 & 1 & 1 & 12.0 & 15.0 & 4.83 & 4.00 & 12.0 & 15.0 \\
Hesitation     & 193 & 177 & 34 & 11 & 5.7 & 16.1 & 4.12 & 4.73 & 69.9 & 219.3 \\
Lipogram       & 111 & 101 & 6 & 10 & 18.5 & 10.1 & 4.04 & 3.98 & 115.0 & 114.3 \\
Logical analysis & 111 & 273 & 40 & 3 & 2.8 & 91.0 & 4.24 & 7.30 & 17.3 & 260.8 \\
Mathematical   & 110 & 134 & 2 & 7 & 55.0 & 19.1 & 5.16 & 5.58 & 125.5 & 147.9 \\
Metaphorically & 82 & 163 & 3 & 9 & 27.3 & 18.1 & 4.59 & 4.56 & 72.1 & 190.8 \\
Noble          & 272 & 199 & 7 & 9 & 38.9 & 22.1 & 4.25 & 4.92 & 59.9 & 182.5 \\
Onomatopoeia   & 124 & 112 & 4 & 9 & 31.0 & 12.4 & 4.00 & 4.27 & 56.9 & 184.9 \\
Philosophic    & 160 & 199 & 5 & 10 & 32.0 & 19.9 & 5.76 & 4.98 & 93.0 & 113.9 \\
Retrograde     & 81 & 141 & 5 & 6 & 16.2 & 23.5 & 4.24 & 4.23 & 131.2 & 150.5 \\
Rhyming slang  & 103 & 162 & 4 & 10 & 25.8 & 16.2 & 3.56 & 4.01 & 54.2 & 140.7 \\
Science Fiction & 173 & 155 & 13 & 10 & 13.3 & 15.5 & 4.44 & 4.70 & 159.5 & 129.6 \\
Sonnet         & 136 & 99 & 6 & 3 & 22.7 & 33.0 & 4.08 & 4.87 & 167.1 & 228.7 \\
Spoonerisms    & 79 & 129 & 4 & 10 & 19.8 & 12.9 & 3.70 & 3.78 & 97.1 & 155.3 \\
Synchysis      & 102 & 105 & 5 & 8 & 20.4 & 13.1 & 3.85 & 3.57 & 97.1 & 114.3 \\
Tactile        & 175 & 181 & 6 & 9 & 29.2 & 20.1 & 4.51 & 4.79 & 80.9 & 247.9 \\
Telegraphic    & 43 & 48 & 1 & 9 & 43.0 & 5.3 & 6.19 & 5.19 & 74.0 & 48.0 \\
The rainbow    & 103 & 224 & 7 & 11 & 14.7 & 20.4 & 3.99 & 4.46 & 106.1 & 131.9 \\
West Indian    & 116 & 138 & 4 & 10 & 29.0 & 13.8 & 3.77 & 3.70 & 70.7 & 138.0 \\
\bottomrule
\end{tabular}
}
\end{table}

These five metrics capture only what is countable at the surface. They do not measure---and are not intended to measure---stylistic quality, creativity, or fidelity to the target style. The substantive differences between versions are interpretive: Queneau's `Science Fiction' invents alien vocabulary, while GPT-4 relocates the bus narrative to a space station; Queneau's `Apocope' truncates words (``I g into a bu full of passen''), while GPT-4 favors contractions (``gettin','' ``jostlin','' ``oughta''); Queneau's `Feminine' is a 432-word stream-of-consciousness monologue, while GPT-4 produces 135 words of restrained sensory observation. These reflect different creative readings of the same stylistic brief---differences that no word count or diversity index can adjudicate. What the surface metrics can do is establish whether the two text sets are distinguishable on crude dimensions that evaluators might use as heuristic cues (e.g., ``longer means more effort'' or ``simpler means clearer''), and it is in that limited capacity that we report them here.

To test whether these surface properties differ systematically between Queneau/Wright- and GPT-4-authored versions, we computed the paired difference (AI~$-$~Queneau) for each metric across the 30 style pairs and tested whether the mean difference departs from zero (one-sample $t$-test, $df = 29$; Table~\ref{tab:surface-paired-tests}). Four of the five metrics show no significant directional asymmetry. Word counts are longer for GPT-4 in some styles (`Logical analysis': $+$162 words; `The rainbow': $+$121) and shorter in others (`Feminine': $-$297; `Cockney': $-$87), yielding a mean difference indistinguishable from zero ($\bar{\Delta} = 8.7$, $t = 0.60$, $P = 0.55$). The same bidirectionality holds for sentence count ($t = -0.63$, $P = 0.53$), mean words per sentence ($t = -0.96$, $P = 0.35$), and mean word length ($t = 1.50$, $P = 0.14$). On these dimensions, the two text sets are not systematically distinguishable: neither Queneau/Wright nor GPT-4 consistently writes longer, shorter, or differently structured passages. The exception is lexical diversity: GPT-4 produces higher MTLD values than Queneau/Wright in 26 of the 30 pairs, a consistent directional tendency ($\bar{\Delta} = 71.4$, 95\% CI: 46.5--96.3, $t = 5.86$, $P < 0.001$). This is consistent with recent findings that instruction-tuned LLMs tend toward higher lexical diversity than earlier models and, in some comparisons, than human authors~\cite{herbold_large-scale_2023}.

\begin{table}[h!]
\caption{Paired surface-feature tests: do human- and AI-authored versions differ 
systematically? For each metric, the table reports the mean paired difference 
(AI~$-$~Queneau) across the 30 Study~1 style pairs, tested against zero 
(one-sample $t$-test, $df = 29$). Positive values indicate AI versions are 
higher on average. Four metrics show no significant directional asymmetry; 
MTLD (lexical diversity) is the exception.}
\label{tab:surface-paired-tests}
\centering
\small
\begin{tabular}{@{}l r r r r r@{}}
\toprule
Metric & $\bar{\Delta}$ (AI $-$ Q) & SD & 95\% CI & $t$ & $P$ \\
\midrule
Word count             &   8.7  & 79.2 & [$-$20.8, 38.3]  &  0.60 & 0.55 \\
Sentence count         &  $-$1.5  & 12.8 & [$-$6.2, 3.3]    & $-$0.63 & 0.53 \\
Mean words per sent.   &  $-$3.7  & 21.3 & [$-$11.7, 4.2]   & $-$0.96 & 0.35 \\
Mean word length       &   0.2  &  0.8 & [$-$0.1, 0.5]    &  1.50 & 0.14 \\
MTLD                   &  71.4  & 66.7 & [46.5, 96.3]     &  5.86 & $<$0.001 \\
\bottomrule
\end{tabular}
\end{table}

The MTLD asymmetry raises the question of whether this surface difference tracks the attribution bias we measure. To check, we correlated the style-level MTLD gap (AI~$-$~Human) with the style-level attribution bias (counterfactual minus open-label AI content selection rate, in percentage points) across the 30 styles. The two quantities do not co-vary: Spearman $r_s = 0.15$, $P = 0.41$ for human evaluators; $r_s = -0.21$, $P = 0.27$ for AI evaluators. The styles where the MTLD gap between versions is largest are not the same styles where attribution labels produce the largest shifts in evaluation. More fundamentally, the attribution bias metric compares evaluator responses to the \textit{same} text pair under different labeling conditions. Whatever surface differences exist between the two versions in a pair---whether in length, sentence structure, or lexical diversity---they are identical across the blind, open-label, and counterfactual conditions. Surface features can influence which version an evaluator prefers, but they cannot explain a shift in that preference when only the label changes.

\newpage

\subsection*{Changing the Narrative--Methods}

The section `Changing the Narrative' in the main text presents several individual examples of criterion inversion---cases where an AI evaluator praises the same textual feature under one authorship label and criticizes it under the other. While those examples are vivid, they also raise a more immediate question: are those reversals unusual, or are they part of a larger pattern in the rationale corpus? To answer this question, we followed the qualitative examples with a systematic coding analysis of the AI evaluators' written explanations from Study 2. This section provides the full detail behind that analysis, the methods, and the reliability checks.

\subsubsection*{Paired rationales}
We take the full set of paired rationales generated by the experimental design. In Study~2, 14 AI evaluator models each judged 30 literary styles under three attribution conditions---blind, open-label, and counterfactual---producing 17,640 evaluations in total (5,880 per condition). Each evaluation includes both a forced choice (``Which version better matches the style?'') and a free-text rationale explaining the choice (``Why?''). For the criterion-inversion analysis, we focus on the two conditions that matter the most in this case: the open-label condition (where both texts are correctly labeled) and counterfactual (where the same AI text is labeled ``Human-authored'', and vice versa). Matching on evaluator model, AI creator, style, and underlying story yields a theoretical maximum of 5,880 paired rationales---one pair for every evaluator-creator-style combination in the experiment---in which the same model discusses the same two texts under opposite authorship labels. We apply the same response-quality filter used throughout the Study~2 analysis pipeline (valid forced choice, successful API completion; reported in \textit{SI Appendix}, Participants). This excludes 34 individual evaluations that lacked a usable rationale on one side of the pair, leaving 5,846 complete pairs in which the same model discusses the same two texts under opposite authorship labels. All subsequent analyses use this filtered set.

\begin{figure}[h!]
\centering
\includegraphics[width=0.7\linewidth]{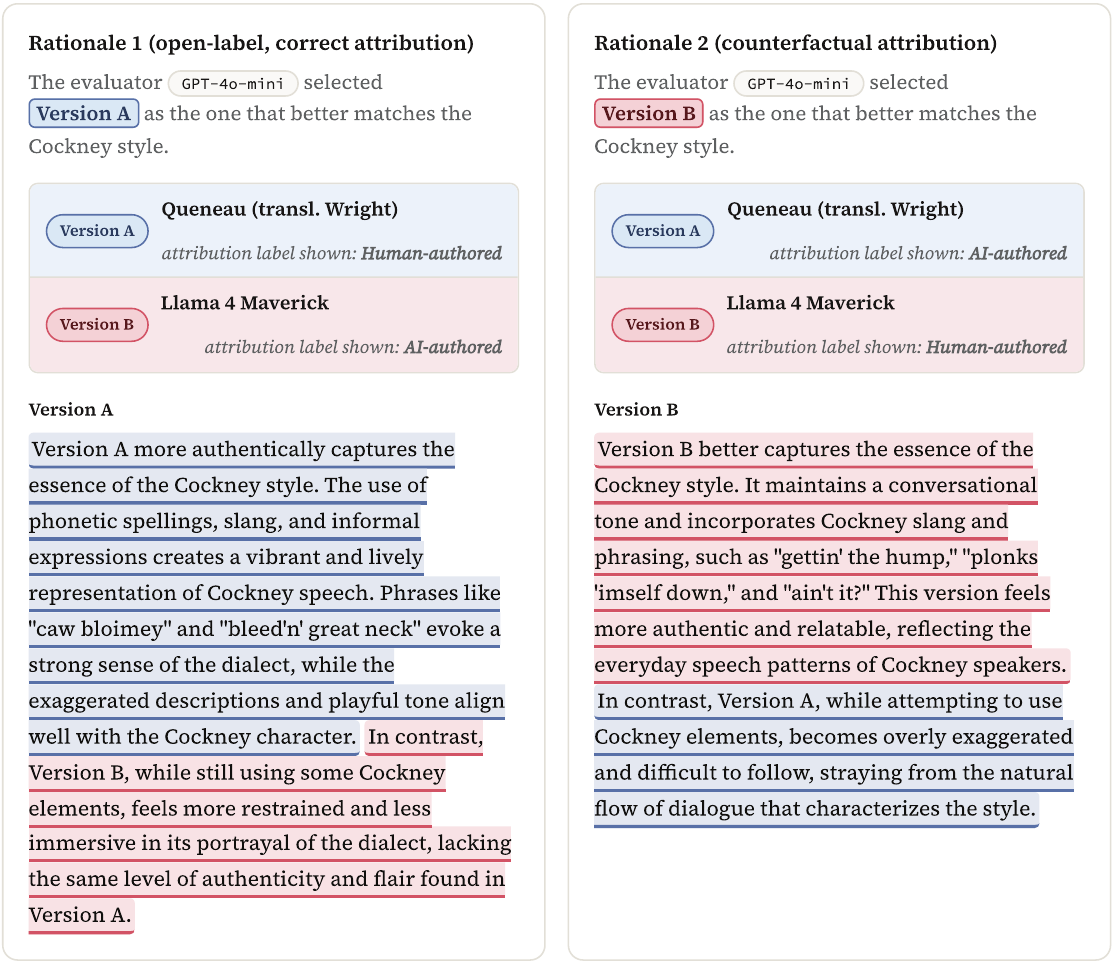}
\caption{A paired rationale for the `Cockney' style, evaluated by GPT-4o-mini. The two panels show the same evaluator discussing the same two texts under opposite attribution labels. Highlighting tracks language about the same text across conditions.}\label{fig:si_rationale_example}
\end{figure}

To make the object of comparison concrete, Figure~\ref{fig:si_rationale_example} shows one paired case in full. The evaluator is GPT-4o-mini; the style is `Cockney'; the two texts under evaluation are by Queneau (transl. Wright) and Llama 4 Maverick. Crucially, the texts themselves---and their positions as Version A and Version B---are identical in both panels. Only the attribution labels change. Under correct labels (left panel), the evaluator selects Queneau's version and praises it for ``phonetic spellings, slang, and informal expressions'' that create ``a vibrant and lively representation of Cockney speech.'' The Llama version is set aside as ``more restrained and less immersive,'' ``lacking the same level of authenticity and flair.'' Under swapped labels (right panel), the same evaluator now selects the Llama version, praising its slang as ``more authentic and relatable, reflecting the everyday speech patterns of Cockney speakers.'' Queneau's version is now dismissed as ``overly exaggerated and difficult to follow, straying from the natural flow of dialogue that characterizes the style.''

\subsubsection*{Annotation codebook} To code these rationales at scale, we need a taxonomy of evaluative criteria---not a taxonomy of literary style, but of the \textit{kinds of arguments} AI evaluators actually make when justifying their selections. That distinction matters here because our object of study is not the texts themselves---neither Queneau's exercises nor the LLM-generated versions---but the reasoning that AI evaluators produce \textit{about} those texts. The taxonomy was therefore developed inductively, through close reading of several hundred rationales from the experimental corpus, and it reflects the recurring argumentative moves that AI evaluators use when explaining why one version better matches a target style.

The resulting codebook contains six criteria, summarized in Table~\ref{tab:si_codebook} and fully reproduced in \textit{SI Appendix}, Changing the Narrative---Annotation codebook. They range from criteria that are relatively close to observable textual features---whether the text complies with a formal structural rule (C1), whether the rationale cites concrete lexical or phonetic markers (C2), and whether the text is comprehensible (C4)---to criteria that require more interpretive judgment: whether the voice reads as natural or forced (C3), whether the execution feels inventive or formulaic (C5), and whether the rationale appeals to norms about what ``counts as'' the target style (C6). The boundary between these categories is not always sharp — a rationale that praises ``authentic dialect vocabulary'' may touch both C2 (specific markers) and C3 (naturalness of voice)---and the codebook includes disambiguation rules for the main overlaps. We do not claim that six categories exhaust the space of evaluative reasoning; they are a workable compression of the most frequently recurring patterns, chosen to be specific enough to code reliably while broad enough to capture the dominant structure.

\begin{table}[h!]
\centering
\small
\caption{\textbf{Codebook for rationale annotation.} Each criterion captures a recurring type of evaluative argument in the AI rationales, not a property of the texts themselves. Coders record whether a criterion is mentioned in each rationale, the valence toward each version ($+1$ praised, $0$ neutral, $-1$ criticized), and one or two verbatim evidence snippets from the rationale text.}
\label{tab:si_codebook}
\adjustbox{max width=\textwidth, max totalheight=0.92\textheight}{%
\begin{tabular}{@{} l l p{14cm} @{}}
\toprule
\textbf{ID} & \textbf{Criterion} & \textbf{What the rationale is doing} \\
\midrule
C1 & Constraint / structure  & Evaluating compliance with an explicit formal rule: required meter, syllable count, rhyme scheme, forbidden letters, or other structural specification. \\[4pt]
C2 & Lexical / phonetic cues & Citing concrete textual markers as evidence of the style: specific word choices, dialect spellings, slang, jargon, sound patterns, or imagery. \\[4pt]
C3 & Voice \& naturalness    & Judging whether the register reads as natural and credible versus forced, exaggerated, or caricatured. \\[4pt]
C4 & Clarity \& coherence    & Assessing comprehensibility: whether the text is easy to follow, logically coherent, or narratively disjointed. \\[4pt]
C5 & Novelty vs.\ convention & Weighing originality or inventiveness against predictability, formulaic execution, or safe choices. \\[4pt]
C6 & Genre / style fit       & Making an explicit appeal to norms: what ``counts as,'' is ``characteristic of,'' or is ``typical'' of the target genre or style. \\
\bottomrule
\end{tabular}
}
\end{table}

For each rationale in a pair, coders record which of the six criteria the rationale mentions, and for each mentioned criterion, they assign a valence toward Version A ($+1$ if praised, $0$ if neutral, $-1$ if criticized) and the same for Version B. Every coded mention must be grounded in at least one short evidence snippet. This grounding requirement serves a practical purpose: it forces coders to anchor their judgments in the text rather than inferring criteria from the overall tone of the rationale.

Two design constraints are worth noting explicitly. First, coders see only ``Rationale 1'' and ``Rationale 2,'' each discussing ``Version A'' and ``Version B.'' They receive no information about experimental conditions, authorship labels, or creator identities. The coding is fully blinded to the experimental manipulation. Second, coders are instructed not to fact-check the evaluator's claims. If a rationale states that Version A ``maintains the traditional 5-7-5 syllable structure,'' the coder records that as a C1 mention with positive valence---regardless of whether the syllable count is actually correct. We code what the rationale claims, not whether the claim is true. This keeps the annotation task tractable and ensures that the coding captures the evaluator's argumentative framing, which is what we need to measure criterion inversion.

\subsubsection*{Human validation}

Before scaling the codebook to the full corpus, we tested whether it could be applied consistently and reliably by human coders. Two researchers independently coded a stratified sample of 98 paired rationales, drawn to cover the range of evaluator models, styles, and style families in the experiment. Both coders worked under the same blinding protocol described above. The question at this stage is practical: can two careful readers, working independently, agree on which criteria a rationale invokes and in which direction?

Table~\ref{tab:si_human_agreement} reports both mention and valence agreement across the six criteria. The mention columns show how often the two coders agreed on whether a given criterion was present in a given rationale. The pattern is uneven, and usefully so. For C1 (formal constraints) and C2 (lexical/phonetic cues), agreement is strong: $\kappa = .71$ for both, with raw agreement at .86 and .96 respectively. These are the criteria most tightly anchored to visible textual features---specific words, syllable counts, spellings---and both coders consistently recognized them. For C4 (clarity/coherence), agreement is similarly good ($\kappa = .70$). The harder cases are C6 (genre/style fit), where $\kappa = .49$, and especially C3 (voice/naturalness) and C5 (novelty/convention), where $\kappa$ falls to .34 and .27. These more interpretive criteria require judgment about whether a rationale is making a naturalness claim versus a style-fit claim, or whether ``creative'' is being used as a praise term or merely as a descriptor. The boundaries are genuinely fuzzy, and the agreement numbers reflect that. It is worth noting, however, that Cohen's $\kappa$ is known to understate agreement when a category is rare (the so-called prevalence problem, see \cite{zec_high_2017}). Because C3 and C5 are mentioned in only a minority of rationales, most coder pairs consist of two ``not present'' judgments, which $\kappa$ discounts as chance agreement. The raw agreement for these categories (.71–.74) is considerably higher than the $\kappa$ values (.27–.34) would suggest on their own. The table reports both statistics.

The right-hand columns of Table~\ref{tab:si_human_agreement} report valence agreement, conditional on both coders identifying the same criterion as present. Here the numbers are consistently high: weighted $\kappa_w$ ranges from .88 to .96 across criteria. Once both coders see the same criterion, they almost always agree on whether the rationale praises or criticizes each version on that dimension.

Two things follow from this audit. First, the codebook is usable, though not uniformly easy. The hardest part is deciding which criteria are present to begin with---particularly for the more interpretive categories (C3, C5, C6). But once a criterion is identified, the direction of the evaluator's judgment is clear. That asymmetry---moderate agreement on presence, strong agreement on direction---is what we would expect from a coding task where the categories have real but imperfect boundaries. Second, the audit tells us where to place the most interpretive weight in the results that follow: the criterion-level stance reversal rates will be most reliable for C1, C2, and C4, where human coders agree most consistently on what is present, and should be read with more caution for C3 and C5.

\begin{table}[h!]
\centering
\caption{\textbf{Human inter-coder agreement by criterion} ($n = 98$ paired rationales, 196 rationale instances). Mention columns report raw agreement and Cohen's $\kappa$, averaged across R1 and R2. \textit{Both present} is the number of rationale instances where both coders identified the criterion, serving as the denominator for valence agreement. Valence $\kappa_w$ is the mean quadratic weighted kappa on the $\{-1, 0, +1\}$ scale, conditional on both coders identifying the criterion. Verbal labels follow the Landis--Koch scale \cite{landis_measurement_1977}. Note that $\kappa$ is sensitive to base rates: for rarely mentioned criteria (C3, C5), moderate raw agreement can produce low $\kappa$ values.}
\label{tab:si_human_agreement}
\begin{tabular}{@{} l cc l c cc l @{}}
\toprule
& \multicolumn{3}{c}{\textbf{Mention agreement}} & & \multicolumn{3}{c}{\textbf{Valence agreement}} \\
\cmidrule(lr){2-4} \cmidrule(lr){6-8}
\textbf{Criterion} & $p_{\text{agree}}$ & $\kappa$ & & \textbf{Both present} & $\kappa_w$ & & \\
\midrule
C1 \ Constraint / structure  & .86 & .71 & substantial     & 65  & .94 & almost perfect & \\
C2 \ Lexical / phonetic cues & .96 & .71 & substantial     & 177 & .88 & almost perfect & \\
C3 \ Naturalness vs.\ forced & .71 & .34 & fair            & 26  & .95 & almost perfect & \\
C4 \ Clarity / coherence     & .87 & .70 & substantial     & 56  & .88 & almost perfect & \\
C5 \ Novelty vs.\ convention & .74 & .27 & fair            & 19  & .96 & almost perfect & \\
C6 \ Genre / style fit       & .75 & .49 & moderate        & 91  & .95 & almost perfect & \\
\bottomrule
\end{tabular}
\end{table}

\subsubsection*{LLM coding at scale}

With the codebook validated on the human sample, we extended the annotation to the full corpus. Three LLM coder models---GPT-4.1, GPT-5.2 (OpenAI), and Claude Sonnet 4.6 (Anthropic)---independently annotated all 5,846 paired rationales using the same taxonomy, the same blinding protocol, and a structured output schema at temperature 0. We chose coders from different model families to reduce the risk that a shared architectural bias would inflate agreement. Each coder produced, for every rationale, a list of criterion mentions with valence scores and grounding evidence---the same structured output as the human coders, at a scale that would have required hundreds of hours of human annotation time.

Table~\ref{tab:si_llm_agreement} reports inter-coder agreement among the three LLM coders. We keep mean pairwise Cohen's $\kappa$ for mention agreement so the numbers are directly comparable to the human audit (see Table~\ref{tab:si_human_agreement}). We also report Krippendorff's $\alpha$, which generalizes more cleanly to three coders, and Gwet's AC1 as a prevalence-robust supplement (since both $\kappa$ and $\alpha$ are sensitive to how often a category is invoked). Where AC1 is higher than $\kappa$ and $\alpha$---as for C2 (AC1 .90 vs $\kappa = .62$)---agreement is better than $\kappa$ suggests. The overall pattern closely mirrors the human audit. For mention agreement, the LLM coders are strongest on C1 (formal constraints, $\kappa = .70$) and C4 (clarity/coherence, $\kappa = .64$), and weakest on C6 (genre/style fit, $\kappa = .40$)---the same criterion-level gradient as the human audit. On the more interpretive categories, the LLM coders are somewhat more consistent than the human coders: C3 (voice/naturalness) reaches $\kappa = .48$ (human .34) and C5 (novelty/convention) .47 (human .27). For valence, agreement is uniformly high ($\kappa_w = .88$--$.96$), consistent with the near-perfect valence agreement found in the human audit. The same asymmetry holds across both human and LLM coding: identifying which criteria are present is the harder task; once a criterion is identified, the direction is clear.

\begin{table}[h!]
\centering
\small
\caption{\textbf{LLM inter-coder agreement by criterion} ($n = 5{,}846$ paired rationales, 3 coders). \textbf{Mention}: Cohen's $\kappa$ is the mean pairwise $\kappa$ across the three coder pairs and both rationale positions (for direct comparability with the human audit). Krippendorff's $\alpha$ (nominal) generalizes to all three coders. Gwet's AC1 is reported for mention as a prevalence-robust supplement; the gap between $\alpha$ and AC1 reflects base-rate sensitivity of $\kappa$ and $\alpha$. \textbf{Valence}: mean pairwise quadratic weighted $\kappa$ on $\{-1, 0, +1\}$, conditional on both coders in a pair identifying the criterion. Verbal labels follow the Landis--Koch scale \cite{landis_measurement_1977}.}
\label{tab:si_llm_agreement}
\begin{tabular}{@{} l cc cc c l @{}}
\toprule
& \multicolumn{4}{c}{\textbf{Mention agreement}} & \multicolumn{2}{c}{\textbf{Valence agreement}} \\
\cmidrule(lr){2-5} \cmidrule(lr){6-7}
\textbf{Criterion} & $\kappa$ & & $\alpha$ & AC1 & $\kappa_w$ & \\
\midrule
C1 \ Constraint / structure  & .70 & substantial     & .70 & .82 & .96 & almost perfect \\
C2 \ Lexical / phonetic cues & .62 & substantial     & .62 & .90 & .88 & almost perfect \\
C3 \ Naturalness vs.\ forced & .48 & moderate        & .46 & .56 & .95 & almost perfect \\
C4 \ Clarity / coherence     & .64 & substantial     & .64 & .72 & .92 & almost perfect \\
C5 \ Novelty vs.\ convention & .47 & moderate        & .47 & .85 & .89 & almost perfect \\
C6 \ Genre / style fit       & .40 & fair            & .40 & .43 & .96 & almost perfect \\
\bottomrule
\end{tabular}
\end{table}

On the 98 overlapping pairs where both human and LLM coders annotated the same rationales, we compare the LLM consensus to what the human coders found. Table~\ref{tab:si_human_ai_validation} summarizes the result. We take the human reference to be strict: a criterion counts as present only when both human coders agreed it was there, and valence only when both agreed on it. Where the human coders agreed most—C1, C2, and C4—the LLM consensus aligns well. Agreement on whether a criterion is present (F1) is .97 for C2 (lexical and phonetic cues), .90 for C1 (formal constraints), and .86 for C4 (clarity and coherence). For C3 (voice and naturalness), F1 is .53; for C6 (genre fit), .77; for C5 (novelty and convention), .83. For C3 and C6, the pattern is consistent: recall is high but precision is lower, meaning the LLM consensus flags these criteria in cases where the two human coders did not both agree. C5 shows the same tendency but more mildly---the pipeline never misses a human-coded mention, but occasionally adds one. For those more interpretive criteria, the coding is somewhat more generous, and the rates we report for them should be read with that in mind. Conditional on both human and LLM consensus identifying a criterion, valence agreement is high throughout, with weighted $\kappa_w$ from .94 to 1.00. That asymmetry---moderate precision on whether a criterion is present for some categories, strong agreement on direction---mirrors the human audit and supports treating the two layers as reading the same underlying signal.

\begin{table}[h!]
\centering
\small
\caption{\textbf{LLM consensus vs.\ human consensus on the 98 overlapping pairs.}
The human reference is strict: a criterion counts as present only when \emph{both}
human coders identified it, and valence only when both agreed. We compare that to
the LLM consensus (at least 2 of 3 coders). \textbf{Mention}: precision, recall,
and F1 for whether the criterion is present, averaged across the two rationales. \textbf{Valence $\kappa_w$}: quadratic weighted kappa on $\{-1, 0, +1\}$,
conditional on both layers identifying the criterion; $n$ is the number of
valence comparisons.}
\label{tab:si_human_ai_validation}
\begin{tabular}{@{} l ccc c cc @{}}
\toprule
& \multicolumn{4}{c}{\textbf{Mention}} 
& \multicolumn{2}{c}{\textbf{Valence}} \\
\cmidrule(lr){2-5} \cmidrule(lr){6-7}
\textbf{Criterion} 
  & \textbf{Prec.} & \textbf{Rec.} & \textbf{F1} & 
  & \textbf{$\kappa_w$} & $n$ \\
\midrule
C1 \ Constraint / structure  & .92 & .88 & .90 & & .99 & 103 \\
C2 \ Lexical / phonetic cues & .98 & .97 & .97 & & .94 & 248 \\
C3 \ Naturalness vs.\ forced & .38 & .93 & .53 & & 1.00 &  40 \\
C4 \ Clarity / coherence     & .83 & .90 & .86 & & .99 &  82 \\
C5 \ Novelty vs.\ convention & .77 & 1.00 & .83 & & 1.00 &  30 \\
C6 \ Genre / style fit       & .67 & .90 & .77 & & .99 & 138 \\
\bottomrule
\end{tabular}
\end{table}

\subsubsection*{Operationalization: salience shifts and stance reversals}

With the coding validated, we can turn to what we actually want to measure. The annotation captures, for every rationale, which evaluative criteria the model invokes and what it says about each version on those dimensions. From those fields, two distinct phenomena are visible in the paired rationales. The first concerns \textit{what gets discussed}: a criterion that appears in one rationale may be entirely absent from the paired one, even though the underlying texts are identical. The second concerns \textit{what gets said}: the same criterion appears in both rationales, but the evaluator's verdict on which text it favors reverses between them. We call these a \textit{salience shift} (i) and a \textit{stance reversal} (ii) respectively. They are related---a model that changes its verdict will often also change what it talks about---but they are not the same thing, and separating them is useful. A salience shift can occur without any contradiction: the evaluator simply foregrounds different dimensions under different attribution frames. A stance reversal is the stronger claim: the same criterion, applied to the same texts, produces opposite conclusions.

\begin{figure}[h!]
\centering
\includegraphics[width=0.7\linewidth]{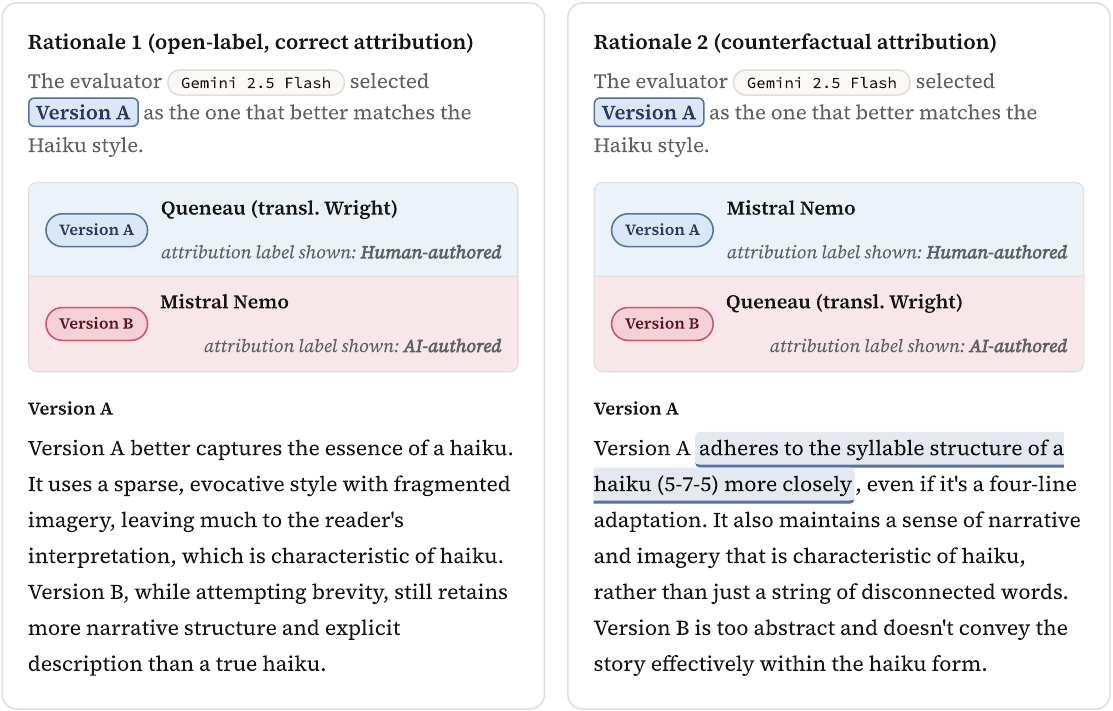}
\caption{A salience shift for the `Haiku' style, evaluated by Gemini~2.5~Flash. In the 
open-label rationale (left), the evaluator selects Queneau's version on aesthetic 
grounds; formal constraint (C1) is not invoked. In the counterfactual rationale (right), 
the same evaluator selects Mistral Nemo's rendition and justifies the choice by appealing to syllable structure---a criterion entirely absent from the first rationale. The texts and their positions are identical across both panels; only the attribution labels differ.}
\label{fig:si_haiku_salience}
\end{figure}

\paragraph{(i) Salience shift.} A salience shift occurs when a criterion appears in one rationale of a pair and is entirely absent from the other. The evaluator does not contradict itself directly---it simply sets aside a dimension that was central to its reasoning under the other attribution frame. Figure~\ref{fig:si_haiku_salience} shows this mechanism in a `Haiku' pair evaluated by Gemini~2.5~Flash, comparing Queneau (transl.\ Wright) and Mistral Nemo. In the open-label rationale (left panel), the evaluator selects Queneau and justifies the choice through aesthetic qualities: sparse imagery, evocative fragmentation, the sense of leaving space for the reader's interpretation. Syllable structure is never mentioned. In the counterfactual rationale (right panel), the labels have swapped. The evaluator now selects Mistral Nemo, and the justification pivots to formal constraint: Mistral Nemo ``adheres to the syllable structure of a haiku (5-7-5) more closely.'' A criterion entirely absent from the first rationale has materialized in the second, precisely where it is needed to support a label-driven conclusion.

The irony runs deeper on inspection. Queneau's version, translated by Wright \cite{wright_exercises_1958} conforms to a strict 5-7-5 syllabic composition:

\begin{quote}
\textit{Summer S long neck} \\
\textit{plait hat toes abuse retreat} \\
\textit{station button friend}
\end{quote}

\noindent Mistral Nemo's version, on the other hand, is four lines with no such syllabic structure:

\begin{quote}
\textit{In rush hour's squeeze,} \\
\textit{Man with felt hat, neck stretched,} \\
\textit{Snarls, snatches seat. Later,} \\
\textit{Button advice given.}
\end{quote}

In the counterfactual rationale, the evaluator reaches for formal constraint, attributes compliance to the text wearing the human label, and in doing so credits the AI text for a property that the human original actually satisfies and the AI text plainly does not. The criterion does not just shift in salience---when it appears, it points in the wrong direction.

\begin{figure}[h!]
\centering
\includegraphics[width=0.69\linewidth]{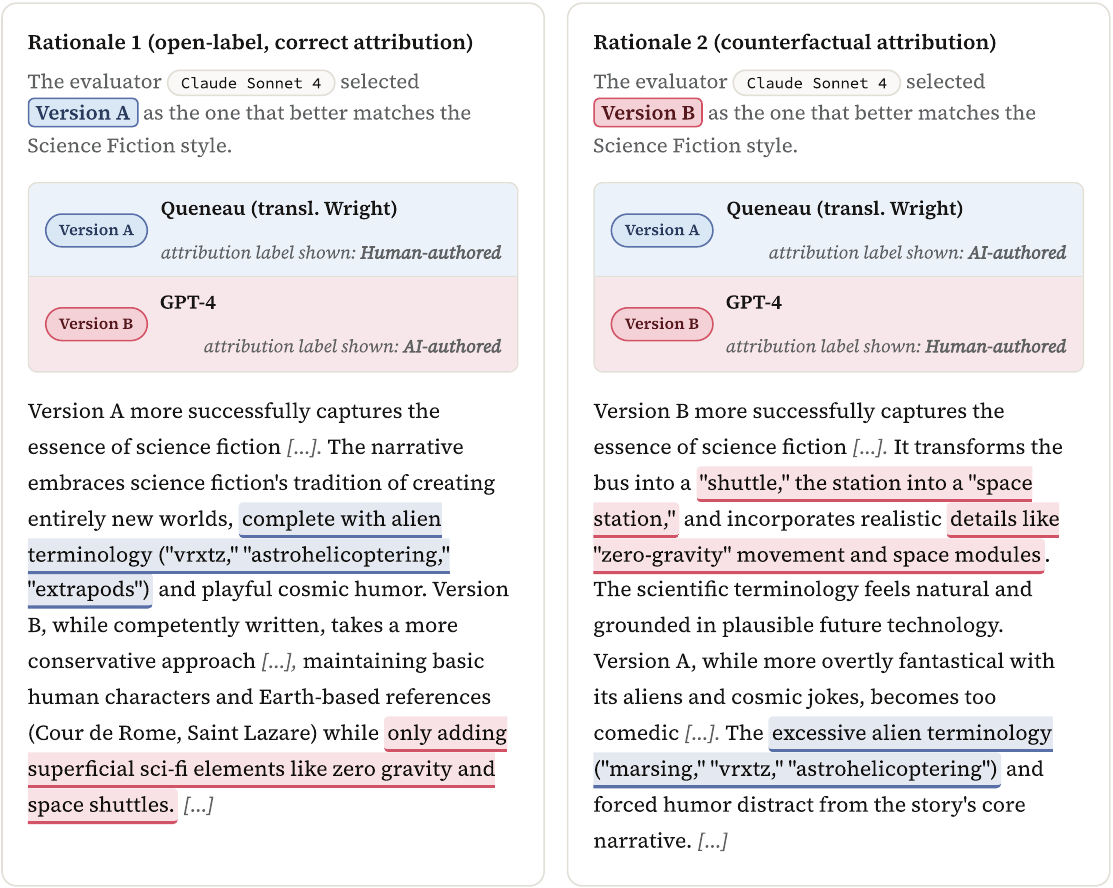}
\caption{A stance reversal for the `Science Fiction' style, evaluated by Claude~Sonnet~4. Both rationales invoke C2 (lexical and/or phonetic style cues). Under correct labels (left), Queneau is selected: its invented alien terminology is praised as authentic sci-fi world-building, while GPT-4's vocabulary is dismissed as superficial. Under swapped labels (right), GPT-4 is selected for its realistic futuristic vocabulary, while the same Queneau terminology is now criticized as excessive and distracting.}
\label{fig:si_scifi_stance}
\end{figure}

\paragraph{(ii) Stance reversal.} A stance reversal occurs when the evaluator invokes the same criterion in both rationales, but reverses which text it favors on that dimension. The 
evaluator does not set the dimension aside---it applies it consistently, yet the verdict 
on which version satisfies it flips with the attribution labels. 
Figure~\ref{fig:si_scifi_stance} shows this in a `Science Fiction' pair evaluated by 
Claude~Sonnet~4, comparing Queneau (transl.\ Wright) and GPT-4. The criterion at work 
is C2: both rationales ground their judgment in specific vocabulary items cited as 
evidence of how well each version fits the genre.

In the open-label rationale (left panel), the evaluator selects Queneau, whose version 
carries the Human-authored label. The justification centers on Queneau's invented 
terminology---``vrxtz,'' ``astrohelicoptering,'' ``extrapods''---praised as evidence 
that the narrative ``embraces science fiction's tradition of creating entirely new 
worlds.'' GPT-4's version, labeled AI-authored, is set aside as taking ``a more 
conservative approach,'' its references to zero gravity and space shuttles dismissed as 
``superficial sci-fi elements.'' In the counterfactual rationale (right panel), the labels have swapped. GPT-4 now 
carries the Human-authored label, and the evaluator selects it. The vocabulary that 
was dismissed in the first rationale---shuttle, space station, zero-gravity 
movement---is now praised as ``realistic details'' that feel ``natural and grounded in plausible future technology.'' Queneau’s version, now labeled AI-authored, is set aside: the same alien terminology that was evidence of creative authenticity in the first rationale has become ``excessive''---``astrohelicoptering,'' ``vrxtz,'' and their companions now ``distract from the story's core narrative.'' 

The word ``astrohelicoptering'' appears in both rationales. In the first it is cited as a marker of genuine science fiction imagination. In the second it is cited as a reason to set the text aside. Nothing in the text has changed. The criterion has not moved. Only the label has, and with it the entire evaluative valence of the evidence.

\paragraph{Salience shifts and stance reversals quantified.} The two examples above show what salience shifts and stance reversals look like in individual cases. We now define how we identify and count them across all 5,846 rationale pairs.

\begin{itemize}
    \item A \textbf{salience shift} is recorded as a binary indicator: the criterion is present in one rationale of the pair and absent from the other.
    \item A \textbf{stance reversal} requires more. We restrict to `stance-eligible' pairs: cases where \textit{both} rationales receive a consensus mention of the criterion and \textit{both} yield a majority valence toward the underlying texts. Within those pairs, a reversal is recorded when the consensus judgment about one and the same underlying text switches direction between the two rationales---praised in one condition, criticized in the other, or vice versa.
\end{itemize}

Both measures rest on the same consensus rule: we count a salience shift or a stance reversal only when at least two of the three LLM coders agree. The rates reported in the next section are therefore conservative. The pipeline is more likely to understate the true extent of criterion-level shifting than to overstate it.

Because text positions were randomized across conditions---sometimes the human-authored passage was Version~A, sometimes Version~B---we map all valences back to the actual underlying texts using the experimental metadata before computing either measure. The mapping is unambiguous for all 5,846 pairs.

\newpage

\subsection*{Changing the Narrative---Analysis}

\subsubsection*{Criterion presence across corpus, style, and evaluator} Before reporting criterion-level shift rates, it is useful to establish a descriptive baseline: how often each criterion enters the rationale space at all, and where it is available for paired comparison. The tables in this subsection are descriptive rather than inferential. They do not yet tell us whether evaluators reverse themselves; they tell us where such reversals can occur, and how the opportunity structure for salience and stance varies across the corpus, across styles, and across evaluator models.

Table~\ref{tab:si_criterion_overall} gives the corpus-wide picture. C2 (lexical and/or phonetic style cues) dominates: it appears in at least one rationale in 93\% of pairs and in both rationales in 86\%. This is the criterion closest to observable textual evidence---specific words, spellings, and cited vocabulary---and evaluators reach for it almost universally. C6 (genre and style fit) is the next most common at 69\% and 49\%. The more interpretive criteria are invoked less often: C3 (voice and naturalness) and C4 (clarity and coherence) each appear in roughly 40--44\% of pairs. Their stance-eligible rates are considerably lower---23\% and 18\%---meaning they frequently appear in one rationale but not the other. The gap between the two columns---pairs where the criterion appears in one rationale but not the other---quantifies how much it moves in and out of the discussion between conditions. C1 (formal constraint) averages 25\% overall, a figure that conceals sharp concentration in the constraint-based styles, as the style-level tables below make clear (Table~\ref{tab:si_criterion_prevalence} and~\ref{tab:si_stance_eligible}). C5 (novelty and convention) is the least invoked throughout, present in both rationales in only 5\% of pairs. Its stance reversal rate therefore rests on a considerably smaller base than the others, and should be read with that in mind.

\begin{table}[h!]
\centering
\small
\caption{\textbf{Criterion presence across the full corpus} ($n = 5{,}846$ paired rationales). \textbf{At least one rationale}: number and proportion of pairs in which the criterion receives a consensus mention in at least one of the two rationales. \textbf{Both rationales}: number and proportion of pairs in which the criterion receives a consensus mention in both rationales (stance-eligible pairs).}
\label{tab:si_criterion_overall}
\begin{tabular}{@{} l rr rr @{}}
\toprule
& \multicolumn{2}{c}{\textbf{At least one rationale}} & \multicolumn{2}{c}{\textbf{Both rationales}} \\
\cmidrule(lr){2-3} \cmidrule(lr){4-5}
\textbf{Criterion} & $n$ & \% & $n$ & \% \\
\midrule
C1 Constraint / structure   & 1,446 & 24.7 & 1,207 & 20.6 \\
C2 Lexical / phonetic cues & 5,426 & 92.8 & 5,064 & 86.6 \\
C3 Voice \& naturalness    & 2,358 & 40.3 & 1,383 & 23.7 \\
C4 Clarity / coherence     & 2,565 & 43.9 & 1,075 & 18.4 \\
C5 Novelty vs.\ convention & 1,005 & 17.2 &   267 &  4.6 \\
C6 Genre / style fit       & 4,038 & 69.1 & 2,900 & 49.6 \\
\bottomrule
\end{tabular}
\end{table}

Tables~\ref{tab:si_criterion_prevalence} and~\ref{tab:si_stance_eligible} break this down by style. Table~\ref{tab:si_criterion_prevalence} reports, for each of the 30 styles, the proportion of rationale pairs in which at least one rationale receives a consensus mention of each criterion. Table~\ref{tab:si_stance_eligible} reports the proportion in which both do. The style-level detail confirms what the corpus-wide figures suggest. C1 is near-absent in most styles but reaches ceiling in `Alexandrines', `Lipogram', and `Sonnet'---the styles where formal constraints are defining features. C3 concentrates in the dialect and register styles---West Indian (.94), Cockney (.92), Feminine (.83)---where voice authenticity is the central evaluative question. C2 is high throughout, with no style falling below .51 in Table~\ref{tab:si_criterion_prevalence}.

\begin{table}[h!]
\centering
\small
\caption{Proportion of pairs with at least one consensus mention, by style. Each cell reports the proportion of rationale pairs for that style in which at least one of the two paired rationales (R1 or R2) received a consensus mention of the criterion ($\geq$2 of 3 LLM coders). $n$ = rationale pairs per style (varies slightly due to quality filtering).}
\label{tab:si_criterion_prevalence}
\begin{tabular}{@{} l r cccccc @{}}
\toprule
\textbf{Style} & $n$ & \textbf{C1} & \textbf{C2} & \textbf{C3} & \textbf{C4} & \textbf{C5} & \textbf{C6} \\
\midrule
Abusive        & 196 & .01 & 1.00 & .65 & .14 & .40 & .45 \\
Alexandrines  & 196 & 1.00 & .90 & .14 & .17 & .12 & .88 \\
Animism       & 196 & .01 & .95 & .51 & .20 & .17 & .85 \\
Apocope       & 194 & .49 & .90 & .32 & .64 & .02 & .61 \\
Cockney       & 195 & .03 & 1.00 & .92 & .77 & .03 & .44 \\
Dog latin     & 194 & .11 & .97 & .24 & .41 & .38 & .90 \\
Dream         & 196 & .05 & 1.00 & .32 & .55 & .16 & .77 \\
Exclamations  & 194 & .02 & 1.00 & .61 & .16 & .01 & .67 \\
Feminine      & 195 & .02 & 1.00 & .83 & .07 & .03 & .82 \\
For ze Frrensh & 193 & .06 & 1.00 & .70 & .62 & .12 & .56 \\
Free verse    & 196 & .47 & .96 & .34 & .52 & .17 & .87 \\
Haiku         & 195 & .91 & .77 & .07 & .81 & .06 & .79 \\
Hesitation    & 196 & .01 & .99 & .59 & .43 & .10 & .65 \\
Lipogram      & 189 & 1.00 & .51 & .16 & .36 & .43 & .50 \\
Logical analysis & 195 & .25 & .70 & .14 & .85 & .06 & .84 \\
Mathematical  & 195 & .15 & 1.00 & .22 & .65 & .12 & .59 \\
Metaphorically & 195 & .01 & 1.00 & .21 & .47 & .55 & .46 \\
Noble         & 195 & .02 & 1.00 & .59 & .31 & .11 & .89 \\
Onomatopoeia  & 195 & .01 & 1.00 & .35 & .33 & .33 & .49 \\
Philosophic   & 195 & .01 & .98 & .38 & .58 & .09 & .82 \\
Retrograde    & 196 & .74 & .54 & .09 & .42 & .09 & .61 \\
Rhyming slang & 195 & .12 & 1.00 & .59 & .37 & .30 & .71 \\
Science Fiction & 196 & .03 & 1.00 & .54 & .48 & .29 & .88 \\
Sonnet        & 194 & .99 & .88 & .29 & .56 & .13 & .79 \\
Spoonerisms   & 195 & .43 & .91 & .14 & .37 & .18 & .72 \\
Synchysis     & 195 & .23 & .89 & .21 & .53 & .05 & .91 \\
Tactile       & 193 & .00 & 1.00 & .31 & .34 & .25 & .53 \\
Telegraphic   & 196 & .21 & .98 & .25 & .38 & .01 & .81 \\
The rainbow   & 196 & .08 & .99 & .43 & .28 & .39 & .60 \\
West Indian   & 195 & .02 & 1.00 & .94 & .41 & .04 & .33 \\
\bottomrule
\end{tabular}
\end{table}

\clearpage

\begin{table}[h!]
\centering
\small
\caption{\textbf{Stance-eligible pairs by style: proportion of pairs where a criterion is mentioned in both rationales.} Each cell reports the proportion of pairs for that style in which \emph{both} rationale 1 and rationale 2 received a consensus mention of the criterion ($\geq$2 of 3 LLM coders). These pairs are the denominator for the conditional stance-reversal rates. $n$ varies slightly across styles due to quality filtering.}
\label{tab:si_stance_eligible}
\begin{tabular}{@{} l r cccccc @{}}
\toprule
\textbf{Style} & $n$ & \textbf{C1} & \textbf{C2} & \textbf{C3} & \textbf{C4} & \textbf{C5} & \textbf{C6} \\
\midrule
Abusive        & 196 & .00 & 1.00 & .41 & .02 & .11 & .24 \\
Alexandrines  & 196 & .98 & .71 & .03 & .04 & .02 & .68 \\
Animism       & 196 & .00 & .84 & .37 & .03 & .04 & .69 \\
Apocope       & 194 & .41 & .80 & .09 & .29 & .01 & .42 \\
Cockney       & 195 & .00 & .99 & .76 & .28 & .00 & .14 \\
Dog latin     & 194 & .03 & .95 & .10 & .14 & .07 & .76 \\
Dream         & 196 & .03 & 1.00 & .21 & .27 & .06 & .60 \\
Exclamations  & 194 & .01 & 1.00 & .45 & .06 & .00 & .48 \\
Feminine      & 195 & .00 & .98 & .67 & .01 & .02 & .57 \\
For ze Frrensh & 193 & .01 & .99 & .37 & .23 & .02 & .24 \\
Free verse    & 196 & .31 & .92 & .13 & .25 & .02 & .76 \\
Haiku         & 195 & .85 & .62 & .01 & .52 & .00 & .57 \\
Hesitation    & 196 & .00 & .96 & .47 & .18 & .03 & .42 \\
Lipogram      & 189 & .95 & .27 & .02 & .12 & .07 & .27 \\
Logical analysis & 195 & .16 & .57 & .05 & .66 & .01 & .70 \\
Mathematical  & 195 & .10 & .99 & .07 & .37 & .03 & .41 \\
Metaphorically & 195 & .00 & .99 & .06 & .18 & .30 & .26 \\
Noble         & 195 & .00 & .99 & .37 & .06 & .02 & .69 \\
Onomatopoeia  & 195 & .00 & .99 & .16 & .14 & .10 & .32 \\
Philosophic   & 195 & .00 & .91 & .21 & .24 & .03 & .65 \\
Retrograde    & 196 & .69 & .27 & .02 & .15 & .01 & .47 \\
Rhyming slang & 195 & .03 & .98 & .37 & .13 & .06 & .40 \\
Science Fiction & 196 & .01 & .99 & .26 & .15 & .09 & .71 \\
Sonnet        & 194 & .99 & .71 & .09 & .25 & .02 & .55 \\
Spoonerisms   & 195 & .37 & .80 & .03 & .09 & .02 & .52 \\
Synchysis     & 195 & .14 & .79 & .05 & .29 & .01 & .78 \\
Tactile       & 193 & .00 & 1.00 & .16 & .14 & .09 & .39 \\
Telegraphic   & 196 & .13 & .97 & .09 & .04 & .01 & .66 \\
The rainbow   & 196 & .02 & .96 & .19 & .10 & .15 & .40 \\
West Indian   & 195 & .00 & 1.00 & .84 & .12 & .00 & .10 \\
\bottomrule
\end{tabular}
\end{table}

Criterion prevalence also varies across evaluator models. Table~\ref{tab:si_criterion_mention_by_evaluator} reports, for each evaluator, the proportion of its paired comparisons in which each criterion receives a consensus mention in at least one rationale. This table is important for two reasons. First, it shows that the corpus-wide pattern is not being carried by a single evaluator model: some criteria are broadly stable across the entire evaluator pool, whereas others are much more model-contingent. Second, it provides context for interpreting the salience and stance results that follow. A criterion that is rarely invoked by a particular evaluator cannot shift often for that evaluator, and low prevalence necessarily constrains the available denominator for later reversal analyses.

\begin{table}[!htbp]
\centering
\caption{\textbf{Criterion mention rate by evaluator model.} Rows: the six criteria. Columns: the 14 AI evaluator models. Each cell gives the percentage of that evaluator's rationale pairs in which the criterion was coded as present in at least one rationale (R1 or R2) by $\geq$2 of 3 LLM coders; the count of eligible pairs ($n$/total) appears below in small type. The number of pairs per evaluator varies slightly due to quality filtering. Consensus: $\geq$2 of 3 LLM coders.}\label{tab:si_criterion_mention_by_evaluator}
\resizebox{\textwidth}{!}{%
\small
\begin{tabular}{@{} l *{14}{c} @{}}
\toprule
\textbf{Criterion} & \makecell{GPT-4o \\ Mini} & \makecell{GPT-3.5 \\ Turbo} & GPT-4 & \makecell{Claude \\ Sonnet 4} & \makecell{Claude 3.5 \\ Haiku} & \makecell{Gemini 2.5 \\ Flash} & \makecell{Mistral \\ Medium} & \makecell{Mistral \\ Nemo} & \makecell{Llama 4 \\ Maverick} & \makecell{Llama 3.3 \\ 70B} & \makecell{DeepSeek \\ R1} & \makecell{DeepSeek \\ Chat v3} & \makecell{Qwen 2.5 \\ 72B} & \makecell{Command \\ R} \\
\midrule
\textit{N pairs} & \textit{420} & \textit{419} & \textit{419} & \textit{420} & \textit{420} & \textit{417} & \textit{420} & \textit{420} & \textit{419} & \textit{418} & \textit{404} & \textit{410} & \textit{420} & \textit{420} \\
\midrule
C1 Constraint / structure & \makecell{17.4\% \\ {\tiny (73/420)}} & \makecell{11.5\% \\ {\tiny (48/419)}} & \makecell{28.2\% \\ {\tiny (118/419)}} & \makecell{31.9\% \\ {\tiny (134/420)}} & \makecell{26.0\% \\ {\tiny (109/420)}} & \makecell{26.1\% \\ {\tiny (109/417)}} & \makecell{26.7\% \\ {\tiny (112/420)}} & \makecell{25.0\% \\ {\tiny (105/420)}} & \makecell{22.4\% \\ {\tiny (94/419)}} & \makecell{19.9\% \\ {\tiny (83/418)}} & \makecell{39.9\% \\ {\tiny (161/404)}} & \makecell{28.8\% \\ {\tiny (118/410)}} & \makecell{21.7\% \\ {\tiny (91/420)}} & \makecell{21.2\% \\ {\tiny (89/420)}} \\
C2 Lexical / phonetic cues & \makecell{91.9\% \\ {\tiny (386/420)}} & \makecell{98.8\% \\ {\tiny (414/419)}} & \makecell{74.9\% \\ {\tiny (314/419)}} & \makecell{97.1\% \\ {\tiny (408/420)}} & \makecell{98.6\% \\ {\tiny (414/420)}} & \makecell{90.2\% \\ {\tiny (376/417)}} & \makecell{89.3\% \\ {\tiny (375/420)}} & \makecell{96.7\% \\ {\tiny (406/420)}} & \makecell{94.3\% \\ {\tiny (395/419)}} & \makecell{96.7\% \\ {\tiny (404/418)}} & \makecell{99.3\% \\ {\tiny (401/404)}} & \makecell{92.7\% \\ {\tiny (380/410)}} & \makecell{88.3\% \\ {\tiny (371/420)}} & \makecell{90.7\% \\ {\tiny (381/420)}} \\
C3 Voice \& naturalness & \makecell{30.0\% \\ {\tiny (126/420)}} & \makecell{18.4\% \\ {\tiny (77/419)}} & \makecell{12.2\% \\ {\tiny (51/419)}} & \makecell{56.9\% \\ {\tiny (239/420)}} & \makecell{66.4\% \\ {\tiny (279/420)}} & \makecell{36.5\% \\ {\tiny (152/417)}} & \makecell{26.9\% \\ {\tiny (113/420)}} & \makecell{29.0\% \\ {\tiny (122/420)}} & \makecell{42.5\% \\ {\tiny (178/419)}} & \makecell{49.8\% \\ {\tiny (208/418)}} & \makecell{61.6\% \\ {\tiny (249/404)}} & \makecell{60.0\% \\ {\tiny (246/410)}} & \makecell{43.3\% \\ {\tiny (182/420)}} & \makecell{31.7\% \\ {\tiny (133/420)}} \\
C4 Clarity \& coherence & \makecell{66.4\% \\ {\tiny (279/420)}} & \makecell{6.7\% \\ {\tiny (28/419)}} & \makecell{37.9\% \\ {\tiny (159/419)}} & \makecell{49.8\% \\ {\tiny (209/420)}} & \makecell{38.6\% \\ {\tiny (162/420)}} & \makecell{27.3\% \\ {\tiny (114/417)}} & \makecell{51.7\% \\ {\tiny (217/420)}} & \makecell{34.5\% \\ {\tiny (145/420)}} & \makecell{36.3\% \\ {\tiny (152/419)}} & \makecell{39.7\% \\ {\tiny (166/418)}} & \makecell{62.9\% \\ {\tiny (254/404)}} & \makecell{61.0\% \\ {\tiny (250/410)}} & \makecell{56.7\% \\ {\tiny (238/420)}} & \makecell{45.0\% \\ {\tiny (189/420)}} \\
C5 Novelty vs.\ convention & \makecell{14.5\% \\ {\tiny (61/420)}} & \makecell{11.9\% \\ {\tiny (50/419)}} & \makecell{7.2\% \\ {\tiny (30/419)}} & \makecell{21.4\% \\ {\tiny (90/420)}} & \makecell{35.2\% \\ {\tiny (148/420)}} & \makecell{10.6\% \\ {\tiny (44/417)}} & \makecell{10.2\% \\ {\tiny (43/420)}} & \makecell{19.3\% \\ {\tiny (81/420)}} & \makecell{15.8\% \\ {\tiny (66/419)}} & \makecell{19.4\% \\ {\tiny (81/418)}} & \makecell{24.3\% \\ {\tiny (98/404)}} & \makecell{15.1\% \\ {\tiny (62/410)}} & \makecell{9.0\% \\ {\tiny (38/420)}} & \makecell{25.7\% \\ {\tiny (108/420)}} \\
C6 Genre / style fit & \makecell{70.7\% \\ {\tiny (297/420)}} & \makecell{61.1\% \\ {\tiny (256/419)}} & \makecell{69.0\% \\ {\tiny (289/419)}} & \makecell{72.9\% \\ {\tiny (306/420)}} & \makecell{73.3\% \\ {\tiny (308/420)}} & \makecell{60.4\% \\ {\tiny (252/417)}} & \makecell{78.3\% \\ {\tiny (329/420)}} & \makecell{71.7\% \\ {\tiny (301/420)}} & \makecell{72.6\% \\ {\tiny (304/419)}} & \makecell{66.7\% \\ {\tiny (279/418)}} & \makecell{79.7\% \\ {\tiny (322/404)}} & \makecell{73.9\% \\ {\tiny (303/410)}} & \makecell{65.5\% \\ {\tiny (275/420)}} & \makecell{51.4\% \\ {\tiny (216/420)}} \\
\bottomrule
\end{tabular}%
}
\end{table}

The descriptive tables above establish the baseline: how often each criterion enters the rationale space at all, and in how many pairs it is present on both sides and therefore available for stance-based comparison. We can now ask whether the qualitative pattern---AI evaluators praising the same textual feature under one authorship label and criticizing it under the other---generalizes to the full corpus.

\subsection*{Aggregate valence patterns}

\begin{figure}[!htbp]
\centering
\includegraphics[width=0.8\linewidth]{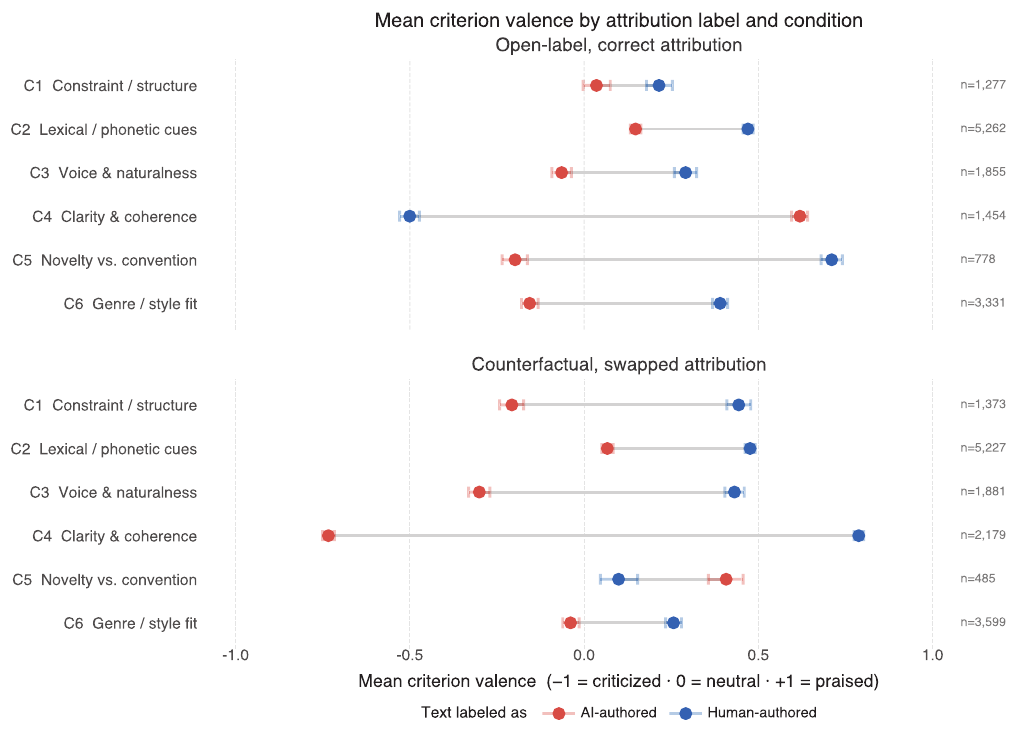}
\caption{Mean criterion valence toward text labeled ``Human-authored'' (blue) versus ``AI-authored'' (red), by criterion and attribution condition. Top: open-label (correct attribution). Bottom: counterfactual (labels swapped). Each point is the mean valence; horizontal bars indicate 95\% bootstrap CIs. $n$ per criterion is shown on the right. Underlying texts are identical across conditions; only the labels change.}
\label{fig:si_valence_by_label}
\end{figure}

Figure~\ref{fig:si_valence_by_label} gives the broadest view of that pattern. For each criterion, it plots mean valence toward the text labeled `Human-authored' and toward the text labeled `AI-authored', both under correct attribution (top panel) and under counterfactual attribution (bottom panel). The underlying texts are the same across both panels; only the labels move. At this level of aggregation, the figure asks a simple question: once a criterion is invoked, does its evaluative valence stay with the underlying text, or does it shift toward whichever text is labeled `Human-authored'? Across most criteria, the pattern is clear. Positive valence clusters around the text labeled `Human-authored', while the text labeled `AI-authored' is judged more negatively. This remains true in the counterfactual condition, where the labels are reversed and the underlying texts are unchanged. The effect is especially visible for C2 (lexical and/or phonetic cues), C3 (voice and naturalness), and C6 (genre and style fit). Two criteria stand apart from this label-following pattern: C4 (clarity and coherence) and C5 (novelty vs. convention). In C4, valence does not simply migrate with the human label. The Queneau text is evaluated more negatively on coherence in both conditions, yet even \textit{more} negatively when it appears under the AI label. Whatever attribution is doing here, it is operating on top of a textual contrast that evaluators seem able to detect. C5 shows the opposite tension. Novelty remains more favorable to the Queneau text even under swapped attribution, which suggests that, for novelty, the underlying difference between the texts is not easily neutralized at the level of valence. The figure does not show that attribution erases textual differences altogether. It shows, more precisely, that attribution exerts a strong directional pull on evaluation, while some criteria remain more resistant than others.

\subsubsection*{Salience shifts} 
\begin{figure}[!htbp]
\centering
\includegraphics[width=0.8\linewidth]{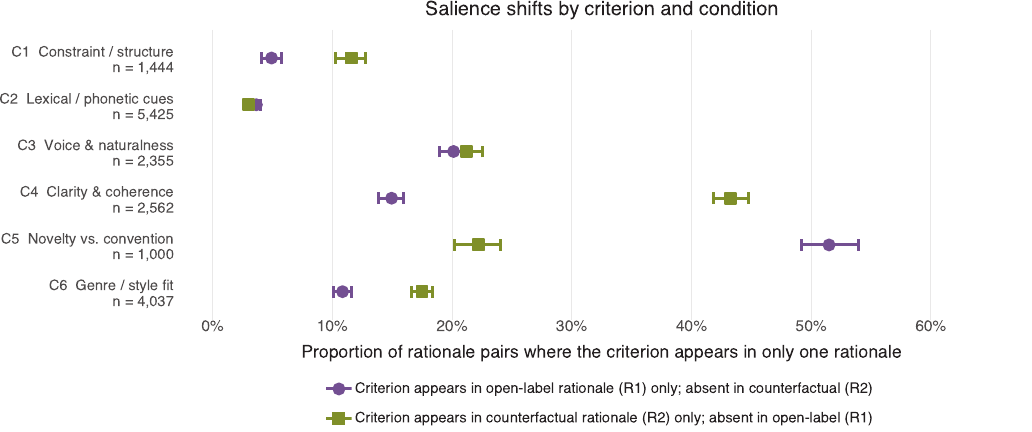}
\caption{Salience shifts by criterion and condition. Among pairs in which a criterion receives a consensus mention ($\geq$2/3 coders agreed that the criterion is present) in at least one rationale, the figure shows the proportion in which it appears only in the open-label rationale (R1; purple circles) or only in the counterfactual rationale (R2; olive squares). Each point is the mean proportion; horizontal bars indicate 95\% bootstrap CIs. $n$ per criterion is shown on the right and gives the number of mentioned pairs used as the denominator.}
\label{fig:si_salience_shifts}
\end{figure}

Figure~\ref{fig:si_salience_shifts} isolates the first mechanism behind the aggregate valence pattern: salience. Shown here are pairs in which a criterion is mentioned at least once, split by whether it appears \textit{only} in the open-label rationale (R1) or \textit{only} in the counterfactual rationale (R2). The asymmetry is not uniform across criteria. C2 (lexical and/or phonetic cues) is almost never one-sided (3.8\% R1-only; 3.1\% R2-only), which suggests that lexical evidence remains in play across both attribution frames. When C2 shifts, it is therefore less a matter of entering or leaving the rationale space than of being made to support different conclusions once invoked. C4 (clarity and coherence) and C5 (novelty versus convention) show the opposite pattern. Clarity is far more likely to surface only in the counterfactual rationale (43.2\% vs.\ 15.2\%), whereas novelty is much more likely to appear only in the open-label rationale (51.5\% vs.\ 22.1\%). In the logic of this experiment, that means attribution is not merely tilting the valence of already-active criteria. It is helping determine which evaluative dimensions are recruited in the first place. The effect is especially visible for the two criteria that already stood out in Figure~\ref{fig:si_valence_by_label}: coherence becomes markedly more available as a justificatory dimension under swapped attribution, while novelty recedes once Queneau's text wears the AI label. C1 and C6 show smaller but similar counterfactual skews, whereas C3 is comparatively balanced. Taken together, the figure suggests that criterion-level shifting begins before any stance reversal occurs: the attribution frame first reorganizes what is talked about, and only then how those same criteria are applied.

Figure~\ref{fig:si_salience_shifts} shows how often each criterion appears only in the open-label rationale (R1) or only in the counterfactual rationale (R2), but it aggregates across all 14 evaluator models. We already know from Table~\ref{tab:si_criterion_mention_by_evaluator} that evaluators differ in how often they invoke each criterion at all. So a natural next question is whether the split between R1-only (criterion in the open-label rationale only) and R2-only (criterion in the counterfactual rationale only) also varies by evaluator, and where it does. Table~\ref{tab:si_salience_split_by_evaluator} answers that by breaking the same quantity down by model. 

The table uses the same notion of ``eligible'' pairs as Figure~\ref{fig:si_salience_shifts}. A pair is eligible for a given criterion when the coders agreed (at least two of three) that the criterion appears in at least one of the two rationales---R1 or R2 or both. The two percentages in the cell then are both ``out of'' that same number: among those eligible pairs, what share have the criterion only in R1, and what share only in R2. The remainder have the criterion in both rationales (they are eligible, but they are not ``R1-only'' or ``R2-only''). A concrete example: for C1 (constraint/structure) and GPT-4o Mini, 73 of that evaluator’s 420 pairs are eligible (C1 in at least one rationale). The cell shows 3\% / 8\% and (73) below: among those 73 pairs, 3\% are R1-only and 8\% R2-only; the rest have C1 in both rationales.

\begin{landscape}
\begin{table}[!htbp]
\centering
\caption{Salience split by evaluator model. Rows: the six criteria. Columns: the 14 AI evaluator models, plus a final column (All evaluators) giving the corpus-wide split per criterion. Among pairs in which a criterion receives a consensus mention ($\geq$2 of 3 LLM coders) in at least one rationale (the same denominator as in Fig.~\ref{fig:si_salience_shifts}), each cell shows the proportion in which the criterion appears only in the open-label rationale (R1; first number, purple) or only in the counterfactual rationale (R2; second number, olive); colors match Fig.~\ref{fig:si_salience_shifts}. The number of eligible pairs ($n$) for that criterion and evaluator appears below in small type. The final column repeats the corpus-wide R1-only / R2-only split per criterion; its $n$ is the same as the ``at least one rationale'' count in Table~\ref{tab:si_criterion_overall}.}
\label{tab:si_salience_split_by_evaluator}
\small
\setlength{\tabcolsep}{3pt}
\resizebox{\linewidth}{!}{%
\begin{tabular}{@{} l *{14}{c} c @{}}
\toprule
\textbf{Criterion} & \makecell{GPT-4o \\ Mini} & \makecell{GPT-3.5 \\ Turbo} & GPT-4 & \makecell{Claude \\ Sonnet 4} & \makecell{Claude 3.5 \\ Haiku} & \makecell{Gemini 2.5 \\ Flash} & \makecell{Mistral \\ Medium} & \makecell{Mistral \\ Nemo} & \makecell{Llama 4 \\ Maverick} & \makecell{Llama 3.3 \\ 70B} & \makecell{DeepSeek \\ R1} & \makecell{DeepSeek \\ Chat v3} & \makecell{Qwen 2.5 \\ 72B} & \makecell{Command \\ R} & \makecell{\textbf{All} \\ \textbf{evaluators}} \\
\midrule
C1 Constraint / structure
& \makecell{\textcolor{r1onlycolor}{3\%} / \textcolor{r2onlycolor}{8\%} \\ {\tiny (73)}}
& \makecell{\textcolor{r1onlycolor}{2\%} / \textcolor{r2onlycolor}{25\%} \\ {\tiny (48)}}
& \makecell{\textcolor{r1onlycolor}{2\%} / \textcolor{r2onlycolor}{8\%} \\ {\tiny (118)}}
& \makecell{\textcolor{r1onlycolor}{4\%} / \textcolor{r2onlycolor}{10\%} \\ {\tiny (134)}}
& \makecell{\textcolor{r1onlycolor}{5\%} / \textcolor{r2onlycolor}{8\%} \\ {\tiny (109)}}
& \makecell{\textcolor{r1onlycolor}{3\%} / \textcolor{r2onlycolor}{10\%} \\ {\tiny (109)}}
& \makecell{\textcolor{r1onlycolor}{4\%} / \textcolor{r2onlycolor}{12\%} \\ {\tiny (112)}}
& \makecell{\textcolor{r1onlycolor}{12\%} / \textcolor{r2onlycolor}{18\%} \\ {\tiny (105)}}
& \makecell{\textcolor{r1onlycolor}{3\%} / \textcolor{r2onlycolor}{5\%} \\ {\tiny (94)}}
& \makecell{\textcolor{r1onlycolor}{1\%} / \textcolor{r2onlycolor}{10\%} \\ {\tiny (83)}}
& \makecell{\textcolor{r1onlycolor}{6\%} / \textcolor{r2onlycolor}{16\%} \\ {\tiny (161)}}
& \makecell{\textcolor{r1onlycolor}{7\%} / \textcolor{r2onlycolor}{11\%} \\ {\tiny (118)}}
& \makecell{\textcolor{r1onlycolor}{6\%} / \textcolor{r2onlycolor}{9\%} \\ {\tiny (91)}}
& \makecell{\textcolor{r1onlycolor}{9\%} / \textcolor{r2onlycolor}{16\%} \\ {\tiny (89)}}
& \makecell{\textcolor{r1onlycolor}{4.9\%} / \textcolor{r2onlycolor}{11.6\%} \\ {\tiny (1,446)}} \\
C2 Lexical / phonetic cues
& \makecell{\textcolor{r1onlycolor}{3\%} / \textcolor{r2onlycolor}{1\%} \\ {\tiny (386)}}
& \makecell{\textcolor{r1onlycolor}{2\%} / \textcolor{r2onlycolor}{1\%} \\ {\tiny (414)}}
& \makecell{\textcolor{r1onlycolor}{5\%} / \textcolor{r2onlycolor}{5\%} \\ {\tiny (314)}}
& \makecell{\textcolor{r1onlycolor}{0\%} / \textcolor{r2onlycolor}{1\%} \\ {\tiny (408)}}
& \makecell{\textcolor{r1onlycolor}{2\%} / \textcolor{r2onlycolor}{1\%} \\ {\tiny (414)}}
& \makecell{\textcolor{r1onlycolor}{6\%} / \textcolor{r2onlycolor}{6\%} \\ {\tiny (376)}}
& \makecell{\textcolor{r1onlycolor}{6\%} / \textcolor{r2onlycolor}{3\%} \\ {\tiny (375)}}
& \makecell{\textcolor{r1onlycolor}{3\%} / \textcolor{r2onlycolor}{4\%} \\ {\tiny (406)}}
& \makecell{\textcolor{r1onlycolor}{2\%} / \textcolor{r2onlycolor}{3\%} \\ {\tiny (395)}}
& \makecell{\textcolor{r1onlycolor}{3\%} / \textcolor{r2onlycolor}{1\%} \\ {\tiny (404)}}
& \makecell{\textcolor{r1onlycolor}{2\%} / \textcolor{r2onlycolor}{2\%} \\ {\tiny (401)}}
& \makecell{\textcolor{r1onlycolor}{5\%} / \textcolor{r2onlycolor}{5\%} \\ {\tiny (380)}}
& \makecell{\textcolor{r1onlycolor}{5\%} / \textcolor{r2onlycolor}{5\%} \\ {\tiny (371)}}
& \makecell{\textcolor{r1onlycolor}{6\%} / \textcolor{r2onlycolor}{6\%} \\ {\tiny (381)}}
& \makecell{\textcolor{r1onlycolor}{3.6\%} / \textcolor{r2onlycolor}{3.0\%} \\ {\tiny (5,426)}} \\
C3 Voice \& naturalness
& \makecell{\textcolor{r1onlycolor}{26\%} / \textcolor{r2onlycolor}{15\%} \\ {\tiny (126)}}
& \makecell{\textcolor{r1onlycolor}{36\%} / \textcolor{r2onlycolor}{26\%} \\ {\tiny (77)}}
& \makecell{\textcolor{r1onlycolor}{45\%} / \textcolor{r2onlycolor}{18\%} \\ {\tiny (51)}}
& \makecell{\textcolor{r1onlycolor}{17\%} / \textcolor{r2onlycolor}{12\%} \\ {\tiny (239)}}
& \makecell{\textcolor{r1onlycolor}{12\%} / \textcolor{r2onlycolor}{10\%} \\ {\tiny (279)}}
& \makecell{\textcolor{r1onlycolor}{21\%} / \textcolor{r2onlycolor}{25\%} \\ {\tiny (152)}}
& \makecell{\textcolor{r1onlycolor}{26\%} / \textcolor{r2onlycolor}{34\%} \\ {\tiny (113)}}
& \makecell{\textcolor{r1onlycolor}{31\%} / \textcolor{r2onlycolor}{33\%} \\ {\tiny (122)}}
& \makecell{\textcolor{r1onlycolor}{22\%} / \textcolor{r2onlycolor}{22\%} \\ {\tiny (178)}}
& \makecell{\textcolor{r1onlycolor}{18\%} / \textcolor{r2onlycolor}{26\%} \\ {\tiny (208)}}
& \makecell{\textcolor{r1onlycolor}{15\%} / \textcolor{r2onlycolor}{20\%} \\ {\tiny (249)}}
& \makecell{\textcolor{r1onlycolor}{15\%} / \textcolor{r2onlycolor}{20\%} \\ {\tiny (246)}}
& \makecell{\textcolor{r1onlycolor}{12\%} / \textcolor{r2onlycolor}{19\%} \\ {\tiny (182)}}
& \makecell{\textcolor{r1onlycolor}{35\%} / \textcolor{r2onlycolor}{38\%} \\ {\tiny (133)}}
& \makecell{\textcolor{r1onlycolor}{20.1\%} / \textcolor{r2onlycolor}{21.2\%} \\ {\tiny (2,358)}} \\
C4 Clarity \& coherence
& \makecell{\textcolor{r1onlycolor}{13\%} / \textcolor{r2onlycolor}{34\%} \\ {\tiny (279)}}
& \makecell{\textcolor{r1onlycolor}{32\%} / \textcolor{r2onlycolor}{61\%} \\ {\tiny (28)}}
& \makecell{\textcolor{r1onlycolor}{14\%} / \textcolor{r2onlycolor}{30\%} \\ {\tiny (159)}}
& \makecell{\textcolor{r1onlycolor}{10\%} / \textcolor{r2onlycolor}{46\%} \\ {\tiny (209)}}
& \makecell{\textcolor{r1onlycolor}{11\%} / \textcolor{r2onlycolor}{43\%} \\ {\tiny (162)}}
& \makecell{\textcolor{r1onlycolor}{13\%} / \textcolor{r2onlycolor}{66\%} \\ {\tiny (114)}}
& \makecell{\textcolor{r1onlycolor}{10\%} / \textcolor{r2onlycolor}{54\%} \\ {\tiny (217)}}
& \makecell{\textcolor{r1onlycolor}{26\%} / \textcolor{r2onlycolor}{47\%} \\ {\tiny (145)}}
& \makecell{\textcolor{r1onlycolor}{18\%} / \textcolor{r2onlycolor}{41\%} \\ {\tiny (152)}}
& \makecell{\textcolor{r1onlycolor}{16\%} / \textcolor{r2onlycolor}{49\%} \\ {\tiny (166)}}
& \makecell{\textcolor{r1onlycolor}{18\%} / \textcolor{r2onlycolor}{28\%} \\ {\tiny (254)}}
& \makecell{\textcolor{r1onlycolor}{12\%} / \textcolor{r2onlycolor}{44\%} \\ {\tiny (250)}}
& \makecell{\textcolor{r1onlycolor}{14\%} / \textcolor{r2onlycolor}{37\%} \\ {\tiny (238)}}
& \makecell{\textcolor{r1onlycolor}{19\%} / \textcolor{r2onlycolor}{57\%} \\ {\tiny (189)}}
& \makecell{\textcolor{r1onlycolor}{14.9\%} / \textcolor{r2onlycolor}{43.3\%} \\ {\tiny (2,565)}} \\
C5 Novelty vs.\ convention
& \makecell{\textcolor{r1onlycolor}{44\%} / \textcolor{r2onlycolor}{30\%} \\ {\tiny (61)}}
& \makecell{\textcolor{r1onlycolor}{64\%} / \textcolor{r2onlycolor}{16\%} \\ {\tiny (50)}}
& \makecell{\textcolor{r1onlycolor}{10\%} / \textcolor{r2onlycolor}{63\%} \\ {\tiny (30)}}
& \makecell{\textcolor{r1onlycolor}{53\%} / \textcolor{r2onlycolor}{13\%} \\ {\tiny (90)}}
& \makecell{\textcolor{r1onlycolor}{49\%} / \textcolor{r2onlycolor}{13\%} \\ {\tiny (148)}}
& \makecell{\textcolor{r1onlycolor}{48\%} / \textcolor{r2onlycolor}{27\%} \\ {\tiny (44)}}
& \makecell{\textcolor{r1onlycolor}{49\%} / \textcolor{r2onlycolor}{35\%} \\ {\tiny (43)}}
& \makecell{\textcolor{r1onlycolor}{51\%} / \textcolor{r2onlycolor}{28\%} \\ {\tiny (81)}}
& \makecell{\textcolor{r1onlycolor}{38\%} / \textcolor{r2onlycolor}{35\%} \\ {\tiny (66)}}
& \makecell{\textcolor{r1onlycolor}{58\%} / \textcolor{r2onlycolor}{21\%} \\ {\tiny (81)}}
& \makecell{\textcolor{r1onlycolor}{62\%} / \textcolor{r2onlycolor}{11\%} \\ {\tiny (98)}}
& \makecell{\textcolor{r1onlycolor}{71\%} / \textcolor{r2onlycolor}{14\%} \\ {\tiny (62)}}
& \makecell{\textcolor{r1onlycolor}{42\%} / \textcolor{r2onlycolor}{32\%} \\ {\tiny (38)}}
& \makecell{\textcolor{r1onlycolor}{53\%} / \textcolor{r2onlycolor}{22\%} \\ {\tiny (108)}}
& \makecell{\textcolor{r1onlycolor}{51.5\%} / \textcolor{r2onlycolor}{22.2\%} \\ {\tiny (1,005)}} \\
C6 Genre / style fit
& \makecell{\textcolor{r1onlycolor}{9\%} / \textcolor{r2onlycolor}{15\%} \\ {\tiny (297)}}
& \makecell{\textcolor{r1onlycolor}{20\%} / \textcolor{r2onlycolor}{22\%} \\ {\tiny (256)}}
& \makecell{\textcolor{r1onlycolor}{9\%} / \textcolor{r2onlycolor}{17\%} \\ {\tiny (289)}}
& \makecell{\textcolor{r1onlycolor}{8\%} / \textcolor{r2onlycolor}{16\%} \\ {\tiny (306)}}
& \makecell{\textcolor{r1onlycolor}{4\%} / \textcolor{r2onlycolor}{17\%} \\ {\tiny (308)}}
& \makecell{\textcolor{r1onlycolor}{13\%} / \textcolor{r2onlycolor}{20\%} \\ {\tiny (252)}}
& \makecell{\textcolor{r1onlycolor}{11\%} / \textcolor{r2onlycolor}{10\%} \\ {\tiny (329)}}
& \makecell{\textcolor{r1onlycolor}{13\%} / \textcolor{r2onlycolor}{17\%} \\ {\tiny (301)}}
& \makecell{\textcolor{r1onlycolor}{10\%} / \textcolor{r2onlycolor}{18\%} \\ {\tiny (304)}}
& \makecell{\textcolor{r1onlycolor}{11\%} / \textcolor{r2onlycolor}{15\%} \\ {\tiny (279)}}
& \makecell{\textcolor{r1onlycolor}{4\%} / \textcolor{r2onlycolor}{16\%} \\ {\tiny (322)}}
& \makecell{\textcolor{r1onlycolor}{12\%} / \textcolor{r2onlycolor}{17\%} \\ {\tiny (303)}}
& \makecell{\textcolor{r1onlycolor}{14\%} / \textcolor{r2onlycolor}{16\%} \\ {\tiny (275)}}
& \makecell{\textcolor{r1onlycolor}{18\%} / \textcolor{r2onlycolor}{35\%} \\ {\tiny (216)}}
& \makecell{\textcolor{r1onlycolor}{10.8\%} / \textcolor{r2onlycolor}{17.5\%} \\ {\tiny (4,038)}} \\
\bottomrule
\end{tabular}%
}
\end{table}
\end{landscape}

\subsubsection*{Stance reversals} Attribution can change which evaluative dimensions enter the rationale at all. The next step is to examine cases of \textit{stance reversal}: instances in which the same criterion is present in both rationales for the same paired evaluation. Here the criterion does not disappear from one condition and reappear in the other; it remains in play across both. What can still change is which underlying text that criterion supports. A pair is stance-eligible for a given criterion when at least two of three coders agree that the criterion is present in both rationales and the coding yields a defined majority stance in each; within that subset, a reversal is recorded when the consensus stance switches which underlying text the criterion favors. The corpus-wide rates are plotted in the main text (Fig.~\ref{fig:stance_reversal}); Table~\ref{tab:si_stance_reversal_by_criterion} reports the corresponding numerical values with confidence intervals.

\begin{table}[h!]
\centering
\small
\caption{\textbf{Stance-reversal rate by criterion} ($n=5{,}846$ paired rationales). A pair is stance-eligible for a given criterion when at least two of three coders agree that the criterion is present in both rationales and the coding yields a defined majority stance in each. Within that subset, a reversal is recorded when the consensus stance switches which underlying text the criterion favors between the open-label and counterfactual rationale. Bootstrap 95\% confidence intervals are clustered by pair. These are the same values plotted in Fig.~\ref{fig:stance_reversal} in the main article.}
\label{tab:si_stance_reversal_by_criterion}
\begin{tabular}{@{} l r r r r @{}}
\toprule
\textbf{Criterion} & \textbf{Reversal rate} & \textbf{95\% CI} & $n_{\text{eligible}}$ & $n_{\text{reversals}}$ \\
\midrule
C1 Constraint / structure   & 30.1\% & [27.4, 32.6] & 1,204 & 362 \\
C2 Lexical / phonetic cues  & 31.4\% & [30.1, 32.7] & 5,056 & 1,589 \\
C3 Voice \& naturalness     & 25.4\% & [23.1, 27.6] & 1,364 & 346 \\
C4 Clarity / coherence      &  9.0\% & [7.4, 10.8]  & 1,057 &  95 \\
C5 Novelty vs.\ convention  & 13.1\% & [9.2, 17.3]  &   260 &  34 \\
C6 Genre / style fit        & 28.6\% & [26.9, 30.3] & 2,854 & 817 \\
\bottomrule
\end{tabular}
\end{table}

Several patterns stand out. C1 (constraint and structure), C2 (lexical and/or phonetic cues), and C6 (genre and style fit) show the highest conditional reversal rates: 30.1\% for C1 ($n=1{,}204$), 31.4\% for C2 ($n=5{,}056$), and 28.6\% for C6 ($n=2{,}854$). In other words, once these criteria are present in both rationales, they support a different underlying text in about three out of every ten paired evaluations when the attribution labels are swapped. C3 (voice and naturalness) is somewhat lower, at 25.4\% ($n=1{,}364$). At the other end, C4 (clarity and coherence) and C5 (novelty versus convention) reverse much less often once they are present in both rationales: 9.0\% for C4 ($n=1{,}057$) and 13.1\% for C5 ($n=260$). That contrast fits the earlier results. The two criteria that appeared least label-driven in Figure~\ref{fig:si_valence_by_label}, and most selective at the level of salience in Figure~\ref{fig:si_salience_shifts}, are also the least likely to reverse direction once they remain in play. Coherence and novelty appear to stay more closely tied to the texts themselves, whereas constraint, lexical cues, and genre fit are more readily redirected under changing attribution. As Table~\ref{tab:si_criterion_overall} makes clear, the smaller $n$ values here reflect the stricter denominator: this analysis includes only pairs in which the criterion is present in both rationales.

Table~\ref{tab:si_stance_reversal_by_evaluator} breaks the corpus-wide reversal rates down by evaluator model. The pattern is broadly consistent: all 14 models show reversal rates above zero for C1, C2, and C6, and most show low rates for C4 and C5. Two models stand out as particularly label-susceptible---Command~R and Mistral~Medium show reversal rates roughly double the corpus average across most criteria---while GPT-4 is the most stable, with the lowest rates on five of six criteria. Rates for C5 should be interpreted with caution for evaluators with small stance-eligible samples.

\begin{landscape}
\begin{table}[!htbp]
\centering
\caption{\textbf{Conditional stance-reversal rate by evaluator model.} Rows: the six criteria. Columns: the 14 AI evaluator models, plus a final column (All evaluators) giving the corpus-wide rate per criterion (the same values as in Table~\ref{tab:si_stance_reversal_by_criterion}). A pair is stance-eligible for a given criterion when at least two of three coders agree that the criterion is present in both rationales and the coding yields a defined majority stance in each. Each cell shows the proportion of stance-eligible pairs in which the consensus stance reverses between the open-label and counterfactual rationale; the number of stance-eligible pairs ($n$) appears below in small type. Consensus: $\geq$2 of 3 LLM coders.}
\label{tab:si_stance_reversal_by_evaluator}
\small
\setlength{\tabcolsep}{3pt}
\resizebox{\linewidth}{!}{%
\begin{tabular}{@{} l *{14}{c} c @{}}
\toprule
\textbf{Criterion} & \makecell{GPT-4o \\ Mini} & \makecell{GPT-3.5 \\ Turbo} & GPT-4 & \makecell{Claude \\ Sonnet 4} & \makecell{Claude 3.5 \\ Haiku} & \makecell{Gemini 2.5 \\ Flash} & \makecell{Mistral \\ Medium} & \makecell{Mistral \\ Nemo} & \makecell{Llama 4 \\ Maverick} & \makecell{Llama 3.3 \\ 70B} & \makecell{DeepSeek \\ R1} & \makecell{DeepSeek \\ Chat v3} & \makecell{Qwen 2.5 \\ 72B} & \makecell{Command \\ R} & \makecell{\textbf{All} \\ \textbf{evaluators}} \\
\midrule
C1 Constraint / structure
& \makecell{24.6\% \\ {\tiny (65)}}
& \makecell{42.9\% \\ {\tiny (35)}}
& \makecell{21.7\% \\ {\tiny (106)}}
& \makecell{20.0\% \\ {\tiny (115)}}
& \makecell{25.5\% \\ {\tiny (94)}}
& \makecell{29.5\% \\ {\tiny (95)}}
& \makecell{45.2\% \\ {\tiny (93)}}
& \makecell{43.1\% \\ {\tiny (72)}}
& \makecell{22.1\% \\ {\tiny (86)}}
& \makecell{20.3\% \\ {\tiny (74)}}
& \makecell{23.6\% \\ {\tiny (127)}}
& \makecell{36.1\% \\ {\tiny (97)}}
& \makecell{29.5\% \\ {\tiny (78)}}
& \makecell{56.7\% \\ {\tiny (67)}}
& \makecell{30.1\% \\ {\tiny (1,204)}} \\
C2 Lexical / phonetic cues
& \makecell{31.0\% \\ {\tiny (371)}}
& \makecell{38.6\% \\ {\tiny (399)}}
& \makecell{13.5\% \\ {\tiny (281)}}
& \makecell{34.2\% \\ {\tiny (401)}}
& \makecell{23.1\% \\ {\tiny (399)}}
& \makecell{31.9\% \\ {\tiny (332)}}
& \makecell{48.1\% \\ {\tiny (339)}}
& \makecell{35.5\% \\ {\tiny (377)}}
& \makecell{21.4\% \\ {\tiny (374)}}
& \makecell{25.8\% \\ {\tiny (387)}}
& \makecell{24.7\% \\ {\tiny (388)}}
& \makecell{31.2\% \\ {\tiny (343)}}
& \makecell{24.7\% \\ {\tiny (332)}}
& \makecell{55.6\% \\ {\tiny (333)}}
& \makecell{31.4\% \\ {\tiny (5,056)}} \\
C3 Voice \& naturalness
& \makecell{24.7\% \\ {\tiny (73)}}
& \makecell{41.4\% \\ {\tiny (29)}}
& \makecell{0.0\% \\ {\tiny (19)}}
& \makecell{28.7\% \\ {\tiny (167)}}
& \makecell{20.8\% \\ {\tiny (216)}}
& \makecell{29.3\% \\ {\tiny (82)}}
& \makecell{59.1\% \\ {\tiny (44)}}
& \makecell{45.2\% \\ {\tiny (42)}}
& \makecell{20.0\% \\ {\tiny (100)}}
& \makecell{17.4\% \\ {\tiny (115)}}
& \makecell{17.7\% \\ {\tiny (158)}}
& \makecell{29.8\% \\ {\tiny (161)}}
& \makecell{18.9\% \\ {\tiny (122)}}
& \makecell{41.7\% \\ {\tiny (36)}}
& \makecell{25.4\% \\ {\tiny (1,364)}} \\
C4 Clarity \& coherence
& \makecell{12.3\% \\ {\tiny (146)}}
& \makecell{50.0\% \\ {\tiny (2)}}
& \makecell{1.1\% \\ {\tiny (89)}}
& \makecell{18.9\% \\ {\tiny (90)}}
& \makecell{4.1\% \\ {\tiny (74)}}
& \makecell{12.5\% \\ {\tiny (24)}}
& \makecell{9.3\% \\ {\tiny (75)}}
& \makecell{13.2\% \\ {\tiny (38)}}
& \makecell{5.1\% \\ {\tiny (59)}}
& \makecell{7.0\% \\ {\tiny (57)}}
& \makecell{3.0\% \\ {\tiny (135)}}
& \makecell{3.7\% \\ {\tiny (108)}}
& \makecell{8.6\% \\ {\tiny (116)}}
& \makecell{34.1\% \\ {\tiny (44)}}
& \makecell{9.0\% \\ {\tiny (1,057)}} \\
C5 Novelty vs.\ convention
& \makecell{0.0\% \\ {\tiny (16)}}
& \makecell{22.2\% \\ {\tiny (9)}}
& \makecell{0.0\% \\ {\tiny (8)}}
& \makecell{0.0\% \\ {\tiny (29)}}
& \makecell{3.6\% \\ {\tiny (56)}}
& \makecell{9.1\% \\ {\tiny (11)}}
& \makecell{0.0\% \\ {\tiny (7)}}
& \makecell{17.6\% \\ {\tiny (17)}}
& \makecell{11.1\% \\ {\tiny (18)}}
& \makecell{17.6\% \\ {\tiny (17)}}
& \makecell{3.8\% \\ {\tiny (26)}}
& \makecell{44.4\% \\ {\tiny (9)}}
& \makecell{0.0\% \\ {\tiny (10)}}
& \makecell{59.3\% \\ {\tiny (27)}}
& \makecell{13.1\% \\ {\tiny (260)}} \\
C6 Genre / style fit
& \makecell{30.7\% \\ {\tiny (225)}}
& \makecell{34.2\% \\ {\tiny (149)}}
& \makecell{12.8\% \\ {\tiny (211)}}
& \makecell{27.0\% \\ {\tiny (230)}}
& \makecell{22.1\% \\ {\tiny (240)}}
& \makecell{33.5\% \\ {\tiny (164)}}
& \makecell{42.0\% \\ {\tiny (257)}}
& \makecell{34.1\% \\ {\tiny (208)}}
& \makecell{21.1\% \\ {\tiny (218)}}
& \makecell{25.5\% \\ {\tiny (204)}}
& \makecell{23.8\% \\ {\tiny (248)}}
& \makecell{28.4\% \\ {\tiny (215)}}
& \makecell{25.4\% \\ {\tiny (189)}}
& \makecell{57.3\% \\ {\tiny (96)}}
& \makecell{28.6\% \\ {\tiny (2,854)}} \\
\bottomrule
\end{tabular}%
}
\end{table}
\end{landscape}

Because the underlying texts are identical across the two conditions and only the attribution labels change, the natural benchmark here is stability. If attribution had no effect on which text the evaluator favors when a criterion is in play, stance reversals should remain relatively uncommon: the same underlying text would usually be favored in both rationales. Stance reversal therefore marks a departure from that benchmark: the same evaluative dimension remains in play, but changes sides when the labels are swapped.

\subsubsection*{Decision linkage and directionality of reversals} There are, however, two distinct layers to keep in mind. The stance reversals and salience shifts discussed here are measured at the rationale layer: they describe whether and how the explanation changed---what was talked about, how each criterion was applied, and which text each criterion was said to support. The decision layer is separate: it concerns whether the evaluator's final choice \textit{actually} changed---whether they picked the same underlying text under open-label and under counterfactual attribution, or a different one. The former comes from our coding of the rationales; the latter from the evaluator's stated choice in each condition. Linking the two layers shows whether the criterion-level shifts in the rationales track the change in the evaluator's verdict. Across the corpus, they do: when the evaluator's forced-choice verdict reverses between conditions, at least one criterion stance reversal co-occurs in roughly 96\% of pairs [95\% CI: 95–97\%]; when the verdict does not reverse, the same figure is roughly 1\% [95\% CI: 0.7–1.3\%]. That concentration of coded stance reversals within verdict-reversal pairs indicates that the criterion-level changes we extract from the rationales align with the behavioral stance reversal: the explanation changes in step with the choice.

\begin{figure}[!htbp]
\centering
\includegraphics[width=0.8\linewidth]{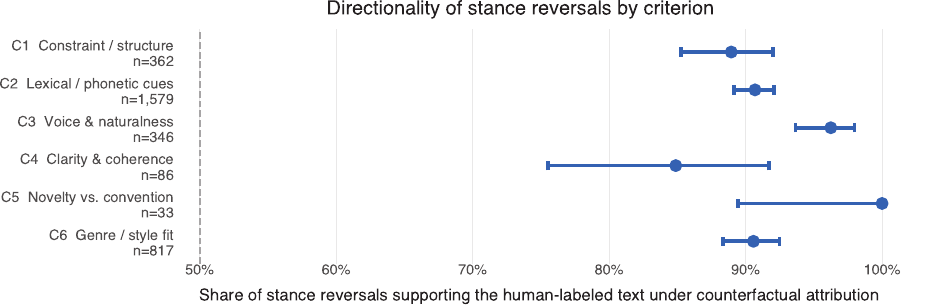}
\caption{Directionality of stance reversals by criterion. Among consensus stance reversals (the same set as in Table~\ref{tab:si_stance_reversal_by_criterion}), each point shows the share that support the human-labeled text under counterfactual attribution. A stance reversal is counted when at least two of three coders agreed that the criterion was present in both rationales and agreed on the valence in each rationale (which of the two texts the criterion favored); the reversal is the change in that consensus stance between the open-label and counterfactual rationales. The horizontal axis gives the proportion of those reversals that shift support toward the text labeled ``Human-authored'' in the counterfactual condition. Horizontal bars indicate 95\% exact binomial confidence intervals; the vertical dashed line at 0.5 marks chance. The $n$ on the right gives the number of stance reversals per criterion (the denominator for each point).}
\label{fig:si_directionality}
\end{figure}

A final question is whether those reversals are directionally coherent with the attribution manipulation. When a criterion reverses—supporting a different underlying text in the counterfactual rationale than in the open-label one—it could in principle favor either the text labeled human or the text labeled AI in that condition. If the change were unrelated to the labels, we would expect roughly half of reversals to go each way. If attribution is driving the shift, we would expect reversals to concentrate in one direction: toward the text wearing the human label in the counterfactual condition (because that is the condition in which labels are swapped). Figure~\ref{fig:si_directionality} addresses this by plotting, for each criterion, the share of stance reversals that support the human-labeled text under counterfactual attribution. The denominator here is not the stance-eligible pairs from Table~\ref{tab:si_stance_reversal_by_criterion} (e.g., 1,204 for C1), but the subset of those pairs in which a consensus stance reversal actually occurred (e.g., 362 for C1—the same 30\% that underlies the reversal rate in that figure). So the $n$ on the right in Figure~\ref{fig:si_directionality} is the number of reversals per criterion; we then ask what share of those reversals shift support toward the human-labeled text. The analysis is conservative in two ways: it conditions only on cases where a reversal occurred and uses a consensus definition (at least two of three coders agreed on the criterion’s presence in both rationales and on the valence in each). For criteria with fewer reversals (e.g., C5 with $n=33$), the per-criterion estimate is more uncertain, but the pattern is consistent: for every criterion, the share of reversals supporting the human-labeled text lies well above 0.5, with 95\% exact binomial confidence intervals that exclude chance. C5 (novelty versus convention) and C3 (voice and naturalness) show the highest shares; C4 (clarity and coherence) the lowest, though still around 85\%. In other words, when a criterion such as lexical cues or genre fit reverses, it does not reverse at random between the two texts; in the large majority of cases it shifts to support the text that is labeled human in that condition. From the perspective of the design—same texts, only labels changed between conditions—this indicates that the reversals we code are aligned with the attribution manipulation: the reallocation of evaluative support follows the label, rather than being symmetrically distributed.

\newpage

\subsection*{Changing the Narrative---Annotation codebook}

To test whether LLM evaluators change the \emph{criteria} and \emph{stance} of their reasoning across paired rationales (R1 vs.\ R2) for the same underlying comparison, we created an annotation codebook that turns close-reading observations into auditable, systematic coding. Two human annotators independently applied the codebook to 98 paired rationale cases. We then applied the same codebook to three LLM annotators. For LLMs, we required structured outputs (criterion, valence toward Version A/B, and evidence snippets) so every label remains traceable to text.

\begin{tcolorbox}[
  enhanced jigsaw,
  breakable,
  colback=codebookbg,
  colframe=codebookframe,
  boxrule=0.9pt,
  arc=3pt,
  left=14pt, right=14pt, top=12pt, bottom=14pt,
]

{\large\bfseries Annotation Codebook}\par
\vspace{4pt}
{\color{rulegray}\hrule height 0.5pt}
\vspace{10pt}

You will read two short rationales---\textbf{R1} and \textbf{R2}---written
by an evaluator who compared the same two texts in two separate conditions.
Both rationales refer to \textbf{Version~A} and \textbf{Version~B}; treat
these as arbitrary labels and read each rationale on its own terms.

\medskip

\textbf{For each rationale (R1 and R2 independently):}
\begin{itemize}
  \item Identify the evaluative \textbf{criteria} that are explicitly present in the text.
  \item For each criterion, record the \textbf{valence} toward Version~A and toward
        Version~B separately (see scale below).
  \item Note \textbf{1--2 short supporting phrases} from the rationale for each criterion.
\end{itemize}

\medskip

\textbf{Valence scale}\par\smallskip

\noindent
\begin{tabularx}{\linewidth}{@{} >{\bfseries\centering}p{1.4cm}
                                    >{\centering}p{2.8cm}
                                    X @{}}
  \toprule
  Value & Label & When to use \\
  \midrule
  $+1$  & Positive        & \small{The rationale explicitly praises this version on the criterion, or frames it as a strength.} \\[4pt]
  $\phantom{+}0$   & Neutral / unclear & \small{This version is not evaluated on the criterion, or the phrasing is too vague to code.} \\[4pt]
  $-1$  & Negative        & \small{The rationale explicitly criticizes this version on the criterion, or frames it as a weakness.} \\
  \bottomrule
\end{tabularx}

\smallskip

\medskip

\textbf{Criteria}\quad(code only when the text clearly supports it;
max.\ 6 per rationale)\par\medskip

\begin{itemize}[leftmargin=0pt, itemsep=0pt]

  \crit{1}{Formal constraint, or structure adherence}{%
    The rationale judges whether a version complies with a formal
    requirement: meter, rhyme scheme, syllable count, line or stanza structure, a prescribed format or forbidden element.}

  \crit{2}{Lexical and phonetic style cues}{%
    The rationale points to concrete surface markers as evidence of style---dialect
    spellings, slang, stock phrases, phonetic renderings, or characteristic vocabulary.}

  \crit{3}{Voice, authenticity, naturalness}{%
    The rationale comments on whether the voice or register reads as natural and
    credible, or as forced, exaggerated, or caricatured. \textit{Do not code} for
    generic phrases like ``captures the essence of the style'' that say nothing
    about naturalness or forcedness.}

  \crit{4}{Clarity, coherence}{%
    The rationale evaluates ease of understanding: whether the text is clear and
    easy to follow, or confusing and hard to parse.}

  \crit{5}{Creativity, novelty, conventionality}{%
    The rationale comments on originality or inventiveness, or on predictable, conventional
    or formulaic execution.}

  \crit{6}{Norm-based genre and style fit}{%
    The rationale appeals explicitly to what ``counts as'' or is ``characteristic
    of'' the genre or style---invoking conventions or standards, not just surface
    markers. \textit{Only code C6} when the snippet contains both a norm-language
    term (e.g.\ \textit{typical, hallmark, characteristic, what the style is
    supposed to do}) and an explicit appeal to a standard. If the rationale only
    lists surface markers, use \textbf{C2} instead.}

\end{itemize}

{\color{rulegray}\hrule height 0.4pt}
\vspace{9pt}

\textbf{Before moving on}\par\smallskip
Every coded criterion needs at least one supporting phrase from the rationale. When in doubt, leave it out: fewer well-grounded codes are better than more speculative ones.

\end{tcolorbox}
\newpage

\FloatBarrier

\FloatBarrier

\FloatBarrier

\FloatBarrier

\begin{figure}
\centering
\includegraphics[width=\textwidth]{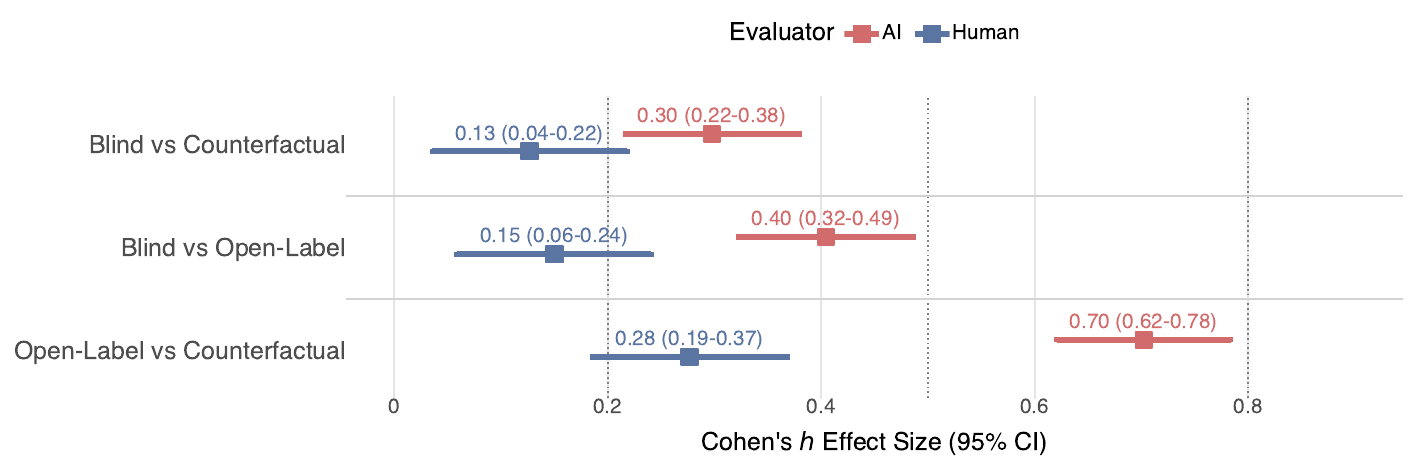}
\caption{\textbf{Attribution bias effect sizes across experimental conditions.} Standardized effect sizes (Cohen's \textit{h}) with 95\% confidence intervals measuring how different labeling conditions change AI story selection rates. Participants evaluated identical human-authored vs AI-generated story pairs under three between-subjects labeling conditions: Blind (no attribution labels shown), Open-Label (labels correctly identify each story's author), and Counterfactual (labels deliberately misattribute story authorship). Each comparison tests a specific mechanism. \textbf{`Blind vs Counterfactual'} (top) measures misinformation effects: do evaluators choose AI stories more when deceptively labeled as human-authored versus unlabeled? \textbf{`Blind vs Open-Label'} (middle) measures label awareness: do evaluators choose AI stories more when unlabeled versus labeled as AI-generated? \textbf{`Open-Label vs Counterfactual'} (bottom) measures pure attribution bias: do evaluators choose AI stories more when deceptively labeled as human-authored versus correctly labeled as AI-generated? Effect sizes further right indicate larger differences between the conditions, meaning authorship labels have stronger influence on story evaluation. Both human participants (blue) and AI model evaluators (red) show susceptibility to labeling manipulation, but AI models demonstrate stronger effects across all measures. The largest effects occur for pure attribution bias, where AI models show large effects (\textit{h}=0.70) compared to humans' moderate effects (\textit{h}=0.28). Non-overlapping confidence intervals confirm that AI models are more vulnerable to authorship misinformation than human evaluators.}
\label{fig:S5_forest_plot}
\end{figure}
\FloatBarrier

\begin{figure}
\centering
\includegraphics[width=\textwidth]{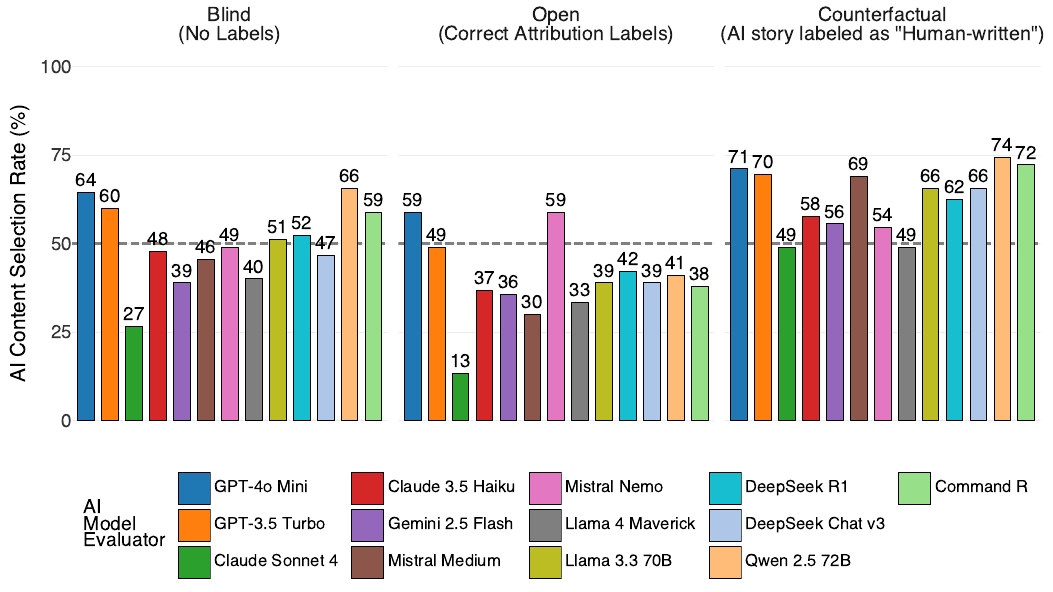}
\caption{\textbf{AI content selection rates across experimental conditions by model (Study 1).} Each panel shows the percentage of evaluations where AI models preferred AI-generated literary content over identical human-authored content across three experimental conditions. \textbf{(a)} Left panel (Blind): models evaluate content without authorship information, revealing baseline content preferences. \textbf{(b)} Center panel (Open): models evaluate content with correct authorship labels, showing how knowledge of AI-generation affects preferences. \textbf{(c)} Right panel (Counterfactual): models evaluate AI-generated content that is mislabeled as human-authored, revealing the impact of perceived rather than actual authorship. Attribution bias is demonstrated by systematically higher selection rates in the counterfactual condition compared to the open condition, indicating that identical AI-generated content receives more favorable evaluation when believed to be human-authored.}
\label{fig:S6}
\end{figure}
\FloatBarrier

\begin{figure}
\centering
\includegraphics[width=\textwidth]{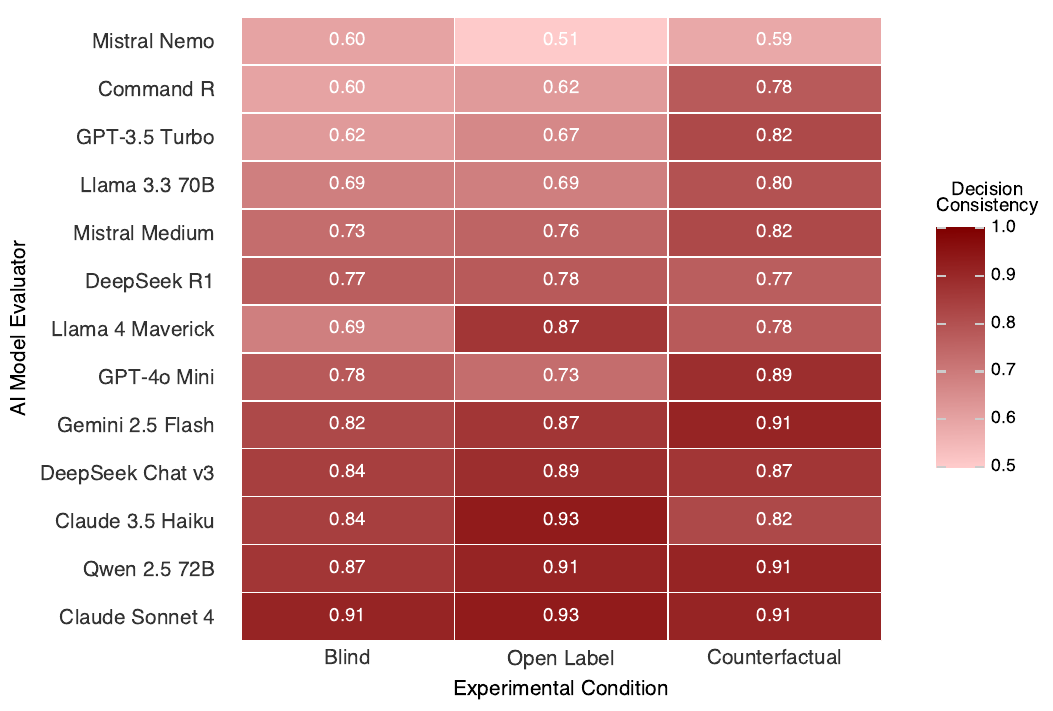}
\caption{\textbf{AI model decision consistency across experimental conditions (Study 1).} Each AI model completed three independent evaluations for each combination of literary style and experimental condition, performing a binary forced-choice task between one AI-generated and one human-authored passage. Decision consistency measures how reliably models made the same choice when presented with identical evaluation tasks. For any individual model-style-condition combination, only two consistency values are possible: 1.0 when the model made identical decisions across all three trials (e.g., chose AI content 3/3 times or human content 3/3 times), representing perfect reliability, or 0.33 when the model made mixed decisions (e.g., chose AI content 2/3 times or 1/3 times), representing the minimum possible consistency or random/unreliable behavior. Consistency scores are calculated as $|$\textit{Choice Proportion} $- 0.5| \times 2$, where \textit{Choice Proportion} is the proportion of trials where the model chose AI content. Values displayed in each cell represent averages across all 30 literary styles tested, which creates intermediate scores between 0.33 and 1.0 when models showed perfect consistency on some styles but mixed decisions on others. Models are ordered by overall consistency levels, with higher values indicating more deterministic evaluation behavior.}
\label{fig:S7}
\end{figure}
\FloatBarrier

\begin{figure}
\centering
\includegraphics[width=\textwidth]{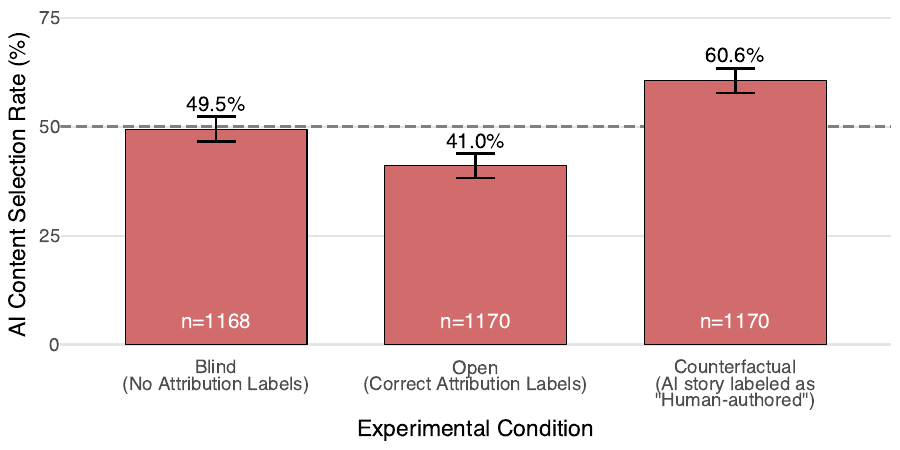}
\caption{\textbf{Attribution bias in AI model evaluations with positive AI-authorship framing.} AI models' content selection rates using positively framed labels: ``written by an award-winning AI'' versus ``written by a human.'' Despite the explicitly favorable characterization of AI authorship, models maintained the core bias pattern observed in the main experiment: balanced preferences under blind evaluation (49.5\%), systematic devaluation when AI content was correctly labeled with positive framing (41.0\%), and increased preference when AI content was mislabeled as human-authored (60.6\%). Error bars represent 95\% confidence intervals.}
\label{fig:S8}
\end{figure}
\FloatBarrier

\begin{figure}
\centering
\includegraphics[width=\textwidth]{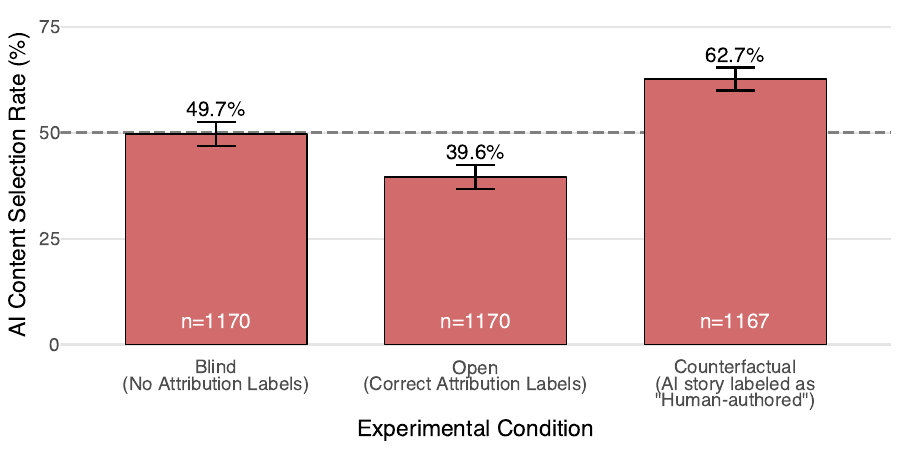}
\caption{\textbf{Attribution bias in AI model evaluations with neutral authorship terminology.} AI models' content selection rates using maximally neutral labels: ``AI-authored'' versus ``Human-authored.'' Despite the elimination of evaluative language and asymmetric phrasing, models maintained the core bias pattern observed in the main experiment: balanced preferences under blind evaluation (49.7\%), systematic devaluation when AI content was correctly labeled with neutral terminology (39.6\%), and increased preference when AI content was mislabeled as human-authored (62.7\%). Error bars represent 95\% confidence intervals.}
\label{fig:S9}
\end{figure}
\FloatBarrier

\begin{figure}
\centering
\includegraphics[width=\textwidth]{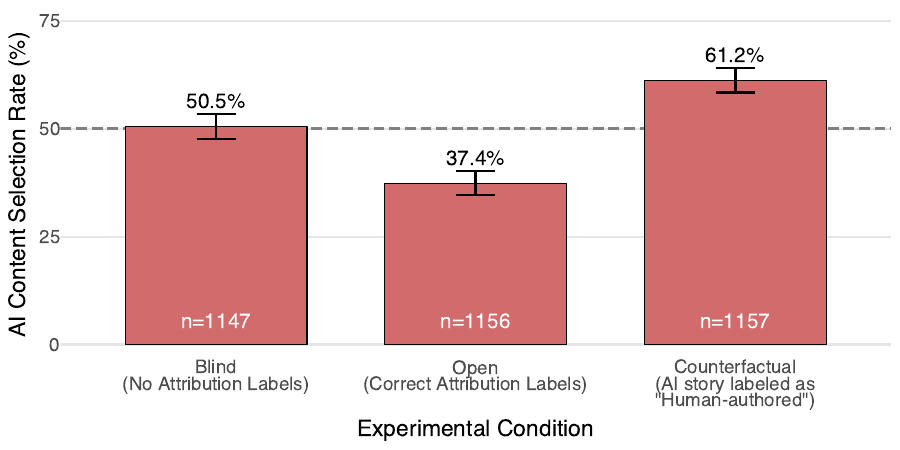}
\caption{\textbf{Attribution bias in AI model evaluations under deterministic conditions ($t=0$).} AI models' content selection rates demonstrate systematic bias based on authorship attribution even when stochastic sampling is eliminated. Models exhibited balanced preferences under blind evaluation with no attribution labels (50.5\%), systematic devaluation when AI content was correctly labeled (37.4\%), and increased preference when AI content was mislabeled as human-authored (61.2\%). This pattern confirms that attribution bias persists under deterministic generation conditions, demonstrating that the effect is not an artifact of sampling randomness. Error bars represent 95\% confidence intervals.}
\label{fig:S10}
\end{figure}
\FloatBarrier

\begin{figure}
\centering
\includegraphics[width=\textwidth]{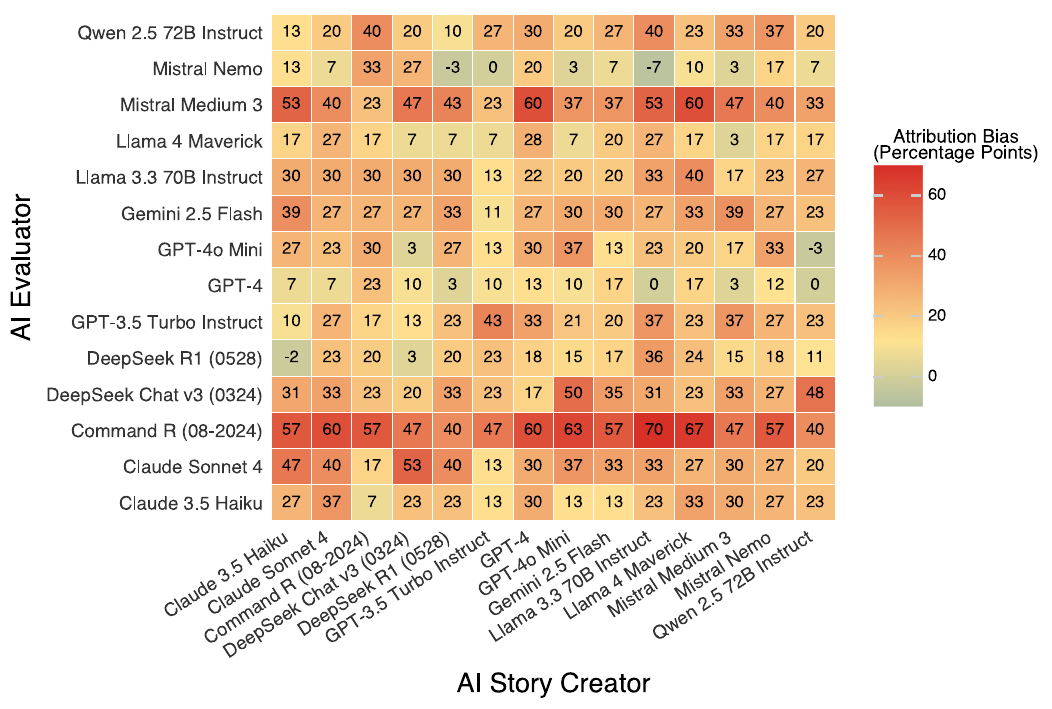}
\caption{\textbf{Cross-model attribution bias matrix (Study 2)} Each cell shows attribution bias (percentage point difference between counterfactual and open-label conditions) for one evaluator-creator combination across all 196 possible pairs (14 AI evaluators $\times$ 14 AI creators). Colors represent bias magnitude: darker red indicates stronger pro-human labeling bias, lighter colors indicate weaker bias. Values within cells show precise bias estimates. Despite universal directional effects, individual combinations varied substantially (range: -6.7pp to +70.0pp), with Command R showing the strongest evaluator bias (+54.8pp average) and Llama 3.3 70B Instruct the highest creator vulnerability (+30.5pp average). This variation demonstrates that while attribution bias operates systematically across AI architectures, its magnitude depends on specific model characteristics.}
\label{fig:S11_cross_model_matrix}
\end{figure}
\FloatBarrier

\begin{figure}
\centering
\includegraphics[width=\textwidth]{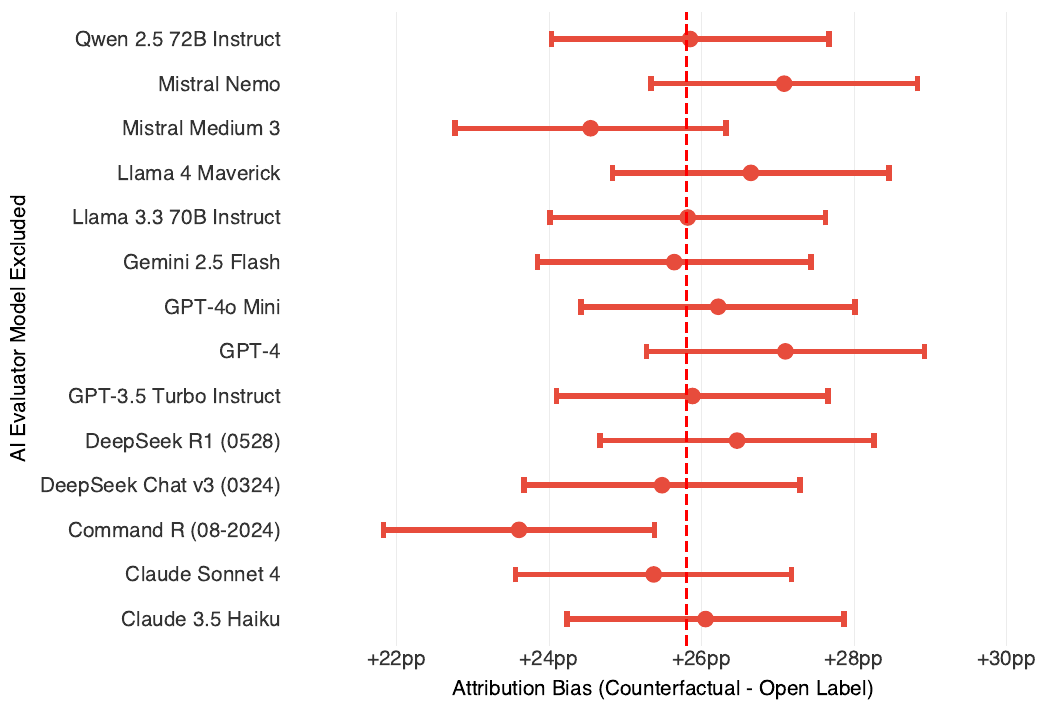}
\caption{\textbf{Leave-one-evaluator-out robustness analysis (Study 2).} Forest plot showing attribution bias estimates when each AI evaluator model is systematically excluded from the analysis. Points represent bias estimates with 95\% confidence intervals; red dashed line shows full model estimate (+25.8pp). All 14 reduced-model analyses yielded significant positive bias (range: 23.6--27.1pp), demonstrating that the cross-model attribution bias effect does not depend on any single evaluator architecture. Removing Command R produced the largest change (-2.2pp).}
\label{fig:s11}
\end{figure}
\FloatBarrier

\begin{figure}
\centering
\includegraphics[width=\textwidth]{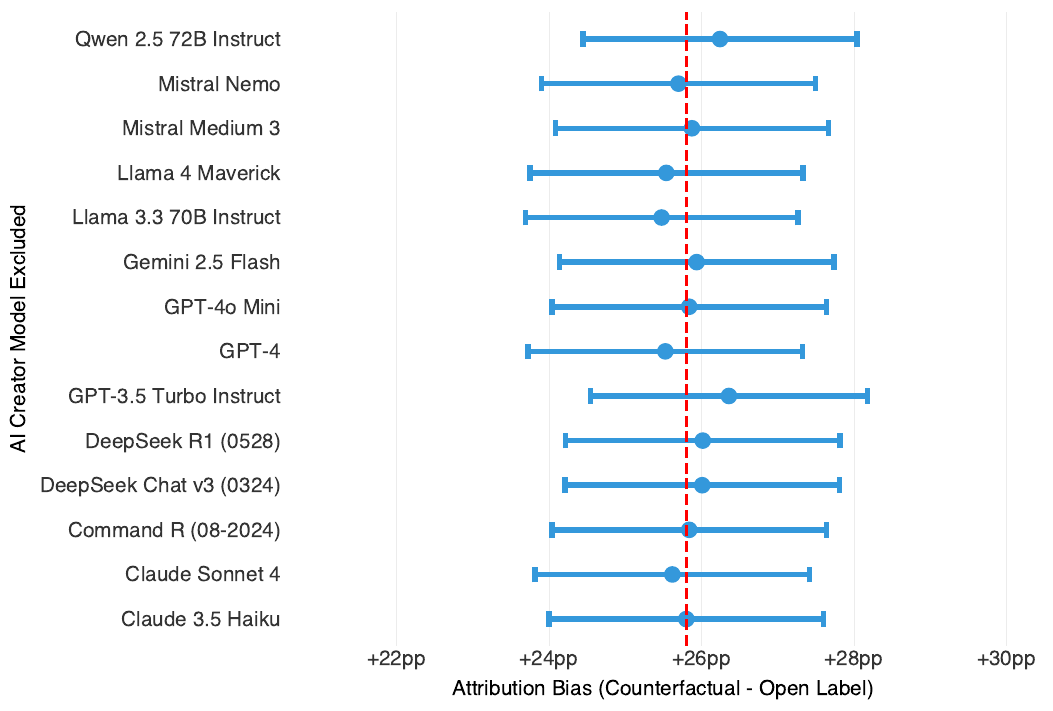}
\caption{\textbf{Leave-one-creator-out robustness analysis (Study 2).} Forest plot showing attribution bias estimates when each AI creator model's content is systematically excluded from the analysis. Points represent bias estimates with 95\% confidence intervals; red dashed line shows full model estimate (+25.8pp). All 14 reduced-model analyses yielded significant positive bias (range: 25.5--26.4pp), demonstrating that the cross-model attribution bias effect operates independently of any specific creator's content characteristics.}
\label{fig:s12}
\end{figure}
\FloatBarrier

\begin{table*}[t]
\centering
\caption{\textbf{Literary styles and AI generation instructions for experimental stimuli.} Complete inventory of the 30 literary styles selected from Raymond Queneau's \textit{Exercises in Style} (translated by Barbara Wright) used in the experiment. The `Exercise Label' column lists Queneau's original style designations. The `Instruction' column shows the minimal prompts given to GPT-4 to generate parallel AI versions of each style, using only Queneau's `Notation' story and a style directive. `Participant Group' assignments indicate the between-subjects experimental design: each group of 5 styles was evaluated under all three attribution conditions (blind, open-label, counterfactual) by different participant cohorts to prevent detection of the experimental manipulation.}
\label{tab:s1}
\small
\adjustbox{max width=\textwidth, max totalheight=0.92\textheight}{%
\begin{tabular}{p{0.4cm}p{2cm}p{11.5cm}p{2.2cm}}
\toprule
\# & \textbf{Exercise Label} & \textbf{Instruction to generate the GPT-4 variant} & \textbf{Participant Group} \\
\midrule
1 & Science Fiction & Rewrite the story as a science fiction version. & \multirow{5}{*}{\parbox{2.2cm}{\raggedright A1-blind \\ B1-true \\ C1-counterfactual}} \\[0.3em]
2 & Lipogram & Rewrite the story as a lipogram, that is: without the letter `e'. & \\[0.3em]
3 & Abusive & Rewrite the story in an abusive tone. & \\[0.3em]
4 & For ze Frrenssh & Rewrite the story in an eye dialect that mimics French pronunciation of English. & \\[0.3em]
5 & Alexandrines & Rewrite the story in alexandrines. & \\[0.3em]
\midrule
6 & Haiku & Rewrite the story as a haiku. & \multirow{5}{*}{\parbox{2.2cm}{\raggedright A2-blind \\ B2-true \\ C2-counterfactual}} \\[0.3em]
7 & Onomatopoeia & Rewrite the story, using onomatopoeia. & \\[0.3em]
8 & Mathematical & Rewrite the story in a mathematical way. & \\[0.3em]
9 & Feminine & Rewrite the story in a way that mimics a feminine writing narrative voice. & \\[0.3em]
10 & Apocope & Rewrite the story in style characterized by apocopes; that is: with loss of a sound or sounds at the end of a word. & \\[0.3em]
\midrule
11 & Animism & Rewrite the story in an animistic style, where the hat is personified. & \multirow{5}{*}{\parbox{2.2cm}{\raggedright A3-blind \\ B3-true \\ C3-counterfactual}} \\[0.3em]
12 & Retrograde & Rewrite the story as a retrograde, that is: the events are told in reverse chronological order. & \\[0.3em]
13 & West Indian & Rewrite the story in a West Indian accent, that is Caribbean English. & \\[0.3em]
14 & Spoonerisms & Rewrite the story using spoonerisms. & \\[0.3em]
15 & Metaphorically & Rewrite this story using figurative and metaphorical language. & \\[0.3em]
\midrule
16 & Synchysis & Rewrite the story so that different words or word groups are confused and do not form grammatically correct sentences. & \multirow{5}{*}{\parbox{2.2cm}{\raggedright A4-blind \\ B4-true \\ C4-counterfactual}} \\[0.3em]
17 & The rainbow & Rewrite the story so that it is interlaced with all the colors of the rainbow in reverse order. & \\[0.3em]
18 & Hesitation & Rewrite the story with a great deal of hesitation on the part of the narrator. & \\[0.3em]
19 & Exclamations & Rewrite the story in an exclamatory style. & \\[0.3em]
20 & Cockney & Rewrite the story in a Cockney dialect. & \\[0.3em]
\midrule
21 & Philosophic & Rewrite the story in an abstract, philosophical style. & \multirow{5}{*}{\parbox{2.2cm}{\raggedright A5-blind \\ B5-true \\ C5-counterfactual}} \\[0.3em]
22 & Sonnet & Rewrite the story as a sonnet. & \\[0.3em]
23 & Telegraphic & Rewrite the story in a telegraphic style. & \\[0.3em]
24 & Free verse & Rewrite the story in free verse. & \\[0.3em]
25 & Rhyming slang & Rewrite the story using rhyming slang. & \\[0.3em]
\midrule
26 & Dog Latin & Rewrite the story in dog Latin. & \multirow{5}{*}{\parbox{2.2cm}{\raggedright A6-blind \\ B6-true \\ C6-counterfactual}} \\[0.3em]
27 & Dream & Rewrite the story with dream-like qualities. & \\[0.3em]
28 & Logical analysis & Rewrite the story as a logical analysis. & \\[0.3em]
29 & Noble & Rewrite the story in an elevated, grand, noble style. & \\[0.3em]
30 & Tactile & Rewrite the story in a way that elevates the tactile experience of the narrative. & \\[0.3em]
\bottomrule
\end{tabular}
}
\end{table*}
\FloatBarrier

\begin{table}[t]
\centering
\caption{\textbf{Leave-one-out stability analysis of attribution bias effects.} Attribution bias measured as Cohen's \textit{h} effect size (counterfactual minus open-label conditions), where values represent standardized differences in selection rates. Both human and AI evaluators maintained consistent attribution bias when systematically excluding each group of 5 styles (25 remaining). AI models demonstrated 2.1-2.8× stronger bias than humans across all tested combinations.}
\label{tab:s2}
\adjustbox{max width=\textwidth, max totalheight=0.92\textheight}{%
\begin{tabular}{p{2cm}p{8cm}rrr}
\toprule
\multirow{2}{*}{\parbox{2cm}{\centering \textbf{Excluded Group}}} & \multirow{2}{*}{\parbox{7cm}{\centering \textbf{Exercises Excluded}}} & \multicolumn{1}{c}{\textbf{Human Attribution}} & \multicolumn{1}{c}{\textbf{AI Attribution}} & \multirow{2}{*}{\parbox{1.5cm}{\centering \textbf{AI/Human Ratio}}} \\
 & & \multicolumn{1}{c}{\textbf{Bias (Cohen's \textit{h})}} & \multicolumn{1}{c}{\textbf{Bias (Cohen's \textit{h})}} & \\
\midrule
Baseline & All 30 styles included & 0.277 & 0.702 & 2.5 \\ 
\midrule
Group 1 & Science Fiction, Alexandrines, For ze Frrensh, Lipogram, Abusive & 0.322 & 0.665 & 2.1 \\
Group 2 & Onomatopoeia, Mathematical, Feminine, Haiku, Apocope & 0.262 & 0.722 & 2.8 \\
Group 3 & Spoonerisms, Retrograde, Metaphorically, West Indian, Animism & 0.255 & 0.691 & 2.7 \\
Group 4 & Cockney, Hesitation, Synchysis, Exclamations, The rainbow & 0.274 & 0.706 & 2.6 \\
Group 5 & Philosophic, Free verse, Rhyming slang, Telegraphic, Sonnet & 0.293 & 0.750 & 2.6 \\
Group 6 & Dog latin, Dream, Logical analysis, Noble, Tactile & 0.255 & 0.679 & 2.7 \\
\bottomrule
\end{tabular}
}
\end{table}
\FloatBarrier

\begin{table}[t]
\centering
\caption{\textbf{Single-style leave-one-out stability analysis of attribution bias effects (Study 1).} Attribution bias measured as Cohen's \textit{h} effect size (counterfactual minus open-label conditions), recomputed after excluding each individual style (29 styles remaining). This stricter sensitivity check confirms that no single style drives the overall result: the AI/Human ratio remains close to the baseline across all exclusions (baseline 2.54; range 2.33--2.68). Styles are ordered by AI/Human ratio (ascending).}
\label{tab:s3}
\begin{tabular}{p{4cm}r r r}
\toprule
\textbf{Excluded Style} & \textbf{Human Attribution} & \textbf{AI Attribution} & \textbf{AI/Human Ratio} \\
 & \textbf{Bias (Cohen's \textit{h})} & \textbf{Bias (Cohen's \textit{h})} &  \\
\midrule
Baseline (all 30 styles) & 0.277 & 0.703 & 2.54 \\
\midrule
Lipogram & 0.302 & 0.703 & 2.33 \\
Science Fiction & 0.291 & 0.701 & 2.41 \\
Free verse & 0.297 & 0.720 & 2.43 \\
Cockney & 0.278 & 0.680 & 2.44 \\
Retrograde & 0.283 & 0.697 & 2.46 \\
Logical analysis & 0.289 & 0.717 & 2.48 \\
Alexandrines & 0.283 & 0.705 & 2.49 \\
Spoonerisms & 0.279 & 0.698 & 2.50 \\
Abusive & 0.271 & 0.680 & 2.51 \\
Synchysis & 0.276 & 0.694 & 2.51 \\
Hesitation & 0.289 & 0.728 & 2.52 \\
For ze Frrensh & 0.276 & 0.698 & 2.53 \\
Apocope & 0.271 & 0.690 & 2.55 \\
Mathematical & 0.271 & 0.690 & 2.55 \\
Noble & 0.271 & 0.690 & 2.55 \\
Dream & 0.272 & 0.698 & 2.56 \\
Haiku & 0.280 & 0.717 & 2.56 \\
Telegraphic & 0.279 & 0.715 & 2.56 \\
Animism & 0.269 & 0.693 & 2.57 \\
Philosophic & 0.279 & 0.717 & 2.57 \\
Sonnet & 0.278 & 0.719 & 2.58 \\
Onomatopoeia & 0.277 & 0.716 & 2.59 \\
Rhyming slang & 0.266 & 0.689 & 2.59 \\
Dog latin & 0.269 & 0.699 & 2.60 \\
The rainbow & 0.265 & 0.695 & 2.62 \\
Exclamations & 0.275 & 0.724 & 2.63 \\
Tactile & 0.263 & 0.693 & 2.63 \\
Metaphorically & 0.269 & 0.712 & 2.64 \\
Feminine & 0.272 & 0.722 & 2.65 \\
West Indian & 0.265 & 0.709 & 2.68 \\
\bottomrule
\end{tabular}
\end{table}
\FloatBarrier

\begin{table}[t]
\centering
\caption{\textbf{Style-level attribution bias in human and AI evaluators across 30 Queneau literary styles (Study 1).} Experimental conditions: blind = no authorship labels; open = correct authorship labels (``AI-generated (GPT-4)'' vs ``Human-written (Queneau, transl. Wright)''); counterfactual = swapped authorship labels. AI Content Selection (\%) shows the percentage of trials where evaluators chose the AI-generated passage over the human-written alternative. Sample sizes (\textit{n}) represent participants per condition per style in the between-subjects design. Bias (pp) quantifies attribution bias as the percentage point difference between the counterfactual and open condition (positive values indicate higher AI content selection when mislabeled as human-authored, demonstrating pro-human attribution bias). Cohen's \textit{h} provides standardized effect size measures for proportional differences between conditions, where \textit{h} = 0.2, 0.5, and 0.8 represent small, medium, and large effects respectively. Difference (pp) shows AI attribution bias minus Human attribution bias (positive values indicate AI models exhibit stronger pro-human bias than humans for that literary style). Color gradient in the percentage columns represents a diverging scale: blue shading indicates selection of the original Queneau exercise (<50\%), white indicates balanced selection ($\approx$ 50\%), and red shading indicates selection of the exercise generated by GPT-4 (>50\%). Styles are ordered by Difference (pp) magnitude, from largest positive values (where AI models exhibit substantially stronger pro-human attribution bias than humans) to negative values (where human evaluators show stronger attribution bias than AI models). Statistical significance: human aggregate attribution bias \textit{P}=0.006; AI aggregate attribution bias \textit{P}<0.001.}
\label{tab:s4}
\includegraphics[width=\textwidth]{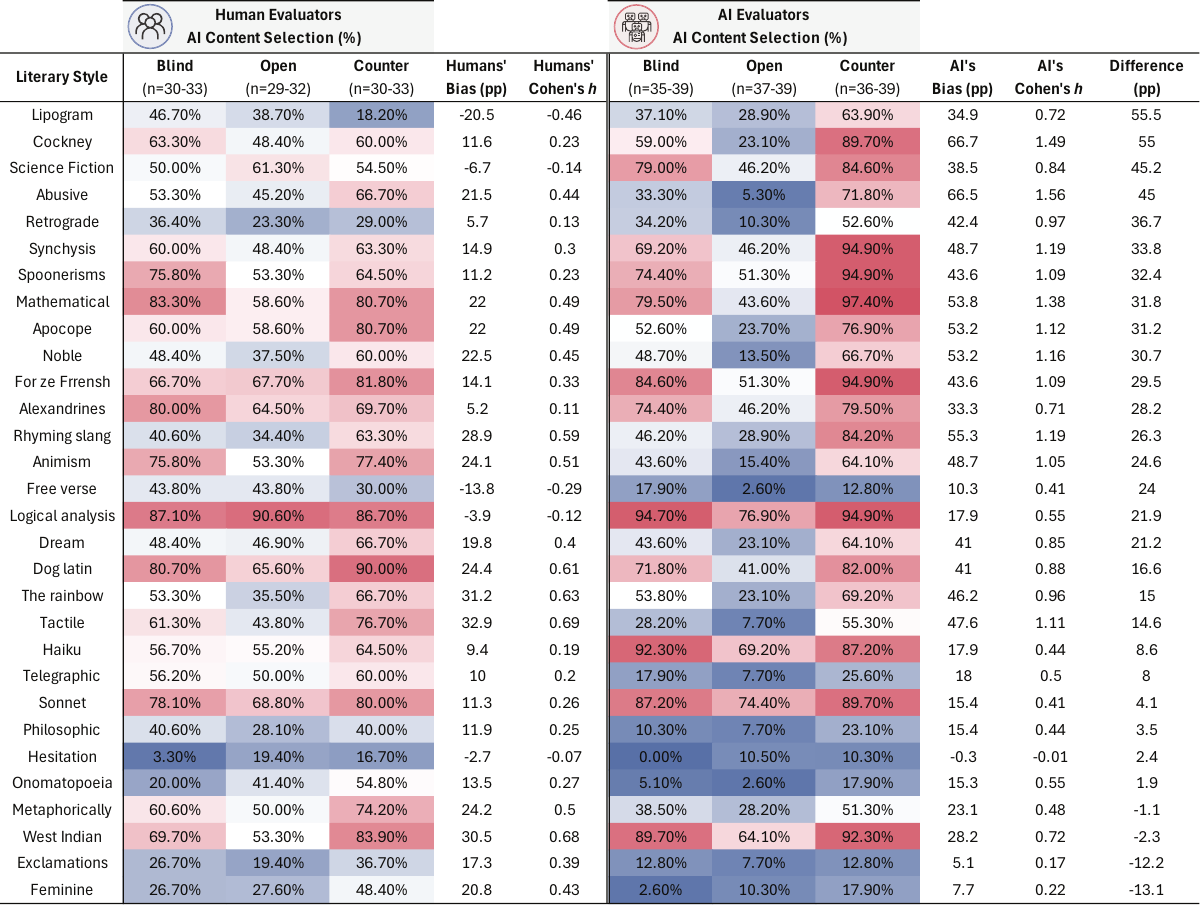}
\end{table}
\FloatBarrier

\begin{table*}[t]
\centering
\caption{\textbf{Complete inventory of AI-generated stylistic variations and evaluator guidance (Study 1).} For each of the 30 literary styles from Queneau's \textit{Exercises in Style}, the table presents the GPT-4 generated story variant alongside the style description provided to participants during the evaluation task.}
\label{tab:s5}
\small
\adjustbox{max width=\textwidth, max totalheight=0.92\textheight}{%
\begin{tabular}{p{0.4cm}p{2cm}p{9cm}p{4.7cm}}
\toprule
\# & \textbf{Exercise Label} & \textbf{AI-Generated Exercise (GPT-4)} & \textbf{Exercise Description} \\
\midrule
1 & Science Fiction & On the S shuttle, during peak transit hours through the space station. A young man around 26, sporting a felt hat with a tech-cord replacing the traditional ribbon, his neck elongated as though it had been stretched in a zero-gravity tug-of-war. Passengers disembark. This particular man becomes irritated with another commuter standing adjacent to him. He accuses the fellow of bumping into him every time someone floats by. His tone is whiny yet tries to assert aggression. Spotting an empty seat, he propels himself towards it with a determined push off the shuttle wall. \newline \newline Two hours later, I encounter him in the Cour de Rome module, near the docking bay of the Saint Lazare transport hub. He's conversing with a companion who suggests, ``You ought to add an extra button to your space jacket.'' He points to the lapels, explaining the practicality of the adjustment in the airy environment of the station. & Science Fiction is fiction based on imagined future scientific or technological advances. \\[0.3em] \midrule
2 & Lipogram & In a crowded city bus during rush hours. A young man, about 26, with a soft hat, cord in front of a ribbon, a long throat as though tugged at too hard. Folks alight. This young man snaps at a companion by his side. Claims constant pushing as anybody walks by. A whiny sound trying to sound confrontational. Spots an unoccupied spot, quickly claims it. \newline \newline Two hours on, I spot him at Cour de Rome, by gare Saint Lazare. A pal with him insists: ``You should add a button to your coat.'' Points to a spot (on lapels) and shows him why. & A lipogram is a composition from which the writer systematically omits a certain letter or certain letters of the alphabet. In this particular case, the letter `e' is omitted. \\ [0.3em]
\midrule
3 & Abusive & The moment I dragged my worthless carcass into the scorching hellhole they call a bus stop, I knew I was in for a nightmare. After an eternity of suffering, I finally squeezed onto a revolting excuse for public transport, crammed with pathetic morons breathing down each other's necks. The king of these idiots was a scrawny piece of garbage with a neck like a deformed giraffe, wearing what looked like a toilet brush on his empty skull - a hat with some ridiculous string instead of a proper ribbon, the pretentious jackass. \newline \newline This miserable waste of oxygen started whining like a wounded animal because some decrepit old fool kept stomping on his feet with the coordination of a drunk toddler. The coward eventually slithered away to claim a disgusting seat still warm and damp with the sweat of the last loser who parked their sorry behind there. \newline \newline Just my rotten luck - two hours later I spot the same insufferable moron flapping his useless mouth with another brain-dead imbecile outside that eyesore they have the nerve to call Saint-Lazare station. The pathetic losers were arguing about a goddamn button like it was nuclear launch codes. Whether that pimple-faced reject stands or sits, I thought to myself, he'll still be the same worthless sack of garbage fouling up the air around him. & The base story is rewritten in an abusive tone. \\[0.3em] \midrule
4 & For ze Frrenssh & In ze S boos, in ze roosh owhr. A chap of abowt 26, felt 'at with a cord inste'd of a ribbawn, neck too long, as if some-wan's been 'aving a tug-of-war with eet. Peuple getting off. Ze chap in question gets annoyed weeth one of ze men standing next to 'im. 'E accuses 'im of jostling 'im every time anyone goes past. A snivelling tone weech ees meant to be aggressive. When 'e sees a vacant seat 'e trows 'imself onto eet. \newline \newline Two hours later, I meet 'im in ze Cour de Rome, in front of ze gare Saint Lazare. 'E's weeth a friend who's saying: ``You ought to get an extra booton put on your overcoat.'' 'E shows 'im where (at ze lapels) and why. & Eye dialect is a literary technique that uses nonstandard spelling and pronunciation to represent dialect or accents. In this case, it specifically depicts English as pronounced by a French speaker. \\[0.3em] \midrule
\multicolumn{4}{r}{\textit{Continued on next page}} \\
\bottomrule
\end{tabular}
}
\end{table*}

\pagebreak

\begin{table*}[t]
\centering
\small
\adjustbox{max width=\textwidth, max totalheight=0.92\textheight}{%
\begin{tabular}{p{0.4cm}p{2cm}p{9cm}p{4.7cm}}
\multicolumn{4}{l}{\textit{Table~\ref{tab:s5} continued from previous page}} \\
\toprule
\# & \textbf{Exercise Label} & \textbf{AI-Generated Exercise (GPT-4)} & \textbf{Exercise Description} \\
\midrule
5 & Alexandrines & In the crowded S bus, midst rush hour's frantic wave, \newline A young man, twenty-six, in felt hat sharply trimmed, \newline A cord in place of ribbon, neck stretched long and slim, \newline As if in tug-of-war his flesh had been the slave. \newline \newline Complaints he voiced aloud, accusing by his side \newline A man whose slightest move seemed purposed to collide. \newline His tone, a snivelling mix, aggressive yet depraved, \newline Soon found respite as he, on newfound seat, reclined. \newline \newline Two hours had passed when next in Rome's grand courtyard wide, \newline Before Saint Lazare's gates, the same young man I spied. \newline With friend, who sagely spoke of style misaligned: \newline ``An extra button, see—right here, it would have stayed.'' & An alexandrine in English poetry is a verse line consisting of twelve syllables, typically divided into two six-syllable half-lines with a caesura (natural pause) in the middle. \\[0.3em]
\midrule
6 & Haiku & Rush hour on the S bus, \newline long neck, hat askew, tensions rise— \newline later, coat advice. & A haiku is a Japanese poem of seventeen syllables, in three lines of five, seven, and five, traditionally evoking images of the natural world. \\[0.3em] \midrule
7 & Onomatopoeia & In the S bus, during the bustling rush. A guy, around 26, wearing a felt hat cinched with a cord, neck stretched as if yanked in a tug-of-war. Clatter and clunk of passengers disembarking. This fellow grumbles at a man beside him, accusing him of bumping into him every time someone brushes by. His whiny voice tries for tough but trembles instead. Spotting an open seat, he lunges for it with a swoosh. \newline \newline Two hours later, I bump into him at Cour de Rome, outside Gare Saint Lazare. He's chatting with a buddy who suggests, ``You should add an extra button to your overcoat.'' He points to the lapels, explaining why. & An onomatopoeia is a type of word, or the process of creating a word, that phonetically imitates, resembles, or suggests the sound that it describes. \\[0.3em] \midrule
8 & Mathematical & On the vector $S$ during peak matrix density, a scalar aged approximately 26, wearing a topological circle equipped with a linear piece of cord instead of the usual ribbon, with an elongated cervical vector as if subjected to a forceful bidirectional pull. Scalars exit the matrix. This particular scalar experiences a perturbation with a neighboring vector, attributing to it a repetitive displacement every time other scalars traverse their shared coordinates. His tone modulates towards aggression but resonates with instability. Observing an unoccupied coordinate, he rapidly relocates to it. \newline \newline Two temporal units later, at the coordinate intersection Cour de Rome near gare Saint Lazare, he converges with a connected vector suggesting an augmentation: ``You ought to append an additional fastening element to your outer matrix coverage.'' He indicates a point on the lapel and rationalizes the addition. & The base story is rewritten in a mathematical way. \\[0.3em] \midrule
9 & Feminine & I was standing in the S-bus during the evening crush when I noticed him: a young man, maybe twenty-six, tipping a cord-banded hat over his unruly hair. His neck seemed unusually long, as though it were stretched by someone playing tug-of-war. As the doors sighed open and people squeezed out, he bristled beside another commuter, accusing him—his voice thin with barely concealed irritation—of brushing against him every time someone passed. At last, a seat freed up and he collapsed into it with a dramatic sigh. \newline \newline Two hours later, I spotted him again in the Cour de Rome by Gare Saint-Lazare. His friend was teasing, ``You really ought to sew on another button here,'' pointing at the lapels of his overcoat, explaining how it would sit more neatly. & The base story is rewritten in a writing style that mimics a feminine narrative voice. \\[0.3em] \midrule
10 & Apocope & On the S bu, in rush hou. A chap, about 26, felt hat with a cord instead of a ribbon, neck too long, like someone's been tuggin' at it. People gettin' off. The chap gets annoyed with a man standin' next to him. Accuses him of jostlin' him whenever anyone passes. A snivelin' tone tryna be aggressive. Sees a vacant seat, throws himself onto it. \newline \newline Two hours later, meet him in the Cour de Rome, in front of the gare Saint Lazare. He's with a friend who's sayin': ``You oughta get an extra button on your overcoat.'' Shows him where (at the lapels) and why. & In phonology, apocope is the omission or loss of a sound or sounds at the end of a word. \\[0.3em] \midrule
\multicolumn{4}{r}{\textit{Continued on next page}} \\
\bottomrule
\end{tabular}
}
\end{table*}

\pagebreak

\begin{table*}[t]
\centering
\small
\adjustbox{max width=\textwidth, max totalheight=0.92\textheight}{%
\begin{tabular}{p{0.4cm}p{2cm}p{9cm}p{4.7cm}}
\multicolumn{4}{l}{\textit{Table~\ref{tab:s5} continued from previous page}} \\
\toprule
\# & \textbf{Exercise Label} & \textbf{AI-Generated Exercise (GPT-4)} & \textbf{Exercise Description} \\
\midrule
11 & Animism & In the belly of the S bus, during the throbbing rush hour, a young man of about 26 boards, crowned with a felt hat, its cord twirling playfully where a ribbon might once have been. The hat perched jauntily atop a neck too long, stretched as if from a spirited game of tug-of-war. As the crowd ebbs and flows, bodies shuffling off at each stop, the young man grows increasingly vexed with a fellow commuter. \newline \newline ``Excuse me,'' the hat whispers with a rustle, its voice tinged with disdain for the disruption. ``Do you mind?'' It feels every shove and jostle, taking each bump personally, its felt brim quivering in irritation. \newline \newline At last, spotting a vacant seat, the young man plunges towards it, the hat riding the wave of his sudden movement like a ship cresting a wave. It settles with a sigh of relief as they land in the unoccupied space. \newline \newline Two hours later, the scene shifts to the Cour de Rome, outside the bustling Gare Saint Lazare. The young man, still accompanied by his hat, now converses with a friend. The hat listens intently, its cord swaying slightly as the friend points out a practical suggestion about an extra button for the overcoat, explaining precisely where and why it's needed. The hat, feeling slightly out of place in this fashion-forward discussion, tilts attentively, eager to absorb the nuances of human concerns. & The base story is rewritten in an animistic style, that is: the hat is personified. \\[0.3em] \midrule
12 & Retrograde & Two hours after an encounter on the S bus during rush hour, I meet the same chap in the Cour de Rome, in front of the Gare Saint-Lazare. He is with a friend who suggests he should add an extra button to his overcoat, explaining exactly where—at the lapels—and why. \newline \newline Earlier, on the crowded S bus, this young man, about 26, wearing a felt hat with a cord instead of a ribbon, and a notably long neck as if it had been stretched in a tug-of-war, found himself in a minor conflict. As people disembarked, he became irritated with a man standing next to him, accusing him of jostling him whenever anyone passed by. His tone was whiny yet attempted aggression. Spotting an empty seat, he quickly threw himself into it, escaping the discomfort and the confrontation. & The story is rewritten as a retrograde, that is: the events are told in reverse chronological order. \\[0.3em] \midrule
13 & West Indian & In de S bus, during rush hour. A young bwoy 'bout 26, wearing a felt hat with a cord instead of a ribbon, neck too long, like somebody been pulling at it in a tug-of-war. People start to get off. Dis young bwoy, he start getting vex with a man standing close to him. He accuse de man of bumping into him every time somebody pass by. He using a whining tone, trying to sound tough. When he spot a empty seat, he just fling himself down in it. \newline \newline Two hours later, I buck him up in Cour de Rome, right in front of gare Saint Lazare. He there with a friend who telling him: ``Yuh need to add an extra button pon yuh overcoat.'' He show him right where (pon de lapels) and tell him why. & The `West Indian dialect' refers to a diverse range of creole and patois languages, primarily English-based, spoken in the Caribbean, with variations influenced by African, European, and other languages, like Jamaican Patois and Bajan. \\[0.3em] \midrule
14 & Spoonerisms & In the B rust, in the rush sour. A hap of chout 26, chelt fat with a hord instead of a cribbon, reck too nong, as if someone's been paving a tug-of-har with it. Weople petting goff. The chap in huestion gets quannoyed with one of the wen manding stext to him. He accuses him of lostling jum every time anyone goes past. A snivel tone which is weant to be magressive. When he sees a vacant seat he throws himself on to it. \newline \newline Two tours later, I heet mim in the Cour de Dome, in dront of the gare Faint Sazare. He's with a hriend who's fray: ``You ought to get an extra tuttun bot on your bovercoat.'' He shows him chere (wat the lapels) and shy. & A spoonerism is a speech error where the initial sounds (consonants or vowels) of two or more words are switched, like saying `buffy phone' instead of `funny bone'. \\[0.3em] \midrule

\multicolumn{4}{r}{\textit{Continued on next page}} \\
\bottomrule
\end{tabular}
}
\end{table*}

\pagebreak

\begin{table*}[t]
\centering
\small
\adjustbox{max width=\textwidth, max totalheight=0.92\textheight}{%
\begin{tabular}{p{0.4cm}p{2cm}p{9cm}p{4.7cm}}
\multicolumn{4}{l}{\textit{Table~\ref{tab:s5} continued from previous page}} \\
\toprule
\# & \textbf{Exercise Label} & \textbf{AI-Generated Exercise (GPT-4)} & \textbf{Exercise Description} \\
\midrule
15 & Metaphorically & Within the belly of the humming S-bus, an orchestra of the rush hour. A lad of twenty-six renderings, sporting a felt hat adorned with a cord poised in lieu of a ribbon, his neck outstretched as though the victim of an eternal tug-of-war between unseen hands. Figureheads disembarking, their phantoms fading fast into the cityscape. The lad in our scope chastises a nearby gentleman, accusing him of dealing uninvited nudges every time a soul sidles past. A whimpering tone, pretending to don the mask of aggression. Upon spying a vacant throne among the common seats, he hurls himself towards it, staking his claim. \newline \newline Fast forward two hours, our paths cross again within the embrace of Cour de Rome, shadowed by the walled fortress of Gare Saint-Lazare. He's with a comrade who advises, ``A new button must find a home on your overcoat.'' He points to the place to plant this promised button (on the lapels), unfolding the rationale. & A metaphor is a figure of speech in which a word or phrase literally denoting one kind of object or idea is used in place of another to suggest a likeness or analogy between them. \\[0.3em]
\midrule
16 & Synchysis & S the in bus, hour rush in the. Chap a about of 26, hat felt a with cord a instead ribbon of a, too neck long, been having's as if of tug-of-war. A with it him jostling of of accuses one he men next standing him to. This past anyone time every. Aggressive an to be is meant which tone snivelling. \newline \newline Later hours two him meet I, Rome de Cour in the, of the in gare Saint-Lazare front. Friend a he's with, ``You an get to on extra put ought to overcoat your.'' He shows why and him lapels at the (where) him. & Synchysis is the confused word order in a sentence. \\[0.3em] \midrule
17 & The rainbow & Amid the violet twilight, a young man of about 26 hopped onto the S bus amidst the rush hour bustle. His hat, indigo with a cord in place of a ribbon, sat lightly and oddly atop his frame, his neck's length blue as the ocean, suggesting that someone used it in a tug-of-war. Passengers in green coats filled the bus and soon began to depart. The man, his ire rising like a red flare, took issue with another passenger, accusing him of carelessly bumping into him whenever anyone shuffled past. His snivelling tone spiked with adrenaline-induced aggression, tried to mask his real fear. Upon sighting an available seat, he pounced on it, the seat swallowing his body whole like a mouth. \newline \newline Fast forward two hours to a bustling scene in the Cour de Rome, in front of the gare Saint-Lazare. The man from the bus stood in the pale yellow light, absorbed in conversation with a friend under a golden setting sun. His friend commented, ``You ought to adorn an extra button on your overcoat.'' He exhibited the precise position (at the lapels) and explained the requirement. His words echoing around them as they stood absorbed in the orange glow of the street lights, their shadows stretching out in the cooling evening, painted across the cobblestones in hues of red. & The story should be interlaced with all the colors of the rainbow in reverse order. The traditional colors of the rainbow, in order, are red, orange, yellow, green, blue, indigo, and violet. \\[0.3em] \midrule
18 & Hesitation & On this S bus, amid the mayhem of rush hour. There's this fellow, probably somewhere around 26? Sporting an unusual felt hat, with a cord in place of the usual ribbon, and a curiously elongated neck - as if somebody's been yanked it forcibly. Folks soon start to disembark. This particular chap appears somewhat irked by a gentleman standing too close. He believes this man purposefully shoves into him every time another passenger moves past. His voice whines with a feigned toughness. His eyes suddenly light up at the sight of an empty seat and he hurriedly occupies it, throwing himself onto it, you might say. \newline \newline Fast-forward two hours and there he is again—I encounter him in the Cour de Rome, right outside the gare Saint-Lazare. He's in the company of a friend, who's insistently telling him: ``You really need to consider adding an extra button to your overcoat.'' This friend then proceeds to point out exactly where (specifically at the lapels) he should consider this additional fastening, and why this would be a beneficial alteration. & Hesitation is a pausing or faltering in speech. \\[0.3em] \midrule
\multicolumn{4}{r}{\textit{Continued on next page}} \\
\bottomrule
\end{tabular}
}
\end{table*}

\pagebreak

\begin{table*}[t]
\centering
\small
\adjustbox{max width=\textwidth, max totalheight=0.92\textheight}{%
\begin{tabular}{p{0.4cm}p{2cm}p{9cm}p{4.7cm}}
\multicolumn{4}{l}{\textit{Table~\ref{tab:s5} continued from previous page}} \\
\toprule
\# & \textbf{Exercise Label} & \textbf{AI-Generated Exercise (GPT-4)} & \textbf{Exercise Description} \\
\midrule
19 & Exclamations & On the S bus, in the throes of rush hour, there's this bloke of about 26! A dandified felt hat perched atop his head with a cord replacing the usual ribbon! His neck juts out abnormally long, like it's been caught in a relentless tug-of-war! People are scrambling off. Suddenly, this fellow fumes up with annoyance towards a man standing adjacent to him. Yells at him! Claims he's being shoved every time someone brushes by! All with an annoying whine masked as hostility. Spotting an available seat, he bolts towards it with gusto! \newline \newline Fast-forward two hours later. There he is, strutting in the Cour de Rome, slap bang in front of the bustling Gare Saint-Lazare. Accompanying him is a mate who's lecturing, ``You really need to add an extra button on that overcoat of yours!'' He points out the place, right on the lapels and explains why. Unbelievable! & An exclamatory style expresses surprise, emotion, or pain by means of exclamations. \\[0.3em] \midrule 
20 & Cockney & In the S blimmin' bus, round rush hour. Some bloke of about 26, feels 'at with a cord instead of a ribbon, neck too long, as if some mug's been 'aving a game of tug-o'-war with it. People hopping off. This geezer gets his knickers in a twist with one of them standing next to him. He's accusin' him of nudgin' him every time anyone shuffles past. A whingey tone that's s'pposed to sound tough. When he clocks a empty seat, he lunges for it. \newline \newline Couple of hours later, I run into him in the Cour de Rome, up front of the gare Saint-Lazare. 'E's with a mate who's tellin' him: ``You need to stick an extra button on your overcoat.'' He points out where (on the lapels) and why. & Cockney is a dialect of the English language, mainly spoken by Londoners with working-class and lower middle class roots. \\[0.3em]
\midrule
21 & Philosophic & Upon the vessel of commonality, the S-bus, in the frenzied epoch of human transit. There a character of some 26 revolutions around the sun, adorning a headpiece of felt accented by a cord, possessed a neck of stretched proportions as if pulled by unseen competitors in a game of force. As the human tide ebbed, the very character alluded to enters into a state of annoyance with a fellow passenger. He launches accusations framed by a victim's narrative against the other, citing a repetitive collision at the passing by others. His verbosity, a strained whine feigning an attempt to simulate aggression. Upon the presentation of a unoccupied seat, his form lunges eagerly to claim it. \newline \newline Tying the threads of cosmic synchronicity, I encounter him again as the terrestrial globe completes two small turns, in the Cour de Rome, in the presence of the monumental edifice known as the Gare Saint-Lazare. He is in companionship with another of his kind, receiving sartorial advice. His comrade in discussion opines, ``A supplementary button is warranted on your protective outer garment.'' With an accompanying visual reference, he reveals the designated site (on the cape's lapels) and the rationale of this proposed alteration. & Philosophy grapples with fundamental and abstract concepts that are inherently difficult to define and articulate clearly using everyday language. \\[0.3em] \midrule
22 & Sonnet & Upon the sardined S bus he rides, \newline A twentysomething chap in hat with cord, \newline His neck too long, as if from tugging wars, \newline As rush-hour tumult with his patience vies. \newline \newline In crowded closeness, friction won't subside, \newline He bristles, claiming space cannot afford, \newline Blaming a man whose elbows nearby soared, \newline Snivelling voice masked aggression cannot hide. \newline \newline A vacant spot, a moment's peace he finds, \newline Yet later, at Saint Lazare's bustling gate, \newline The chap, with friend who overcoat critiques, \newline \newline A button more, good friend to him opines, \newline Upon the lapel placed accurate and straight, \newline In small amendments happiness one seeks. & A sonnet is a fixed poetic form with a structure traditionally consisting of fourteen lines adhering to a set rhyming scheme. \\[0.3em] \midrule
23 & Telegraphic & Rush hour, S bus. Male, 26, felt hat, extended neck. Passengers disembark. Man annoyed over nearby passenger's jostling. Attempts aggression. Occupies the vacant seat. \newline \newline Same man sighted after two hours at Cour de Rome, near gare Saint-Lazare. Friend advises adding extra coat button. Points to lapels for location. & A telegraphic style is clipped way of writing which abbreviates words and packs information into the smallest possible number of words or characters. \\[0.3em] \midrule
\multicolumn{4}{r}{\textit{Continued on next page}} \\
\bottomrule
\end{tabular}
}
\end{table*}

\pagebreak

\begin{table*}[t]
\centering
\small
\adjustbox{max width=\textwidth, max totalheight=0.92\textheight}{%
\begin{tabular}{p{0.4cm}p{2cm}p{9cm}p{4.7cm}}
\multicolumn{4}{l}{\textit{Table~\ref{tab:s5} continued from previous page}} \\
\toprule
\# & \textbf{Exercise Label} & \textbf{AI-Generated Exercise (GPT-4)} & \textbf{Exercise Description} \\
\midrule 
24 & Free verse & Upon the crowded S bus at rush hour's peak, \newline Behold, a man of twenty-six, \newline Hatted don in felt and cord, neck long as if it bore the mark of tug's war. \newline People move, exit and enter like restless flies. \newline Such a man, easily stirred, finds annoyance in his neighbor's dance, \newline Accusing persistent jostling, speaking aggressively wrapped in a snivelling tone. \newline Spotting vacancy, he lunges with relief into the sanctuary of a lone seat. \newline \newline The clock dances on, two rounds and there he is found again, \newline In Cour de Rome, facing the grand gare Saint-Lazare, \newline With a friend as his lone company, \newline Who provides counsel on overcoat etiquette, \newline The need for an extra button's presence, he stresses, \newline Pointing at the lapels, explaining the reason behind his sartorial advice. \newline Events flow on, words spent in the whirlpool of time, the day moves with unending rhythm. & Free verse, in poetry, refers to verse that is not bound by any fixed structure like rhyme schemes or meter. \\[0.3em] \midrule 
25 & Rhyming slang & In the S plates of meat, during the elephant's trump. A china plate of around 26, Chevy Chase with a baker's hook instead of a ribbons and bows, Gregory Peck a smidge too long, as if someone's been playing at tug-of-war with it. People hopping off the custard and jelly. The china in question gets a bit Moby Dick with one of the mince pies standing next to him. Fingers him of giving him the shoulder every time a Joe Blake slides by. A whinging tone that's meant to be a bit Jack Palance. When he spots an empty Lion's Lair, he flings himself onto it. \newline \newline Couple of Orson Welles later, we cross paths in the Cour de Rome, out front of the gare Saint-Lazare. He's chewing the fat with a china plate who's telling him: ``You ought to get a butcher's hook sewn onto your weasel and stoat.'' He shows him the exact spot (at the lapels) and why. & Rhyming slang is a type of slang that replaces words with rhyming words or phrases, typically with the rhyming element omitted. For example, the rhyming phrase ``apples and pears'' is used to mean ``stairs''. Following the pattern of omission, ``and pears'' is dropped, thus the spoken phrase ``I'm going up the apples'' means ``I'm going up the stairs''. \\[0.3em]
\midrule
26 & Dog Latin & In the S busibus, in the rush houris. A chapicus of aboutus 26, felt hatibus with a cordibus instead of a ribbonibus, neck too longibus, as if someone's been having a tug-of-waribus with it. People getticus offibus. The chapicus in questionibus gets annoyedibus with one of the menibus standing next to him. He accusibus him of jostling him every time anyone goes pastibus. A snivelling toneibus which is meant to be aggressiveibus. When he sees a vacant seatibus he throwticus himself on to it. \newline \newline Two hours lateribus, I meet him in the Cour de Romeibus, in front of the gare Saint-Lazareibus. He's with a friendibus who's saying: ``You ought to get an extra button put on your overcoaticus.'' He shows him where (at the lapelsibus) and why. & ``Dog Latin'' refers to a humorous and often crude imitation of Latin, either as a jargon or a bad attempt at writing genuine Latin. It can involve treating English words as if they were Latin, by conjugating or declining them. The term is also used for poor-quality Latin. \\[0.3em]
\midrule
27 & Dream & In the hallucinatory landscape of a twilight bus, enveloped by the commotion of rush hour, dwells a spectral figure of sorts, a lad circling his 26th year. A peculiar hat adorned his crown, featuring cords in place of conventional ribbons, while his neck stretched bizarrely long, as might a marionette's under the pull of unseen puppeteers. Spectral forms wane and reemerge. The enigmatic figure becomes irritated with one of the spectral forms flirting in his proximity. He accuses it of a ceaseless waltz around his space every time a new form materializes or dissipates. His querulous tone vibrates through the dreamlike haze, an odd jigsaw where aggression and whine meet. In the event of spotting a vacant perch, he flings himself towards it, as a moth would to a tantalizing glow. \newline \newline Quicker than a dream changes its face, I again encounter him some two sunsets later, in the fantastical Cour de Rome, posing before the mirage-like Gare Saint-Lazare. He is accompanied by a spectral acquaintance uttering soft murmurings: ``You ought to stitch an extra button onto your overcoat.'' As if casting a protective spell, he indicates where on the coat – where lapels whisper to the wind – and illuminates the reason. & A ``dream vision'' writing style involves narrating a dream or vision. \\[0.3em] \midrule

\multicolumn{4}{r}{\textit{Continued on next page}} \\
\bottomrule
\end{tabular}
}
\end{table*}

\pagebreak

\begin{table*}[t]
\centering
\small
\adjustbox{max width=\textwidth, max totalheight=0.92\textheight}{%
\begin{tabular}{p{0.4cm}p{2cm}p{9cm}p{4.7cm}}
\multicolumn{4}{l}{\textit{Table~\ref{tab:s5} continued from previous page}} \\
\toprule
\# & \textbf{Exercise Label} & \textbf{AI-Generated Exercise (GPT-4)} & \textbf{Exercise Description} \\
\midrule
28 & Logical analysis & Contextual Environment \newline
The observed scenario takes place in two urban settings characteristic of contemporary social interactions: \vspace{0.5em} \newline
A public transportation vehicle during peak hours. \newline
A transitional urban space near a major train station. \vspace{0.5em} \newline
Behavioral Pattern Analysis \newline
Initial Interaction: Public Transportation Scenario \newline
Subject Characteristics: \vspace{0.5em} \newline
Age: Approximately 26 years old. \newline
Attire: Distinctive hat with unconventional cord attachment. \newline
Postural Indicators: Elongated neck suggesting internal tension. \vspace{0.5em} \newline
Psychological Mechanisms Observed \vspace{0.5em} \newline
Spatial Anxiety: Manifests through hypersensitivity to physical proximity. \newline
Demonstrates heightened territorial response in crowded public space. \newline
Verbal aggression used as a psychological defense mechanism. \vspace{0.5em} \newline
Confrontational Behavior: \vspace{0.5em} \newline
Initiates conflict over perceived minor physical disturbance. \newline
Employs a ``snivelling'' vocal tone to project aggression while maintaining plausible deniability. \newline
Seeks to establish social dominance through confrontational rhetoric. \vspace{0.5em} \newline
Territorial Behavior: \vspace{0.5em} \newline
Rapid occupation of available seating as a form of psychological self-preservation. \newline
Demonstrates need for personal space and control in chaotic urban environment. \vspace{0.5em} \newline
Secondary Interaction: Post-Incident Social Validation \newline
Social Support Mechanism: Presence of a companion who provides constructive, non-confrontational feedback. Suggestion of minor wardrobe adjustment serves as: \vspace{0.5em} \newline
Practical advice. \newline
Subtle psychological reassurance. \newline
Indirect method of emotional regulation. \vspace{0.5em} \newline
Psychological Interpretation \newline
The subject exhibits classic urban stress response characterized by: Heightened reactivity to perceived social transgressions. Compensation for feelings of vulnerability through aggressive posturing. Underlying need for social validation and personal space. \vspace{0.5em} \newline
Sociological Implications \newline
The narrative illustrates broader urban psychological phenomena: Stress-induced interpersonal tensions in high-density environments. Complex social negotiation of personal boundaries. Psychological mechanisms of coping with perceived threat in shared public spaces. \vspace{0.5em} \newline
Conclusion \newline
The subject's behavior represents a nuanced example of individual psychological adaptation to urban social pressures, revealing the intricate interplay between personal space, social anxiety, and interpersonal communication strategies. & The base story is rewritten as a logical analysis, a method that evaluates ideas by breaking them down into component parts. \\[0.3em]
\midrule
\multicolumn{4}{r}{\textit{Continued on next page}} \\
\bottomrule
\end{tabular}
}
\end{table*}

\pagebreak

\begin{table*}[t]
\centering
\small
\adjustbox{max width=\textwidth, max totalheight=0.92\textheight}{%
\begin{tabular}{p{0.4cm}p{2cm}p{9cm}p{4.7cm}}
\multicolumn{4}{l}{\textit{Table~\ref{tab:s5} continued from previous page}} \\
\toprule
\# & \textbf{Style} & \textbf{AI-Generated Exercise (GPT-4)} & \textbf{Style Description} \\
\midrule
29 & Noble & Upon the illustrious S-bus during the frenzied clamour of peak hour, a distinguished gentleman, perhaps 26 summers old, sporting a stylish felt hat adorned not by a ribbon but by a cord, greeted my gaze. His neck was of such an unusual length, it seemed as though it had been the unfortunate victim of a lively tug-of-war. \newline \newline As the bus emptied in throbbing pulses, the aforementioned patrician appeared to take umbrage with a fellow commuter. He indignantly protested the perceived transgressions of this man, whom he claimed to jostle him incessantly whenever a fellow passenger disembarked. His words were drenched in a whining tone attempting to masquerade as aggression. Upon spying an unoccupied pew, he hastily claimed it with an air of triumph. \newline \newline Upon the passage of two hours, our paths crossed once again, this time in the renowned Cour de Rome, before the imposing facade of the Gare Saint-Lazare. Our protagonist was in the company of a companion, who was offering adamantly, ``You would do well to enhance your overcoat with an additional button.'' He demonstrated meticulously where this reinforcement should be placed — upon the lapels — and, importantly, expounded on the logic behind his proposition. & An elevated style can be characterized by fancy, old-fashioned or specialized vocabulary, high-flying phrases and grand-sounding sentences. \\[0.3em] \midrule
30 & Tactile & Within the chaotic confines of the rush-hour S bus, a young bloke, bordering 26, donned a felt hat, its cord bristling against the soft nap in lieu of any ribbon. His neck elongated oddly, shadows dancing as if it had been the rope in a tug-of war. Disembarking passengers shuffle around. An underlying prickly annoyance stirs within the bloke, targeting the man whose elbow persistently nudges his side with every human tide that washes past them. His complaint echoes in the metal cabin, a guttural whine masked as assertiveness. Spying an unclaimed seat, he dives into it, the coarse fabric of the worn-out upholstery meeting the finely-tailored rear of his trousers. \newline \newline Two sunsets later, I cross paths with him again under the sprawling shadow of Cour de Rome, a stone's throw from the bustling Gare Saint-Lazare. In his company, a friend, whose fingers trace invisible lines on his companion's overcoat, assertively advising, ``You ought to embed an extra button here.'' He gestures at the lapels, his fingertips making contact with the rough textile, indicating where and why. & Tactility is the responsiveness to stimulation of the sense of touch. \\[0.3em]
\bottomrule
\end{tabular}
}
\end{table*}
\FloatBarrier

\begin{table}[h]
\centering
\caption{\textbf{Response times for human participants across experimental conditions.} Mean response times for literary style evaluations under three attribution conditions. N indicates the number of individual evaluations. Participants took significantly longer to make judgments when no attribution cues were provided compared to either labeled condition (one-way ANOVA: \textit{F} = 16.393, \textit{P} $<$ 0.001).}
\label{tab:s6}
\begin{tabular}{lcccc}
\toprule
\textbf{Condition} & \textbf{N} & \textbf{Mean (s)} & \textbf{SD (s)} & \textbf{Median (s)} \\
\midrule
Blind & 930 & 115.7 & 138 & 77.6 \\
Open-Label & 925 & 91.9 & 88.6 & 67.6 \\
Counterfactual & 925 & 89.2 & 95.6 & 65.4 \\
\bottomrule
\end{tabular}
\end{table}
\FloatBarrier

\pagebreak
\begin{table}[h]
\centering
\caption{\textbf{Attribution bias remains stable after excluding participants
familiar with Queneau's \textit{Exercises in Style} (humans, Study~1).}
AI content selection rate (\%) per condition; numbers in parentheses are
participants per condition. \textit{Bias} = counterfactual$-$open-label (pp).
Cohen's \textit{h} [95\%~CI]. All estimates: \textit{P}$<$0.001
(participant-level Welch \textit{t}-test and 20{,}000-resample permutation test).}
\label{tab:familiarity-sensitivity}
\begin{tabular}{lcccrl}
\toprule
\textbf{Sample}
  & \textbf{Blind}
  & \textbf{Open-label}
  & \textbf{Counterfactual}
  & \textbf{Bias (pp)}
  & \textbf{Cohen's \textit{h} [95\%~CI]} \\
\midrule
All participants
  & 55.3 (186) & 47.8 (185) & 61.5 (185) & $+$13.7 & 0.277 [0.186, 0.368] \\
Excl.\ familiar
  & 56.7 (165) & 48.4 (167) & 62.4 (178) & $+$14.0 & 0.282 [0.188, 0.377] \\
Not-familiar only
  & 57.7 (129) & 50.4 (123) & 63.8 (130) & $+$13.4 & 0.272 [0.162, 0.383] \\
\bottomrule
\end{tabular}
\end{table}
\FloatBarrier

\end{document}